\PassOptionsToPackage{numbers,sort&compress}{natbib}
\documentclass{article}

\usepackage[preprint]{neurips_2026}

\makeatletter
\renewcommand{\@noticestring}{%
  \scriptsize
  $^{*}$Equal contribution\quad
  $^{\bigstar}$Project Lead\quad
  $^{\text{\Letter}}$Corresponding author%
}
\makeatother

\usepackage[utf8]{inputenc}
\usepackage[T1]{fontenc}
\usepackage[colorlinks=true,citecolor={rgb,1:red,0.21;green,0.49;blue,0.74},linkcolor={rgb,1:red,0.21;green,0.49;blue,0.74},urlcolor={rgb,1:red,0.21;green,0.49;blue,0.74}]{hyperref}
\usepackage{url}
\usepackage{booktabs}
\usepackage{amsfonts}
\usepackage{amssymb}
\usepackage{nicefrac}
\usepackage{microtype}
\usepackage[table]{xcolor}

\usepackage{amsmath}
\usepackage{graphicx}
\usepackage{tikz}
\usepackage{marvosym}
\usepackage{multirow}
\usepackage{wrapfig}

\newcommand{\eg}{\emph{e.g.}}

\definecolor{cUniAD}{HTML}{6B7280}       
\definecolor{cDiffDrive}{HTML}{374151}    
\definecolor{cRAP}{HTML}{9CA3AF}          
\definecolor{cAutoVLA}{HTML}{F59E0B}      
\definecolor{cReCogDrive}{HTML}{D97706}   
\definecolor{cAlpamayo}{HTML}{DC2626}     
\definecolor{cMindVLA}{HTML}{2563EB}      

\makeatletter
\def\@listI{\leftmargin\leftmargini \parsep 1pt \topsep 2pt \itemsep 1pt}
\let\@listi\@listI
\@listi
\makeatother
\selectfont

\title{MindVLA-U1: VLA Beats VA with Unified Streaming Architecture for Autonomous Driving}
\vspace{-0.5cm}

\author{%
  \parbox{\textwidth}{\centering
    Yuzhou Huang$^{1,2*}$\quad
    Benjin Zhu$^{2,3*\text{\Letter}\bigstar}$\quad
    Hengtong Lu$^{2}$\quad
    Victor Shea-Jay Huang$^{1,2}$\\
    Haiming Zhang$^{2}$\quad
    Wei Chen$^{2}$\quad
    Jifeng Dai$^{3}$\quad
    Yan Xie$^{2}$\quad
    Hongsheng Li$^{1\text{\Letter}}$\\[0.3em]
    {\mdseries\small $^{1}$CUHK MMLab\quad
    $^{2}$Li Auto\quad
    $^{3}$Tsinghua University\\[0.2em]
    \ttfamily
    1155253472@link.cuhk.edu.hk,\quad
    zhubenjin@lixiang.com\\[0.2em]
    \normalfont\small
    Project page:\, \url{https://mind-omni.github.io/}}
  }
}

\begin{document}

\maketitle
\vspace{-0.5cm}

\begin{abstract}
Autonomous driving has progressed from modular pipelines toward end-to-end unification, and Vision-Language-Action (VLA) models 
are a natural extension of this journey beyond Vision-to-Action (VA). 
In practice, driving VLAs have often trailed VA on planning quality, suggesting that the difficulty is not simply model scale but the interface through which semantic reasoning, temporal context, and continuous control are combined. We argue that this gap reflects how VLA has been built --- as isolated subtask improvements that fail to compose into coherent driving capabilities --- rather than what VLA is.
We present \textbf{MindVLA-U1}, the first unified streaming VLA architecture for autonomous driving. A unified VLM backbone produces autoregressive (AR) language tokens (optional) and flow-matching (diffusion-style) continuous action trajectories in a single forward pass over one shared representation, \emph{preserving the natural output form of each modality}. 
A full streaming design processes the driving video framewise rather than as fixed video-action chunks under costly temporal VLM modeling. Planned trajectories evolve smoothly across frames while a learned streaming memory channel carries temporal context and updates. 
The unified architecture enables fast/slow systems on dense/sparse \emph{Mixture-of-Transformers} (MoT) backbones via flexible self-attention context management, and exposes a measurable language-control path for action: a language-predicted driving intent steers the action diffusion via classifier-free guidance (CFG), turning language-side intent into control signals for continuous action planning. 
On the long-tail WOD-E2E benchmark, MindVLA-U1 \emph{surpasses experienced human drivers for the first time} ($8.20$ RFS vs.\ $8.13$ GT RFS) with only $2$ diffusion steps, achieves state-of-the-art planning ADEs over prior VA/VLA methods by large margins, and matches VA-class latency ($\sim$$16$ FPS vs.\ RAP's $\sim$$18$ FPS at matched $\sim$1B scale) while preserving natural language interfaces for human--vehicle interaction.
\end{abstract}

\section{Introduction}
\label{sec:intro}

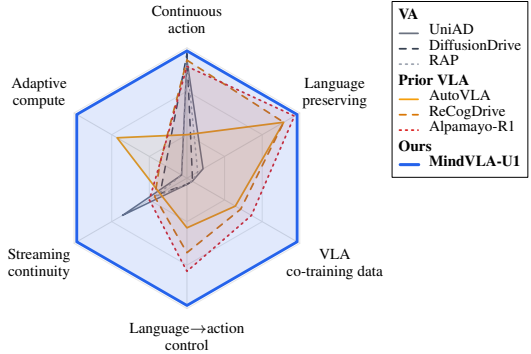
\begin{wrapfigure}[13]{r}{0.5\linewidth}
  \vspace{-1.2em}
  \centering
  \resizebox{\linewidth}{!}{%
    \begin{tikzpicture}[font=\scriptsize, line cap=round, line join=round]
      \foreach \r in {0.7,1.4,2.1} {
        \draw[gray!20] (90:\r) -- (30:\r) -- (-30:\r) -- (-90:\r) -- (-150:\r) -- (150:\r) -- cycle;
      }
      \foreach \a in {90,30,-30,-90,-150,150} {
        \draw[gray!25] (0,0) -- (\a:2.1);
      }

      \node[align=center] at (90:2.55) {Continuous\\action};
      \node[align=center] at (30:2.75) {Language\\preserving};
      \node[align=center] at (-30:2.75) {VLA \\ co-training data};
      \node[align=center] at (-90:2.55) {Language$\to$action \\ control};
      \node[align=center] at (-150:2.75) {Streaming\\continuity};
      \node[align=center] at (150:2.75) {Adaptive\\compute};

      \path[fill=cUniAD, fill opacity=0.06, draw=cUniAD, line width=0.7pt]
        (90:1.90) -- (30:0.30) -- (-30:0.10) -- (-90:0.10) -- (-150:1.20) -- (150:0.10) -- cycle;
      \path[fill=cDiffDrive, fill opacity=0.05, draw=cDiffDrive, line width=0.7pt, dashed]
        (90:2.00) -- (30:0.10) -- (-30:0.10) -- (-90:0.10) -- (-150:0.50) -- (150:0.40) -- cycle;
      \path[fill=cRAP, fill opacity=0.05, draw=cRAP, line width=0.7pt, dotted]
        (90:1.90) -- (30:0.20) -- (-30:0.10) -- (-90:0.10) -- (-150:0.60) -- (150:0.10) -- cycle;
      \path[fill=cAutoVLA, fill opacity=0.10, draw=cAutoVLA, line width=0.8pt]
        (90:0.70) -- (30:1.80) -- (-30:0.90) -- (-90:0.80) -- (-150:0.50) -- (150:1.30) -- cycle;
      \path[fill=cReCogDrive, fill opacity=0.08, draw=cReCogDrive, line width=0.8pt, dashed]
        (90:1.90) -- (30:1.80) -- (-30:1.00) -- (-90:1.20) -- (-150:0.60) -- (150:0.50) -- cycle;
      \path[fill=cAlpamayo, fill opacity=0.08, draw=cAlpamayo, line width=0.8pt, dotted]
        (90:1.80) -- (30:2.00) -- (-30:1.20) -- (-90:1.50) -- (-150:0.70) -- (150:0.50) -- cycle;
      \path[fill=cMindVLA, fill opacity=0.14, draw=cMindVLA, line width=1.3pt]
        (90:2.05) -- (30:2.05) -- (-30:2.05) -- (-90:2.05) -- (-150:2.05) -- (150:2.05) -- cycle;

      \node[anchor=north west, draw=black, line width=0.3pt, inner sep=3pt, fill=white,
            font=\scriptsize] at (3.30, 2.85) {%
        \begin{tabular}{@{}l@{\hspace{4pt}}l@{}}
        \multicolumn{2}{@{}l@{}}{\textbf{VA}} \\
        \tikz\draw[cUniAD,    line width=0.8pt]            (0,0) -- (0.30,0); & UniAD \\
        \tikz\draw[cDiffDrive,line width=0.8pt, dashed]    (0,0) -- (0.30,0); & DiffusionDrive \\
        \tikz\draw[cRAP,      line width=0.8pt, dotted]    (0,0) -- (0.30,0); & RAP \\[2pt]
        \multicolumn{2}{@{}l@{}}{\textbf{Prior VLA}} \\
        \tikz\draw[cAutoVLA,    line width=0.8pt]          (0,0) -- (0.30,0); & AutoVLA \\
        \tikz\draw[cReCogDrive, line width=0.8pt, dashed]  (0,0) -- (0.30,0); & ReCogDrive \\
        \tikz\draw[cAlpamayo,   line width=0.8pt, dotted]  (0,0) -- (0.30,0); & Alpamayo-R1 \\[2pt]
        \multicolumn{2}{@{}l@{}}{\textbf{Ours}} \\
        \tikz\draw[cMindVLA, line width=1.4pt]             (0,0) -- (0.30,0); & \textbf{MindVLA-U1} \\
        \end{tabular}};
    \end{tikzpicture}%
  }
  \caption{\textbf{AD capability radar}}
  \label{fig:comparison_matrix}
\end{wrapfigure}

Driving, at its core, is two things at once --- a continuous act of physical control, and a continuous act of understanding. Most of it happens by reflex: the routine lane changes, the gentle braking, the thousand small adjustments that a skilled driver makes without thinking. But the moments that separate competent driving from merely adequate driving are the moments when reflex is not enough --- when something in the world demands interpretation, and the correct response depends on knowing what that something means. A system that drives well must do both, and must do them within a single coherent architecture.

Vision-to-Action (VA) models~\cite{hu2023planning,jiang2023vad,chen2024vadv2,sun2025sparsedrive,zheng2024genad,Weng_2024_CVPR,li2024ego,chitta2022transfuser,liao2025diffusiondrive,zheng2025diffusion,feng2025rap} represent the state of the art on the first half of this problem. They map sensor input to trajectories with centimeter-level precision, dominate planning benchmarks, and produce genuinely useful driving within the distribution of scenarios they have seen.
But the semantic structure behind their behavior remains implicit, encoded in latent features tied to the co-occurrence patterns of the training distribution. Such representations can support strong pattern-conditioned action, but they are not explicit, inspectable, or reliably compositional when appearance, category, intent, and required response no longer compose in familiar ways. Scaling can improve the mapping from pixels to waypoints, but it does not by itself turn that mapping into an explicit, compositional representation of the world.
Vision-Language-Action (VLA) models~\cite{shao2024lmdrive,tian2024drivevlm,jiang2024senna,sima2024drivelm,hwang2024emma,wang2025omnidrive,chi2025impromptu,zhou2025autovla,zeng2025futuresightdrive,yuan2025autodrive,rowe2025poutine,fu2025orion,renz2025simlingo,zhang2025adadrive,luo2025adathinkdrive,li2025recogdrive,li2025drivevla,wang2025alpamayo,peng2025counterfactual,huang2026automot} are a natural architecture for systems that must do both. A Vision-Language Model (VLM) backbone~\cite{chen2024internvl,qwen2.5,yang2025qwen3,qwen3.5,guo2025deepseek} provides the explicit representations VA lacks: compositional, open-vocabulary understanding grounded in language. In principle, this is the missing half of the driving problem. In practice, driving VLAs have often trailed VA on planning quality, which has encouraged the view that language and planning precision are in tension, and that adding understanding costs control.

We argue that this view confuses a design failure with a paradigm limitation. Prior driving VLAs share a consistent set of interface-level shortcomings the framework itself does not impose.
First, the action interface is mismatched to control precision~\cite{sima2024drivelm,hwang2024emma,wang2025omnidrive,chi2025impromptu,zhou2025autovla,zeng2025futuresightdrive,yuan2025autodrive,rowe2025poutine}. Some designs decode trajectories as discrete numerical tokens through the language head: VLMs can describe rough positions, not high-precision coordinates, so the language stream imposes a precision floor below what physical control demands and exposes action to language-style hallucination.
Others run separate action experts~\cite{black2024pi_0,intelligence2025pi_} that read VLM features through cross-attention~\cite{tian2024drivevlm,jiang2024senna,li2025recogdrive}: action tokens never enter the VLM's self-attention, so the VLM representation is built without action context. In driving, where centimeter-level control depends on tight VLM-action coupling, this routing reduces the VLM to a feature encoder feeding a VA-style head and forfeits much of the VLM's representational potential.
Second, temporal modeling is short-window or chunk-scoped~\cite{zhou2025autovla,yuan2025autodrive,rowe2025poutine,renz2025simlingo,luo2025adathinkdrive,li2025recogdrive,wang2025alpamayo,peng2025counterfactual,huang2026automot}: because the planner predicts fixed-length \emph{action chunks}, the VLM is asked to absorb temporal context directly from multi-frame inputs, but multi-view driving video carries heavy token redundancy, efficient temporal VLM modeling remains an open problem~\cite{nguyen2026video}, and chunked output creates discontinuities at chunk boundaries.
Third, language has no explicit path into action~\cite{shao2024lmdrive,zeng2025futuresightdrive,fu2025orion,zhang2025adadrive,li2025drivevla}. Most driving VLAs initialize from a pretrained VLM but feed only templated driving commands at inference, collapsing the language capability into a fixed prefix and effectively running the system as VA with VLM weights. The root cause is the absence of driving-specific language--action data: real driving logs lack language supervision, and generic VQA data lacks planning signal.
Reported planning gains with language enabled are therefore asserted from the existence of a language head, not demonstrated through a measured language-to-action route. These are interface failures, not paradigm failures, and together they leave the central VLA question open: how to combine semantic understanding, temporal context, and continuous control inside one model without sacrificing precision.

So what makes for a good VLA system for autonomous driving? To address driving as a whole, it should derive its interface from the task rather than from inherited VLM conventions.
Driving requires centimeter-level physical control, so actions should remain continuous. The long tail requires open-vocabulary and compositional semantic representations, so the model needs an explicit language-grounded representation rather than only implicit visual features. VLA's promise depends on language knowledge becoming usable by action, so the language pathway must have a measurable route into continuous planning rather than merely sharing a backbone.
Driving arrives as a framewise stream, not as a sequence of fixed-length chunks, and the temporal interface must reflect this efficiently rather than forcing the VLM to attend over redundant multi-view fixed-frame videos at constant intervals.
Finally, routine and difficult moments impose different compute demands. A common criticism of VLA is that adding a VLM backbone fundamentally increases planning cost by a large margin; we view this as another interface-level misinterpretation --- a single architecture should admit both fast and slow modes depending on what the scene requires (Figure~\ref{fig:comparison_matrix}).
Many of these requirements have never been addressed in previous VA/VLAs, and jointly satisfying all of them inside a single VLA is structurally difficult --- they pull in different architectural directions --- so the capabilities that emerge from joint satisfaction have not been demonstrated by any prior driving VLA: smooth long-horizon planning, controllable language-driven action generation, and fast/slow execution within one model without losing planning quality while preserving the language interface. 

We present \textbf{MindVLA-U1}, the first unified streaming VLA architecture to address them all.
The core of MindVLA-U1 is a \emph{unified shared backbone for scene understanding and action generation}: optional autoregressive (AR) language tokens and flow-matching~\cite{ho2020denoising,esser2024scaling,liu2022flow} (diffusion-style) continuous action trajectories are produced from one shared representation in a single forward pass, preserving the natural output form of each modality --- action remains continuous, language remains explicit, and both modalities share the same weights.
The backbone runs under a full \emph{streaming paradigm}: each step processes only the current multi-view frame rather than fixed video-action chunks under costly temporal VLM modeling, while a learned streaming memory channel propagates compact visual memory across frames and updates along whole streams, so planned trajectories evolve smoothly framewise without the redundant multi-frame attention that prior VLAs rely on.
Through flexible self-attention context management, the unified architecture admits multiple inference modes (language-then-action, action-then-language, action-only) that give the dense VLA backbone fast/slow execution paths from a single checkpoint, and extends naturally to sparse \emph{Mixture-of-Transformers} (MoT)~\cite{vaswani2017attention,liang2024mixture,deng2025emerging} variants for higher action efficiency and extensibility.
Finally, for the first time in VLA, MindVLA-U1 exposes a measurable language-to-action bridge: it autoregressively predicts a driving intent token and feeds it as classifier-free guidance~\cite{ho2022classifier} into the action diffusion, turning language-side intent into a control signal for continuous trajectory generation rather than a capability merely asserted by the presence of VLM pretrained weights.
Contributions are summarized as follows:
\begin{enumerate}
  \item \textbf{The first unified shared VLA architecture for autonomous driving.} A single VLM backbone jointly performs scene understanding and continuous action generation, preserving action precision and language capability with explicit language-to-action controllability and flexible fast/slow execution on dense / sparse MoT backbones.

  \item \textbf{Streaming paradigm with efficient temporal modeling.} A framewise streaming paradigm with efficient memory propagation that supports long-horizon prediction, eliminates action chunking, and provides a computation-efficient temporal modeling scheme for VLA.

  \item \textbf{State-of-the-art performance on WOD-E2E}~\cite{xu2025wod}\textbf{.} MindVLA-U1 surpasses experienced human drivers for the first time ($8.20$ RFS vs.\ $8.13$ GT RFS) with only 2 diffusion steps, achieving state-of-the-art planning ADEs over all prior VA/VLA methods by large margins while preserving a natural-language interface for human--vehicle interaction.
\end{enumerate}

\section{Method}
\label{sec:method}

\begin{figure*}[htbp]
  \centering
  \includegraphics[width=0.95\linewidth]{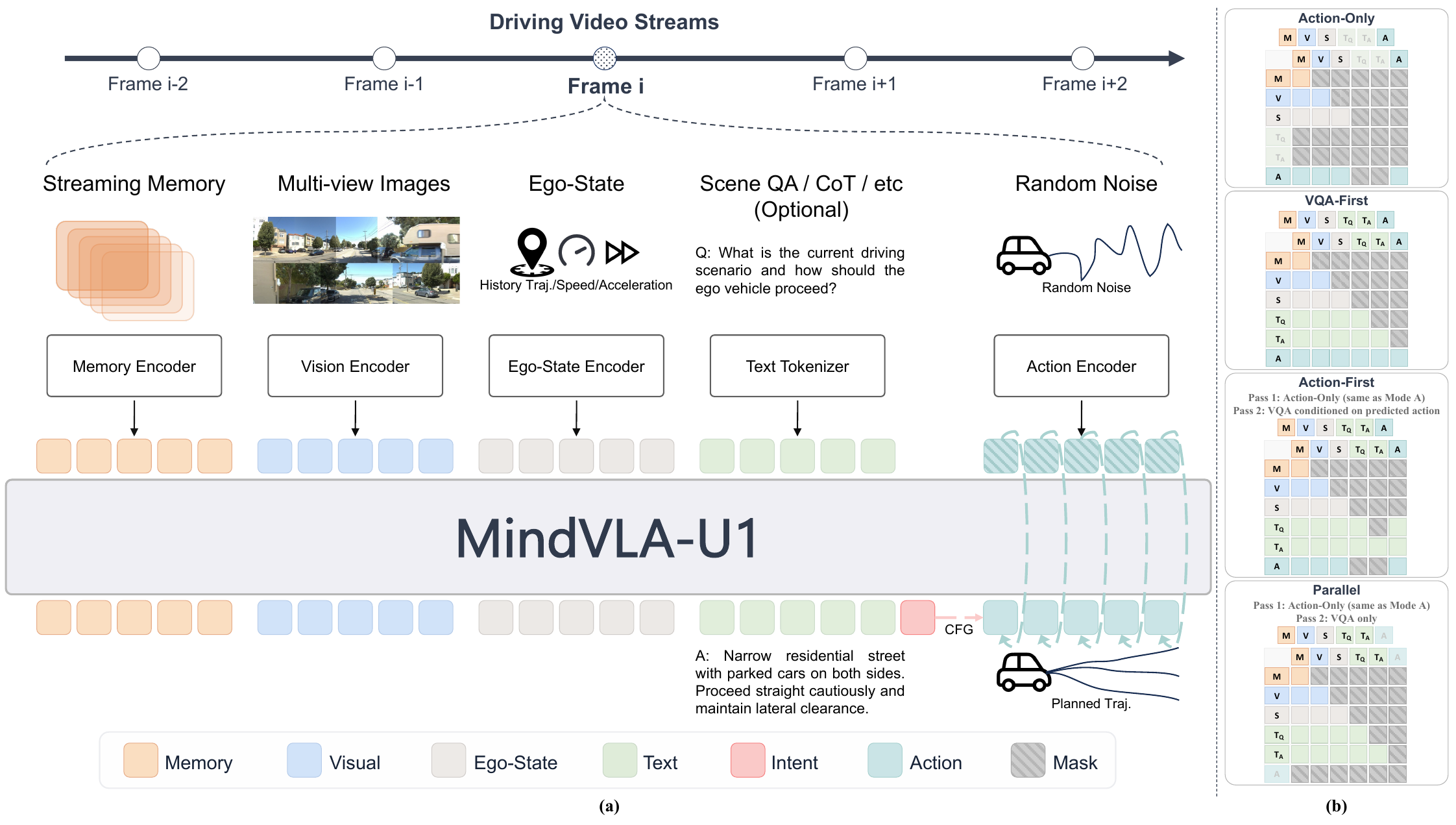}
  \caption{\textbf{Overview of MindVLA-U1.} Vision, ego-state, language, memory, and noisy action tokens flow through a shared VLM backbone in one forward pass; the LM head and the flow-matching action head read out at their respective token positions (\S\ref{sec:unified}). A FIFO memory channel propagates compact temporal context across frames, motion-aligned on read and refreshed after each pass (\S\ref{sec:streaming_memory}). Attention-mask composition exposes four inference orderings (\texttt{vqa\_first/only}, \texttt{action\_first/only}) for fast/slow execution from the same weights.}
  \label{fig:framework}
\end{figure*}

MindVLA-U1 combines a \emph{unified shared backbone} (\S\ref{sec:unified}) that jointly performs scene understanding and continuous action generation in a single forward pass over one representation, with a \emph{streaming memory} (\S\ref{sec:streaming_memory}) paradigm that propagates a compact memory channel across frames in whole driving sessions rather than re-attending to multi-frame video chunks (Figure~\ref{fig:framework}). To let language guide continuous action measurably and to match compute to scene demand, we further develop a language-to-action route and a fast/slow execution scheme that runs on dense or sparse MoT variants of this shared backbone.

\subsection{Unified Shared Backbone}
\label{sec:unified}

MindVLA-U1 keeps language autoregressive and action continuous: a single VLM backbone jointly processes vision, ego-state, language, memory, and noisy action tokens through the same self-attention and FFN weights at every layer. 
Noisy action tokens flow through every transformer layer~\cite{vaswani2017attention} of the backbone exactly like vision and language tokens with no separate computation path. Gradients from both training objectives flow through every layer back to all token modalities; A language LM head decodes language tokens and a thin flow-matching MLP reads out the action velocity field at action positions, both in a single forward pass (Figure~\ref{fig:framework}).

\paragraph{Joint AR + Flow Matching on Shared VLM backbones.}
For a single frame-step with multi-view vision tokens $\mathbf{I}$, ego-state history $\mathbf{e}$, language query $\mathbf{q}$, language answer tokens $\mathbf{a}=(a_1,\ldots,a_L)$, 
and a length-$N$ sequence of noisy action tokens $\mathbf{x}_t=(x_{t,1},\ldots,x_{t,N})$, the shared backbone $f_\theta$ runs all tokens through every transformer layer in a single forward pass. 
The prefix $(\mathbf{I},\mathbf{e},\mathbf{q},\mathbf{a})$ is causally masked and the action tokens are bidirectionally fully visible. Taking language-then-action path as example, the LM head at language position $l-1$ sees only $(\mathbf{I},\mathbf{e},\mathbf{q},a_{<l})$, while the action MLP at any action position sees the full prefix $(\mathbf{I},\mathbf{e},\mathbf{q},\mathbf{a})$ and all $N$ noisy action tokens. 
The LM head produces the autoregressive loss over answer tokens (\emph{e.g.}, scene QAs):
\begin{equation}
  \mathcal{L}_{\mathrm{AR}}
  = -\sum_{l=1}^{L} \log p_\theta\!\left(a_l \mid \mathbf{I},\mathbf{e},\mathbf{q},a_{<l}\right).
  \label{eq:ar_loss}
\end{equation}
The action MLP at action positions predicts a velocity field for flow matching. Given ground-truth waypoints $\mathbf{x}_0 = \mathbf{w}$ and noise $\boldsymbol{\epsilon} \sim \mathcal{N}(0, I)$, the noisy action tokens take the form $\mathbf{x}_t = t\boldsymbol{\epsilon} + (1-t)\mathbf{x}_0$ at $t \sim \mathrm{Beta}(1.5, 1.0)$ under the convention $t{=}0$ is data and $t{=}1$ is noise~\cite{ho2020denoising,esser2024scaling,liu2022flow}, giving
\begin{equation}
  \mathcal{L}_{\mathrm{FM}}
  = \bigl\|\,v_\theta(\mathbf{x}_t, t;\, \mathbf{I},\mathbf{e},\mathbf{q},\mathbf{a}) - (\boldsymbol{\epsilon} - \mathbf{x}_0)\,\bigr\|^2,
  \label{eq:fm_loss}
\end{equation}
where $v_\theta$ is the action MLP's velocity prediction. The training objective $  \mathcal{L} = \mathcal{L}_{\mathrm{AR}} + \mathcal{L}_{\mathrm{FM}}$, 
and at inference trajectories are recovered by ODE integration~\cite{song2020score} over $v_\theta$.

Training the discrete AR cross-entropy and the continuous flow-matching MSE on shared backbone parameters is stable in practice because the two losses apply at \emph{different output positions} through \emph{separate readout heads}: they share gradients but not targets, and the scene-grounded features each loss needs compose rather than compete. 
This shared-backbone design separates MindVLA-U1 from three prior VLA patterns: \emph{discrete-token action VLAs} (full-autoregressive~\cite{zhou2025autovla} or block-diffusion-style~\cite{ma2025dvlm}) decode trajectories as discrete tokens through the language head and inherit a tokenizer-imposed precision floor; \emph{discrete tokens with downstream trajectory decoding} (\eg, Alpamayo-R1~\cite{wang2025alpamayo}) recovers continuous action from those discrete tokens via a separate decoder, leaving action generation outside the backbone forward pass; 
and \emph{VLM with separate action experts} (\eg, pi-0~\cite{black2024pi_0}) routes action through a cross-attention expert with its own attention/FFN, so VLM features are built without action context. 
Action of our MindVLA-U1 is decoded continuously inside the shared self-attention.

\paragraph{Language-to-Action Bridge via Intent-CFG.}
\label{sec:semantic_intent}
Intent is intuitively a natural and compact bridge from language reasoning to continuous action. No matter how complex the situation, after sufficient scene analysis and reasoning a human driver settles on a single driving intent (go straight, change lane, yield) and acts on it. 
MindVLA-U1 implements this bridge inside the VLM forward pass: the language head is supervised to predict the current-scene intent label $z$ (\emph{e.g.}, Left, Right, Go Straight) via standard next-token prediction, and the predicted intent token is embedded and added to the action MLP's time embedding before velocity prediction. 
During training, the conditioning intent is occasionally replaced by an unconditional token $\emptyset$ (CFG dropout)~\cite{ho2022classifier}, so the same action tokens learn both conditional and unconditional velocity fields. At inference, MindVLA-U1 decodes an intent token through the language head and runs two backbone forward passes --- conditional on $z$ and on $\emptyset$ --- mixing the velocity fields by guidance scale $s$:
\begin{equation}
  v_{\mathrm{cfg}}(\mathbf{x}_t, t)
  = v_\theta(\mathbf{x}_t, t;\, \emptyset)
  + s\left(v_\theta(\mathbf{x}_t, t;\, z) - v_\theta(\mathbf{x}_t, t;\, \emptyset)\right),
  \label{eq:intent_cfg}
\end{equation}
where $v_\theta(\,\cdot\,;\, z)$ extends the velocity field of Eq.~\ref{eq:fm_loss} with intent conditioning $z$ (the prefix $\mathbf{I},\mathbf{e},\mathbf{q},\mathbf{a}$ is suppressed). 
This language-to-action controllability inherits from MindVLA-U1's end-to-end, unified shared VLA architecture design: discrete-token action heads have no continuous velocity field for CFG to mix.
Post-hoc trajectory decoders or separate action experts detach the gradients between intent prediction and action denoising that U1 maintains.

\begin{figure}[!t]
\begin{minipage}[t]{0.48\linewidth}
  \centering
  \resizebox{\linewidth}{!}{%
  \begin{tikzpicture}[
    font=\scriptsize, >=latex,
    on/.style={draw, rounded corners=0.5pt, minimum size=4mm, inner sep=0pt},
    off/.style={draw=gray!55, dashed, rounded corners=0.5pt, minimum size=4mm, inner sep=0pt, text=gray!60},
    ctxblk/.style={draw=blue!65, fill=blue!8, rounded corners=2pt, inner sep=2pt, align=center},
    actblk/.style={draw=orange!85!black, fill=orange!12, rounded corners=2pt, inner sep=2pt, align=center},
    futureblk/.style={draw=gray!55, dashed, fill=gray!8, rounded corners=2pt, inner sep=2pt, align=center, text=gray!60},
    shareblk/.style={draw=black!55, fill=black!6, rounded corners=2pt, inner sep=2pt, align=center, font=\scriptsize\itshape},
  ]
    \node[on, fill=blue!22]   (V) at (0.45, 3.30) {V};
    \node[on, fill=blue!22]   (L) at (0.95, 3.30) {L};
    \node[off]                (W) at (1.45, 3.30) {W};
    \node[off]                (P) at (1.95, 3.30) {P};
    \node[on, fill=orange!28] (M) at (3.30, 3.30) {M};
    \node[on, fill=orange!28] (S) at (3.80, 3.30) {S};
    \node[on, fill=orange!28] (A) at (4.30, 3.30) {A};
    \node[off]                (C) at (4.80, 3.30) {C};
    \node[font=\tiny, color=blue!75]          at (1.20, 3.70) {context tokens};
    \node[font=\tiny, color=orange!90!black]  at (4.05, 3.70) {action tokens};

    \node[ctxblk, minimum width=2.4cm, minimum height=0.55cm] (qkvoctx) at (1.20, 2.40)
      {Q/K/V/O$_{\rm ctx}$\\[-1pt]\tiny perception modalities};
    \node[actblk, minimum width=2.4cm, minimum height=0.55cm] (qkvoact) at (4.05, 2.40)
      {Q/K/V/O$_{\rm act}$ (narrow)\\[-1pt]\tiny motor / proprio modalities};
    \node[font=\tiny, color=black!70, anchor=west] at (5.45, 2.40) {encoder};

    \node[shareblk, minimum width=5.3cm, minimum height=0.40cm] (sa) at (2.625, 1.50)
      {Shared self-attention --- universal K/V pool};

    \node[ctxblk, minimum width=1.15cm, minimum height=0.55cm] (ffnctx)    at (0.70, 0.60)
      {FFN$_{\rm ctx}$\\[-1pt]\tiny context};
    \node[actblk, minimum width=1.15cm, minimum height=0.55cm] (ffnact)    at (1.95, 0.60)
      {FFN$_{\rm act}$\\[-1pt]\tiny action};
    \node[futureblk, minimum width=1.15cm, minimum height=0.55cm] (ffnreas) at (3.30, 0.60)
      {FFN$_{\rm reason}$\\[-1pt]\tiny reasoning};
    \node[futureblk, minimum width=1.15cm, minimum height=0.55cm] (ffnsafe) at (4.65, 0.60)
      {FFN$_{\rm safety}$\\[-1pt]\tiny safety};
    \node[font=\tiny, color=black!70, anchor=west] at (5.45, 0.60) {decoder};

    \node[font=\tiny, color=black!70] at (2.625, 0.05) {one MoT layer (stacked $\times N$)};

    \draw[->]                  (V.south) -- (V.south|-qkvoctx.north);
    \draw[->]                  (L.south) -- (L.south|-qkvoctx.north);
    \draw[->, gray!65, dashed] (W.south) -- (W.south|-qkvoctx.north);
    \draw[->, gray!65, dashed] (P.south) -- (P.south|-qkvoctx.north);
    \draw[->]                  (M.south) -- (M.south|-qkvoact.north);
    \draw[->]                  (S.south) -- (S.south|-qkvoact.north);
    \draw[->]                  (A.south) -- (A.south|-qkvoact.north);
    \draw[->, gray!65, dashed] (C.south) -- (C.south|-qkvoact.north);

    \draw[->] (qkvoctx.south) -- (qkvoctx.south|-sa.north);
    \draw[->] (qkvoact.south) -- (qkvoact.south|-sa.north);

    \draw[->]                  (ffnctx.north|-sa.south)  -- (ffnctx.north);
    \draw[->]                  (ffnact.north|-sa.south)  -- (ffnact.north);
    \draw[->, gray!65, dashed] (ffnreas.north|-sa.south) -- (ffnreas.north);
    \draw[->, gray!65, dashed] (ffnsafe.north|-sa.south) -- (ffnsafe.north);
  \end{tikzpicture}}
  \caption{\textbf{Fast/Slow systems on Sparse MoT.} Each layer splits into two parallel expert groups --- context (V, L) and action (M, S, A) --- joined by a shared self-attention pool so every query sees both groups. Per-modality Q/K/V/O projections feed the shared SA; per-functionality FFN experts (ctx, act, plus extension slots: reason, safety) decode after it. Fast mode (\texttt{action\_only}) skips language decoding (\S\ref{sec:supp_mot_arch}).}
  \label{fig:mot}
\end{minipage}\hfill
\begin{minipage}[t]{0.48\linewidth}
  \centering
  \includegraphics[width=\linewidth]{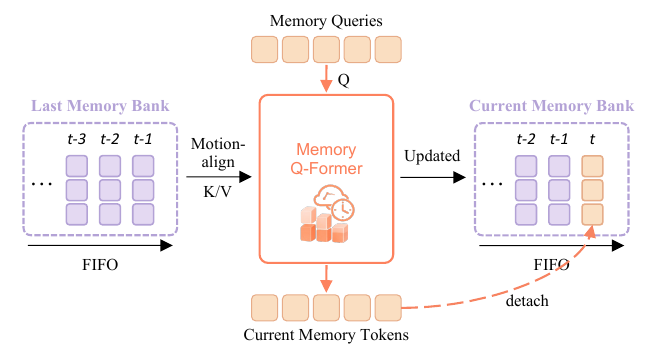}
  \caption{\textbf{Streaming Memory.} At frame $i$, a bounded FIFO channel of compressed memory tokens $\mathbf{m}_i$ from prior frames is read into the backbone (motion-aligned to the current ego pose); a Q-Former-style propagation transformer writes a fresh memory token set after the forward pass and evicts the oldest. Gradients flow across the channel, so loss at frame $i$ supervises memory written at frame $i{-}1$.}
  \label{fig:memory_bank}
\end{minipage}
\end{figure}

\paragraph{Fast/Slow Systems.}
MindVLA-U1's unified shared backbone enables fast/slow systems easily as self-attention context management --- no extra modules, gating heads, or runtime selectors are required. 
Four orderings cover the design space via attention-mask composition alone (Figure~\ref{fig:framework}): 
\texttt{vqa\_first/only} decode language first and feeds the answer back as conditioning for action denoising, while \texttt{action\_first/only} denoises action first and conditions language on the predicted trajectory. 
These modes serve complementary purposes: \texttt{vqa\_first} enhances action diffusion --- scene QA enriches the action-side context with structured semantic features, and CoT-style~\cite{wei2022chain} reasoning can run before the trajectory is committed --- while \texttt{action\_first} produces a natural-language commentary on the just-predicted trajectory, useful in safety-critical scenarios where users need to inspect why a particular maneuver was chosen. 
The fast path \texttt{action\_first} drops language entirely, giving a strictly shorter forward pass (VA systems). All orderings are sampled at training to enable flexible inference mode during deployment under one set of weights, which empirically preserves planning quality across modes showing that the VLM backbone does not impose a cost on action even when language is present. The same attention-mask mechanism extends naturally to additional functional experts (reasoning, safety, world model) beyond the language and action heads instantiated here.

The same backbone admits a sparse \emph{Mixture-of-Transformers (MoT)} variant~\cite{liang2024mixture,deng2025emerging} for higher action efficiency and extensibility (Figure~\ref{fig:mot}). The MoT splits each layer into two parallel expert groups --- a context group serving visual and language tokens, and an action group serving memory, ego-state, and action tokens --- joined by a shared self-attention pool so every query attends over both groups' representations in one forward pass. The two-group structure scaffolds future extensions: new perceptual modalities or cognitive functions enter as additional context-side token-roles, and behavior-style or control heads as additional action-side token-roles. 
We develop this into a foundation-architecture vision applicable beyond driving in \S\ref{sec:mot_foundation}; implementation details for the deployed two-group instance are in \S\ref{sec:supp_mot_arch}.

\vspace{-0.2cm}
\subsection{Streaming Paradigm}
\label{sec:streaming_memory}

MindVLA-U1 processes driving as a framewise stream: at frame $i$, the VLM consumes only the current multi-view frame --- which can arrive at flexible, even non-uniform intervals --- plus a compact memory feature $\mathbf{m}_i$ read from a bounded First-In-First-Out (FIFO) memory channel that stores per-frame memory tokens (each a compressed summary of a past frame's backbone state, not raw visual tokens; Figure~\ref{fig:memory_bank}). Historical context propagates through the FIFO channel rather than via repeated multi-frame attention, so planning evolves smoothly across long driving sessions at bounded per-frame cost, without fixed action chunks or redundant temporal-VLM input. Because each token set was written under an earlier ego pose, it is motion-aligned to the current ego frame on read, keeping memory spatially consistent with the current viewpoint as the vehicle moves. The streaming schedule (frames processed in order with channel state carried forward) and the FIFO channel itself are separable components.

\vspace{-0.2cm}
\paragraph{Streaming memory updating.}
After the frame-$i$ forward pass, a Q-Former-style~\cite{li2023blip} propagation transformer cross-attends learnable query tokens to the backbone outputs to compute a fresh set of compressed memory tokens, which is appended to the FIFO channel; the oldest set is evicted once the token budget is exceeded, and the first-frame forward simply receives an empty channel. Gradients flow through the propagation transformer across frames, so the loss at frame $i$ supervises the memory written at frame $i{-}1$. This converts the channel from a passive cache into an actively supervised state: memory features are shaped by the same flow-matching and language objectives that supervise current-frame action and language, with no auxiliary reconstruction loss. The same read--forward--write loop runs at training and inference, so streaming behavior at deployment matches what was trained. Qualitative streaming examples --- per-frame inference smoothness and long-horizon multi-clip stitches --- are shown in \S\ref{sec:supp_streaming_qualitative}.

\section{Experimental Results}
\label{sec:experiments}

\subsection{Dataset, Benchmark, and Implementation Details}
\label{sec:setup}

\paragraph{Dataset, benchmark, and VLA co-training data.}
We train and evaluate on \emph{Waymo Open Dataset End-to-End (WOD-E2E)}~\cite{xu2025wod}: $4{,}021$ $\sim$20-second driving segments captured by an $8$-camera $360^\circ$ rig, covering long-tail scenarios (event frequency $<$0.003\%). 
The official split provides $2{,}037$ training and $479$ validation sequences; the test split releases the first $12$\,s of each segment and holds the final $8$\,s for evaluation. We provide a richer breakdown alongside the official WOD-E2E metrics. For language quality, we report BLEU-4~\cite{papineni2002bleu} and ROUGE-L~\cite{lin2004rouge} on the VQA answer head. For trajectory quality, we report: (i) the \emph{Rater Feedback Score (RFS)}, in which each predicted trajectory is matched to the closest of three human-annotated reference trajectories (scored in $[0,10]$), with trajectories inside a speed-scaled trust region receiving the rater's score and those outside decaying exponentially to a floor of $4$; (ii) RFS-matched ADE at $3$/$5$\,s, computed against the rater-matched reference; and (iii) RFS-GT ADE at $3$/$5$\,s, computed against the logged ground-truth trajectory --- the convention used by external methods and the official challenge leaderboard. For the long-horizon streaming setting that MindVLA-U1 targets, we additionally report sequence-level ADE at $3$/$5$/$10$/$15$/$20$/$25$\,s on the validation split. 

Beyond logged trajectories, real driving logs supply no language or trajectory-preference supervision; we close this gap with \emph{MindLabel}, an auto-labeling pipeline that, on the same training frames, generates scene-grounded VQA, intent-conditioned dreamed alternative trajectories, chain-of-thought rationales, and trajectory-evaluation QA. Across the full WOD-E2E benchmark, MindLabel annotates ${\sim}18.8$K clips, each independently labeled by both annotation backbones (Qwen3-VL and Qwen3.5-Plus), yielding ${\sim}3.8$M VQA pairs and ${\sim}250$K dreamed trajectories in aggregate (per-stage statistics in \S\ref{sec:supp_mindlabel}). For the results reported in this paper, we co-train only on the basic scene-grounded VQA and the official WOD-E2E ground-truth intent label alongside the flow-matching action loss in a single forward pass through the shared backbone (\S\ref{sec:unified}); harnessing the richer MindLabel outputs --- multiple dreamed trajectories per scene, CoT rationales, and trajectory-related QA --- is left to future work.

\paragraph{Network architecture and implementation.}
The MindVLA-U1 architecture is VLM-agnostic: the unified shared backbone, streaming memory, and Intent-CFG bridge are all defined over a generic VLM forward pass, so any modern VLM (\eg, InternVL, Qwen2.5/3/3.5-VL, DeepSeek-R1~\cite{chen2024internvl,qwen2.5,yang2025qwen3,qwen3.5,guo2025deepseek}) can serve as the backbone. We report main results on Qwen3-VL-2B; backbone-size results under Qwen3.5-VL are in \S\ref{sec:backbone_scaling}. The vision encoder is frozen; the visual merger, language model, ego-history encoders, the streaming memory module, and the lightweight flow-matching action head are jointly trained. The action head predicts position, velocity, and acceleration over a $5$-second horizon at $4$\,Hz with $2$ Euler integration steps at inference. End-to-end training in the sequence-streaming paradigm takes approximately $7$ hours on $8{\times}$H200 GPUs for $50$ epochs under BF16 with DeepSpeed ZeRO-2~\cite{rasley2020deepspeed}; the SE(2) pose-chain preprocessing that stitches per-clip local frames into a single global frame for sequence-level streaming training is described in \S\ref{sec:supp_pose_recovery}. Full architectural dimensions, optimizer hyperparameters, and the flow-matching component weights are deferred to \S\ref{sec:supp_setup}.


\subsection{Main Results}
\label{sec:main_results}

\begin{table*}[!htbp]
\caption{WOD-E2E \emph{validation split}. Methods grouped as VA / VLA / MindVLA-U1 (Ours), with the real-human-driver reference (logged GT trajectory scored under the same RFS metric) listed for comparison. \textbf{AD-PT}: external autonomous-driving pretraining ($\checkmark$ yes, $\times$ no, ? not reported). \emph{L B/R}: VQA BLEU-4 / ROUGE-L. \textbf{Bold} / \underline{underline} = best / second-best per column among VLA + Ours.}
\label{tab:waymo_e2e_val}
\footnotesize
\centering
\resizebox{\linewidth}{!}{%
\begin{tabular}{l|l|c|c|c|c|>{\columncolor[gray]{0.95}}c}
\specialrule{1pt}{0pt}{1pt}
\toprule
Method & Backbone & AD-PT & L B/R$\uparrow$ & RFS-GT ADE 3/5$\downarrow$ & RFS-matched ADE 3/5$\downarrow$ & RFS$\uparrow$ \\
\midrule
\multicolumn{7}{l}{\textbf{VA Methods}}\\
VAD~\cite{jiang2023vad}                                   & ImageNet vision encoder~\cite{dosovitskiy2020image}    & $\times$     & --                     & 3.19 / 5.81             & --                       & 4.45 \\
UniAD~\cite{hu2023planning}                               & ImageNet ViT~\cite{dosovitskiy2020image}               & $\times$     & --                     & 6.50 / 10.81            & --                       & 5.78 \\
RAP-DINO~\cite{feng2025rap}                              & DINOv3-H~\cite{simeoni2025dinov3}                   & $\checkmark$     & --                     & 0.97 / 2.20             & --                       & 7.91 \\
\midrule
\multicolumn{7}{l}{\textbf{VLA Methods (action-only at inference)}}\\
Poutine-Base~\cite{rowe2025poutine}                      & Qwen2.5-VL-3B~\cite{qwen2.5}              & $\checkmark$ & --                     & 1.27 / 2.94             & --                       & \underline{8.12} \\
\midrule
\multicolumn{7}{l}{\textbf{Real human driver (logged trajectory, RFS-scored)}}\\
Human Driver~\cite{rowe2025poutine}                          & ---                        & ---          & ---                    & ---                     & ---                      & 8.13 \\
\midrule
\multicolumn{7}{l}{\textbf{MindVLA-U1 (Ours)}}\\
MindVLA-U1                                       & Qwen3-VL-2B                & $\times$     & 0.30 / 0.49             & 0.92 / 2.14                      & 0.50 / 1.05                         & 7.83 \\
MindVLA-U1 + Intent-CFG                                   & Qwen3-VL-2B                & $\times$     & \textbf{0.31 / 0.52}    & \textbf{0.86} / \underline{2.13} & \textbf{0.47} / 1.07                & 7.92\footnotemark{} \\
MindVLA-U1 + MoT                                          & Qwen3-VL-2B                & $\times$     & \underline{0.30 / 0.51} & \underline{0.89} / \textbf{2.11} & \underline{0.49} / \underline{1.05} & 7.92 \\
MindVLA-U1 + RL                                           & Qwen3-VL-2B~\cite{yang2025qwen3}                & $\times$     & --                     & 1.01 / 2.28             & 0.51 / \textbf{1.03}     & \textbf{8.20} \\
\bottomrule
\specialrule{1pt}{1pt}{2pt}
\end{tabular}
}
\end{table*}
\footnotetext{Improved version reaches RFS \textbf{8.00} (RFS-GT ADE $0.86/2.06$, RFS-matched ADE $0.47/1.06$, L B/R $0.31/0.52$); we report the 7.92 checkpoint in the row above to keep all ablations (Tables~\ref{tab:language_helps_action},~\ref{tab:abl_mot}) on the same model.}

\begin{table*}[!htbp]
\caption{WOD-E2E \emph{official test split}. Conventions as in Table~\ref{tab:waymo_e2e_val}.}
\label{tab:waymo_e2e_test}
\footnotesize
\centering
\begin{tabular}{l|l|c|c|>{\columncolor[gray]{0.95}}c}
\specialrule{1pt}{0pt}{1pt}
\toprule
Method & Backbone & AD-PT & RFS-GT ADE 3/5$\downarrow$ & RFS$\uparrow$ \\
\midrule
\multicolumn{5}{l}{\textbf{VA Methods}}\\
ViT-Adapter-GRU                                   & n/r                        & ?            & 1.20 / 2.70                         & 7.50 \\
Swin-Trajectory                                   & Swin Transformer~\cite{liu2021swin}           & ?            & 1.21 / 2.81                         & 7.54 \\
DiffusionLTF                                      & DiffusionDrive~\cite{liao2025diffusiondrive}             & ?            & 1.36 / 2.98                         & 7.72 \\
UniPlan (DiffusionDrive)                          & DiffusionDrive~\cite{liao2025diffusiondrive}             & ?            & 1.31 / 2.99                         & 7.69 \\
\midrule
\multicolumn{5}{l}{\textbf{VLA Methods (action-only at inference)}}\\
Open-LLaMA                      & LLaMA-Vision~\cite{touvron2023llama}               & ?            & 1.31 / 3.22                         & 7.43 \\
NaiveEMMA                         & Gemini~\cite{team2023gemini}                     & $\times$     & 1.32 / 3.02                         & 7.53 \\
AutoVLA~\cite{zhou2025autovla}                           & Qwen2.5-VL-3B~\cite{qwen2.5}              & $\checkmark$ & 1.35 / 2.96                         & 7.56 \\
dVLM-AD~\cite{ma2025dvlm}                            & LLaDA-V~\cite{you2025llada} + SigLIP2~\cite{tschannen2025siglip}          & $\checkmark$ & 1.29 / 3.02                         & 7.63 \\
HMVLM~\cite{wang2025hmvlm}                               & Qwen2.5-VL-3B~\cite{qwen2.5}              & ?            & 1.33 / 3.07                         & 7.74 \\
\midrule
\multicolumn{5}{l}{\textbf{MindVLA-U1 (Ours, Language-preserving)}}\\
MindVLA-U1                                       & Qwen3-VL-2B                & $\times$     & \underline{1.16} / \underline{2.67} & \underline{7.77} \\
MindVLA-U1 + RL                                           & Qwen3-VL-2B~\cite{yang2025qwen3}                & $\times$     & \textbf{1.09 / 2.66}                & \textbf{7.87} \\
\bottomrule
\specialrule{1pt}{1pt}{2pt}
\end{tabular}
\end{table*}

On WOD-E2E, MindVLA-U1 closes the planning-quality gap on which driving VLAs have trailed VA. On the official test split, MindVLA-U1 + RL achieves the highest reported RFS ($\mathbf{7.87}$) and the lowest RFS-GT ADE ($\mathbf{1.09 / 2.66}$\,m), ahead of the closest previous VLA dVLM-AD ($1.29 / 3.02$). The architecture, not RL, produces this result: without RL or external AD-data pretraining, MindVLA-U1 + Intent-CFG reaches RFS $7.92$ on val, on par with RAP-DINO ($7.91$, the strongest pure-VA system), and MindVLA-U1 alone places second-best on test ($7.77$, $1.16 / 2.67$\,m), outperforming every previous VLA and VA method on ADE 3s/5s.

MindVLA-U1 reaches these results while preserving its natural-language interface at inference. Every previous VLA in our comparison discards or does not use the language head once trajectories are decoded (``--'' in the L B/R column of Table~\ref{tab:waymo_e2e_val}), effectively running as VA with VLM weights; MindVLA-U1 reports VQA BLEU-4/ROUGE-L on the same checkpoint that produces the trajectory, so language and planning precision coexist on a single set of weights. The remaining val$\to$test RFS gap is distributional, not architectural: the test split over-represents urgent-stopping and yielding scenes that the Waymo $3$-class GT intent ontology used at training (\emph{left} / \emph{right} / \emph{straight}) cannot supervise (intent statistics in \S\ref{sec:supp_mindlabel_dist_shift}). Reasoning over richer driving intents in complex scenes to guide action diffusion is left to future work.
Unless otherwise stated, ``MindVLA-U1'' refers to the dense configuration in Table~\ref{tab:waymo_e2e_val} (RFS $7.83$) in later ablations.

\subsection{Language-to-Action Controllability via Intent-CFG}
\label{sec:unified_capability_diagnostics}

A central claim of the unified shared backbone is that language-side state has a measurable causal route into continuous action. We isolate one capability of this route: predicting a driving intent from the shared representation and using it to steer action diffusion via CFG. Three conditioning-signal sources are compared against a no-intent baseline --- trajectory-derived, GT-supplied, and our next-token-predicted (NTP) variant; embedder implementation details are in \S\ref{sec:supp_intent_cfg}.

\begin{table*}[!htbp]
\caption{Intent-conditioned action diffusion on WOD-E2E val. Variant definitions are described in the surrounding text. $^\dagger$ Ours uses a refined NTP-predicted intent embedding; implementation in \S\ref{sec:supp_intent_cfg}.}
\label{tab:language_helps_action}
\footnotesize
\centering
\resizebox{\linewidth}{!}{%
\begin{tabular}{l|c|c|c|>{\columncolor[gray]{0.95}}c}
\specialrule{1pt}{0pt}{1pt}
\toprule
Variant & L B/R$\uparrow$ & RFS-GT ADE 3/5$\downarrow$ & RFS-matched ADE 3/5$\downarrow$ & RFS$\uparrow$ \\
\midrule
\multicolumn{5}{l}{\textbf{No-intent baseline}}\\
MindVLA-U1 (no intent)                & 0.30 / 0.49           & 0.92 / 2.14                      & 0.50 / \textbf{1.05}                & 7.83 \\
\midrule
\multicolumn{5}{l}{\textbf{Alternative Intent-Injection Mechanisms}}\\
+ Trajectory-derived Intent-CFG       & 0.30 / 0.49           & 0.88 / 2.14                      & 0.49 / 1.08                         & 7.81 \\
+ GT-supplied Intent-CFG              & 0.29 / 0.47           & 0.89 / 2.15                      & 0.50 / 1.06                         & 7.83 \\
\midrule
\multicolumn{5}{l}{\textbf{NTP-predicted Intent-CFG (Ours)}}\\
+ NTP-predicted Intent-CFG (Ours)$^\dagger$ & \textbf{0.31 / 0.52} & \textbf{0.86 / 2.13}             & \textbf{0.47} / 1.07              & \textbf{7.92} \\
\bottomrule
\specialrule{1pt}{1pt}{2pt}
\end{tabular}
}
\end{table*}

The three alternatives isolate the conditioning-signal source (Table~\ref{tab:language_helps_action}). Trajectory-derived intent ($7.81$) and GT-supplied $3$-class intent ($7.83$) both essentially match the no-intent baseline ($7.83$): each exposes a controllable input but does not improve aggregate planning quality. With NTP-predicted intent and a prototype-grounded embedding refinement (\S\ref{sec:supp_intent_cfg}), MindVLA-U1 + Intent-CFG reaches RFS $\mathbf{7.92}$ ($+0.09$ over baseline), the strongest single-axis improvement among MindVLA-U1's component contributions. Intent-CFG is thus both a controllability result --- language-side state steers continuous action --- and an aggregate planning improvement under the right embedding.

\begin{figure}[!htbp]
\centering
\begin{minipage}[t]{0.48\linewidth}
  \centering
  \includegraphics[width=\linewidth]{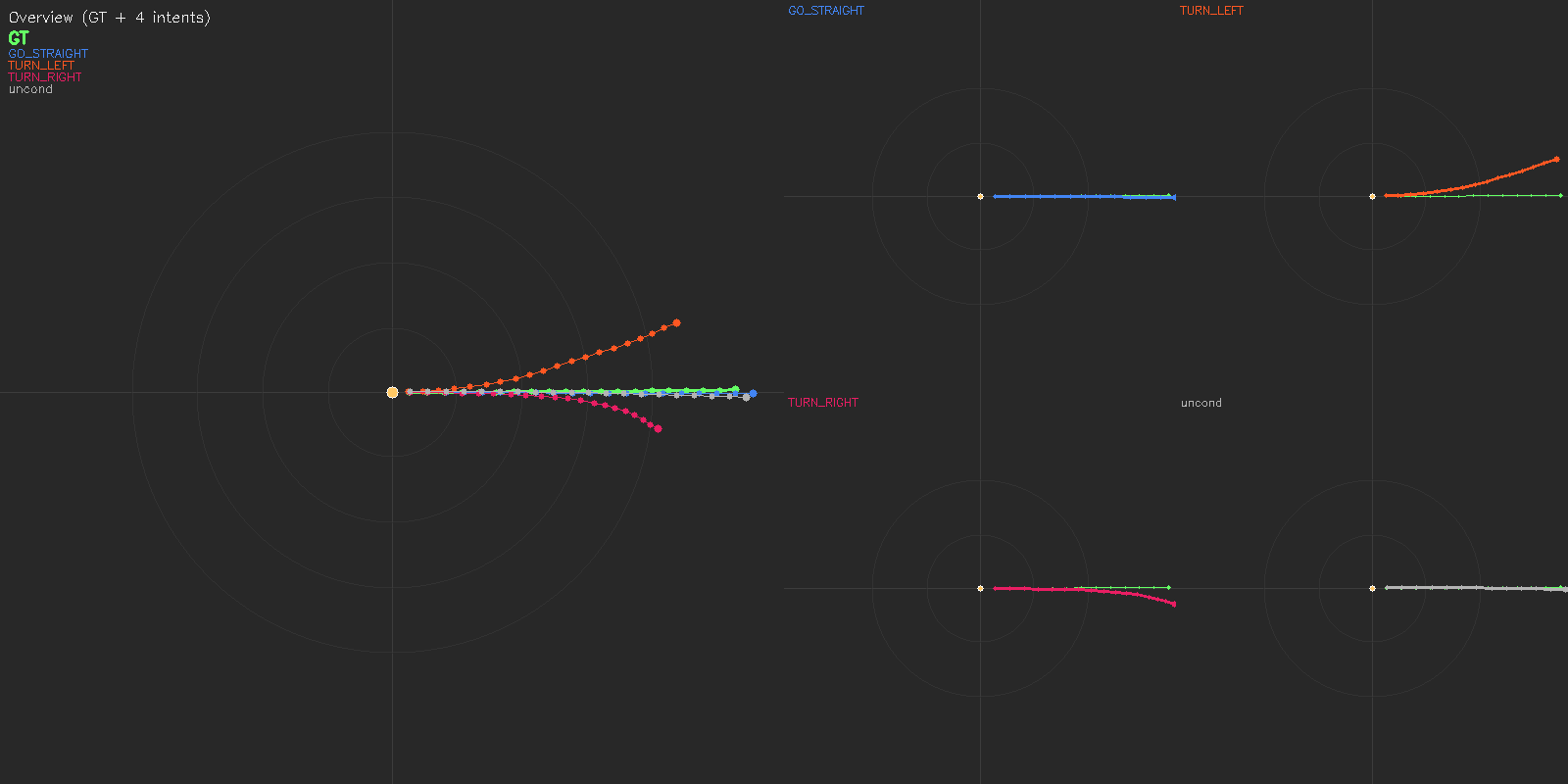}\\[-0.2em]
  {\footnotesize (a) Deployed: WOD-E2E $3$-class intent + uncond.}
\end{minipage}\hfill
\begin{minipage}[t]{0.48\linewidth}
  \centering
  \includegraphics[width=\linewidth]{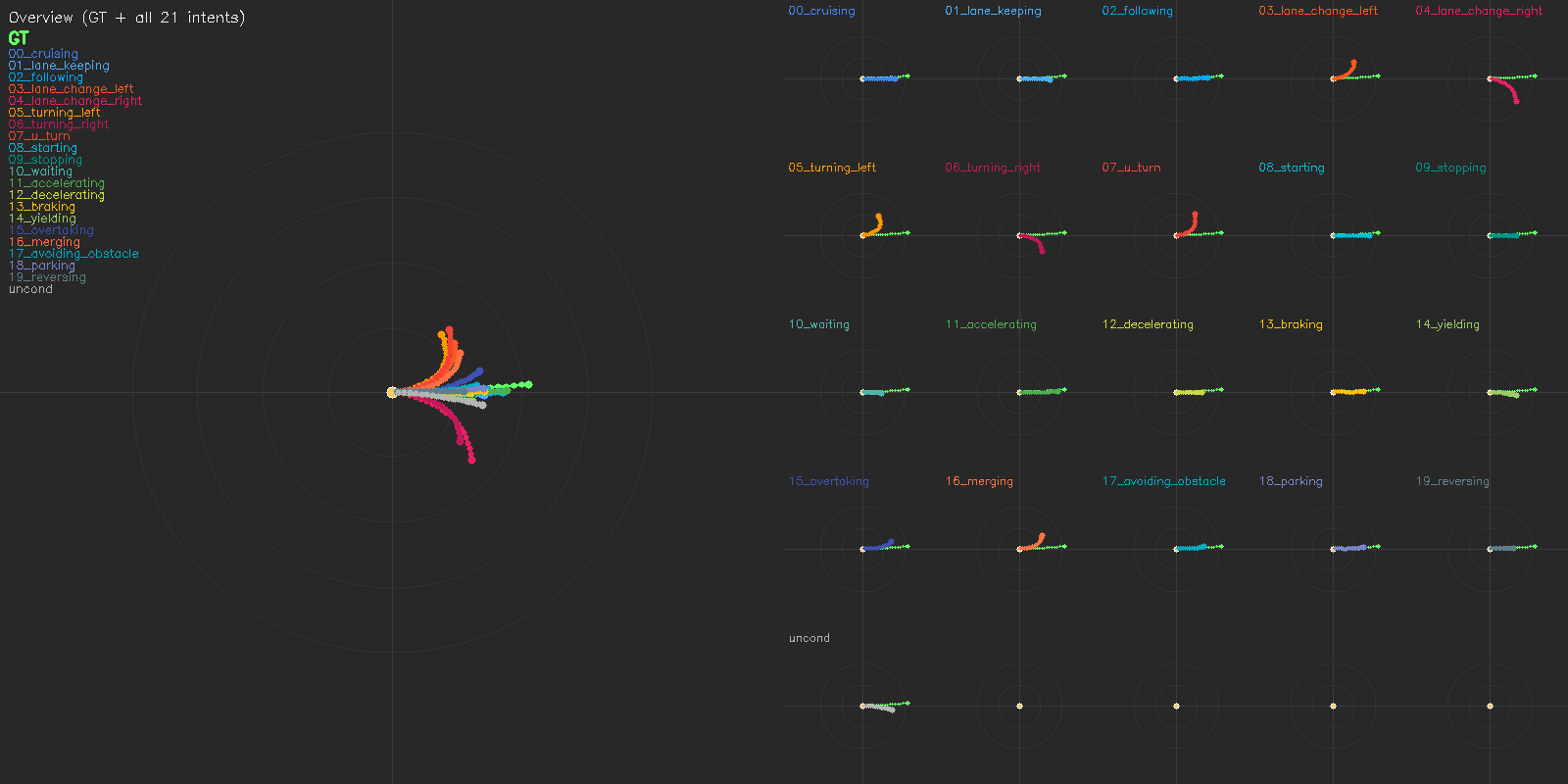}\\[-0.2em]
  {\footnotesize (b) Extended: MindLabel $20$-class intent + uncond.}
\end{minipage}
\caption{\textbf{Intent-CFG as a structural multi-modality mechanism.} Per-intent trajectories on one WOD-E2E frame; \emph{left} of each panel: BEV overview with GT (green); \emph{right}: per-intent subplots. (a) uses the $3$-class GT intent; (b) uses MindLabel's $20$-class extension on the same checkpoint.}
\label{fig:intent_cfg_modes}
\end{figure}

\paragraph{Intent-CFG as a structural multi-modality mechanism.}
Figure~\ref{fig:intent_cfg_modes} visualizes Intent-CFG's qualitative behavior across MindLabel's $20$-class intent vocabulary (\S\ref{sec:supp_mindlabel_dreaming}) on a representative WOD-E2E frame: conditioning the action diffusion on each of the $20$ MindLabel intents (plus an unconditional baseline) at inference produces $21$ corresponding trajectories from the same checkpoint. The trajectories fan out systematically --- each intent steers the action head into a qualitatively distinct, semantically consistent maneuver, while the unconditional baseline collapses to a default behavior close to the GT. The pattern holds regardless of the scene's natural intent: conditioning on \emph{u\_turn}, \emph{starting}, or \emph{reversing} produces the corresponding behavior even when the scene's GT is straight cruising. Intent-CFG therefore exposes multi-modality \emph{structurally}: it is a property of the conditioning interface (the intent token), not of scene content, and the same mechanism applies frame-by-frame regardless of what the scene looks like. Per-intent trajectory \emph{fidelity}, however, is bounded by training-data coverage: rare intents in WOD-E2E (\eg \emph{u\_turn}, \emph{reversing}, \emph{parking}) have few or no GT examples in the training set, so the conditioned trajectory follows the intent direction without recovering a textbook execution of the maneuver. A more balanced training distribution --- which MindLabel's $20$-class intent labels make tractable to construct --- is the natural way to close this gap. Together with the aggregate-RFS result above, the figure shows that the language-to-action route gives the action diffusion an addressable modal axis, and we view this as broad potential for surfacing controllable multi-modality in VLA systems generally --- a direction left to follow-up work (\S\ref{sec:future_roadmap}).

\subsection{Fast/Slow Execution and MoT Design}
\label{sec:ablations_fast_slow}

MindVLA-U1's unified backbone supports sparse MoT routing for fast/slow execution (\S\ref{sec:unified}). We evaluate the MoT design space here on WOD-E2E val with MindVLA-U1's training schedule (\S\ref{sec:setup}); end-to-end throughput against a VA reference is reported in the throughput paragraph below.

\paragraph{MoT expert grouping.}
Two MoT routings are compared against the dense baseline (Table~\ref{tab:abl_mot}), varying which token modalities share an expert; all variants use MindVLA-U1's training schedule and AR+FM objective. Both improve over dense ($7.83$ RFS): the \emph{context vs.\ action} grouping (V,L,M,S$+$A) peaks at $\mathbf{8.01}$ RFS, while the \emph{context vs.\ proprio$+$action} grouping (V,L$+$M,S,A) reaches $7.92$ with the lowest RFS-GT ADE 3s ($0.89$\,m, tied lowest at 5s). We recommend the latter despite its slightly lower headline RFS: grouping memory and state with action gives a modality-pure motor group that runs independently of the context group at fast inference --- under (V,L,M,S)+(A), the fast-mode subgraph would still have to attend memory and ego-state tokens into the context group, undoing the throughput separation that motivates MoT in the first place. The (V,L)+(M,S,A) routing principle (perceptual $\to$ context, motor/proprioceptive $\to$ action) also extends cleanly to additional modalities (perception, world model, safety) inside the same scaffold. MoT thus extends the dense Pareto rather than strictly dominating it: one checkpoint carries slow semantic reasoning and fast action-only execution without collapsing planning quality.

\begin{table*}[!htbp]
\caption{MoT variant comparison on WOD-E2E val. RFS values are best-of-seeds.}
\label{tab:abl_mot}
\footnotesize
\centering
\resizebox{\linewidth}{!}{%
\begin{tabular}{l|c|c|c|>{\columncolor[gray]{0.95}}c}
\specialrule{1pt}{0pt}{1pt}
\toprule
System & L B/R$\uparrow$ & RFS-GT ADE 3/5$\downarrow$ & RFS-matched ADE 3/5$\downarrow$ & RFS$\uparrow$ \\
\midrule
\multicolumn{5}{l}{\textbf{Dense Backbone}}\\
MindVLA-U1 (Dense)                                                            & 0.30 / 0.49             & 0.92 / 2.14                      & 0.50 / \underline{1.05}             & 7.83 \\
\midrule
\multicolumn{5}{l}{\textbf{MoT with Different Expert Groupings}}\\
+MoT, $\{\text{V},\text{L},\text{M},\text{S}\}{+}\{\text{A}\}$ (context vs.\ action)  & \textbf{0.31} / \underline{0.49} & \underline{0.92} / \textbf{2.11} & \textbf{0.46 / 1.03}                 & \textbf{8.01} \\
+MoT, $\{\text{V},\text{L}\}{+}\{\text{M},\text{S},\text{A}\}$ (context vs.\ proprio$+$action, Ours) & \underline{0.30} / \textbf{0.51} & \textbf{0.89 / 2.11} & \underline{0.49 / 1.05} & \underline{7.92} \\
\bottomrule
\specialrule{1pt}{1pt}{2pt}
\end{tabular}
}
\end{table*}

  \paragraph{Closing the VLA throughput gap to VA.}
  \label{sec:throughput}
  Beyond planning quality, deployment also requires real-time control on driving hardware. We benchmark MindVLA-U1's end-to-end inference cost against the strongest VA reference (RAP-DINO~\cite{feng2025rap}) on the same hardware and characterise the fast/slow operating points the unified backbone admits (Table~\ref{tab:fast_slow_fps}). At matched $\sim$$1$B parameter scale, MindVLA-U1's fast inference path reaches near-VA throughput at comparable planning quality without forfeiting the capabilities on which prior VA falls short (Figure~\ref{fig:comparison_matrix}): continuous-action precision, a measurable language$\to$action conditioning route (\S\ref{sec:semantic_intent}), streaming continuity, and the natural-language interface itself. At the deployed Qwen3-VL-$2$B backbone, slow and fast paths are alternative inference orderings of the same unified backbone (\S\ref{sec:unified}) sharing one set of weights: the slow path keeps language-side reasoning available for safety-critical scenes, while the fast path delivers real-time control within $\sim$$0.01$ RFS of slow. The slow--fast cost gap is interface-level rather than capacity-level: autoregressive VQA decoding dominates slow latency and is physically excluded from the action subgraph in fast mode. Distilling the $2$-step flow into a single step, and reusing the prefix KV cache across Euler steps or consecutive frames, are natural next steps toward tighter fast-path budgets.

  \begin{table}[!htbp]
  \caption{Inference latency and throughput on WOD-E2E val (bs$=1$, $1{\times}$\,H200). \emph{Top}: per-stage timing of MindVLA-U1 dense (Qwen3-VL-$2$B) under the slow path (\texttt{vqa\_first\_decoding}); bracket codes --- $V$: visual, $L$: language prompt ($Q$: VQA question), $A$: decoded answer, $\textrm{Act}$: noisy action, $M$: memory. \emph{Bottom}: throughput across slow/fast inference paths sharing one set of weights, with RAP-DINO~\cite{feng2025rap} as the strongest VA reference; the InternVL-2 $1$B row reports the same fast configuration at $\sim$$1$B scale matched to RAP. \textbf{Bold} = recommended Pareto points.}
  \label{tab:fast_slow_fps}
  \label{tab:stage_fps}
  \footnotesize
  \centering
  \resizebox{\linewidth}{!}{%
  \begin{tabular}{l|c|c|c|>{\columncolor[gray]{0.95}}c}
  \specialrule{1pt}{0pt}{1pt}
  \toprule
  Configuration & Latency$\downarrow$ & FPS$\uparrow$ & VRAM$\downarrow$ & RFS$\uparrow$ \\
  \midrule
  \multicolumn{5}{l}{\textbf{Stage-level timing} (Qwen3-VL-$2$B); bracket lists tokens processed per stage.}\\
  \quad Prefix embedding \,[$V{+}L{+}M$, $\approx 624$ tok; vision + past-state + text embed]            & $\sim$$30$\,ms       & ---              & ---         & --- \\
  \quad VQA prefilling \,[$V{+}L{+}M$, $\approx 624$ tok; prefill for VQA decoding]                       & $\sim$$40$\,ms       & ---              & ---         & --- \\
  \quad VQA decoding \,[$A$: $\approx 81$ tok, autoregressive at $30.4$\,ms/tok]                                     & $\sim$$2{,}450$\,ms  & ---              & ---         & --- \\
  \quad Action diffusion, $1$ Euler step \,[$\textrm{Act}$: $\approx 20$ tok appended to $V{+}L{+}M{+}Q{+}A$]   & $\sim$$38$\,ms       & ---              & ---         & --- \\
  \quad Action diffusion, $2$ Euler steps \,[$\textrm{Act}$: $\approx 20$ tok appended to $V{+}L{+}M{+}Q{+}A$] & $\sim$$74$\,ms       & ---              & ---         & --- \\
  \midrule
  \multicolumn{5}{l}{\textbf{End-to-end throughput} of MindVLA-U1 (Qwen3-VL-$2$B Dense, No Intent); compared with best-performing VA models.}\\
  RAP-DINO~\cite{feng2025rap} (VA, $\sim$$0.88$B)                                         & $57$\,ms     & $17.68$            & ---         & $7.91$ \\
  MindVLA-U1 (InternVL-2 1B, template QA, $2$ steps)                                       & $64$\,ms       & $\mathbf{15.55}$   & $2{,}478$\,MB & $\mathbf{7.78}$ \\
  MindVLA-U1 (\texttt{vqa\_first\_decoding}, no intent, $2$ steps)                         & $2{,}594$\,ms  & $0.39$             & $5{,}431$\,MB & $7.83$ \\
  MindVLA-U1 (\texttt{action\_only}, $2$ steps)                                            & $103$\,ms      & $9.70$             & $5{,}412$\,MB & $7.74$ \\
  MindVLA-U1 (\texttt{vqa\_first\_fast}, template QA, $2$ steps)                           & $108$\,ms      & $\mathbf{9.26}$    & $5{,}412$\,MB & $\mathbf{7.82}$ \\
  MindVLA-U1 (\texttt{vqa\_first\_fast}, template QA, $1$ step)                            & $63$\,ms       & $15.92$            & $5{,}412$\,MB & $7.67$ \\
  \bottomrule
  \specialrule{1pt}{1pt}{2pt}
  \end{tabular}%
  }
  \end{table}

  The stage breakdown (Table~\ref{tab:fast_slow_fps}, top) localises the cost: autoregressive $A$ decoding accounts for $\sim$$94\%$ of slow-path latency, while every other stage costs at most one LM forward and is $1$--$2$ orders of magnitude cheaper. Fast inference physically removes $A$ from the action subgraph (\S\ref{sec:unified}), so the same unified backbone supports both deliberative semantic reasoning and reflexive control without separate model copies.

\begin{figure}[!htbp]
\centering
\begin{minipage}[t]{0.19\linewidth}
  \centering
  \includegraphics[width=\linewidth]{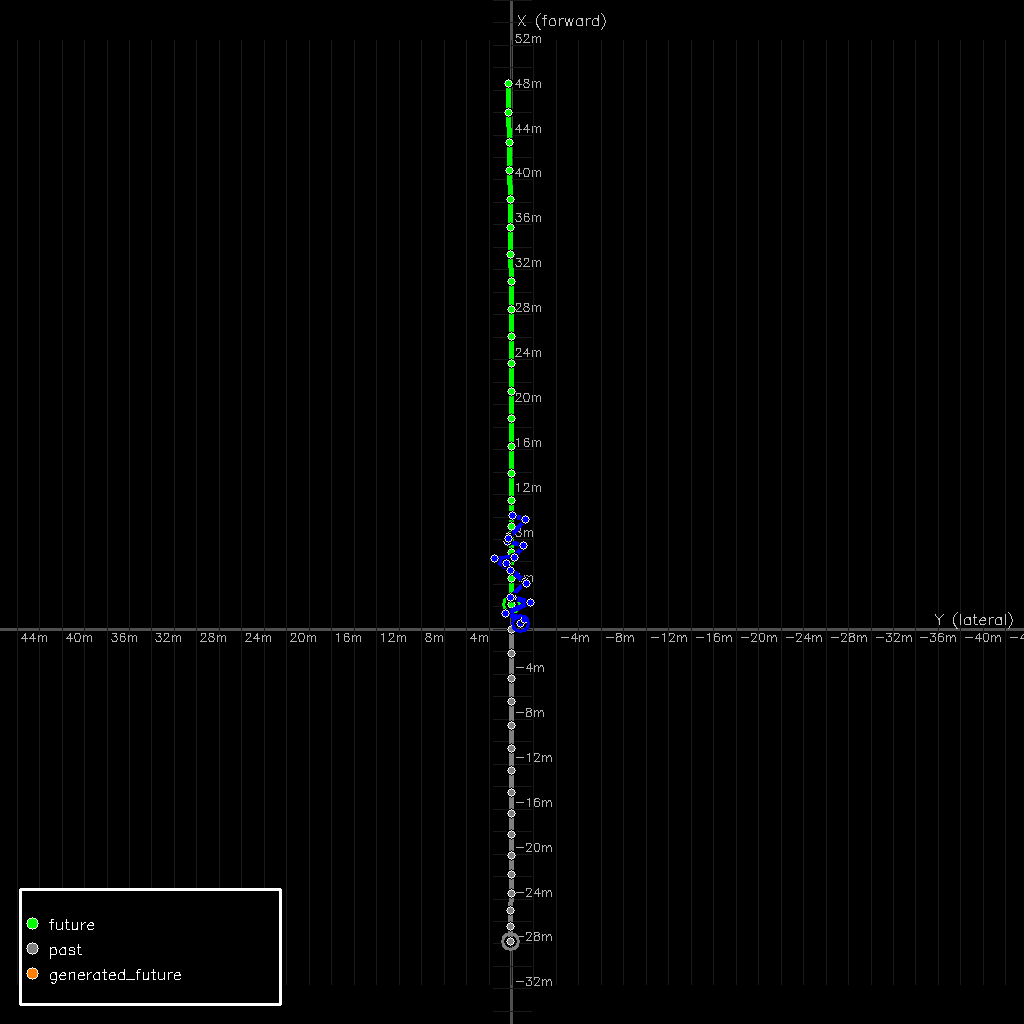}\\[-0.4em]
  {\footnotesize Step $1$}
\end{minipage}\hfill
\begin{minipage}[t]{0.19\linewidth}
  \centering
  \includegraphics[width=\linewidth]{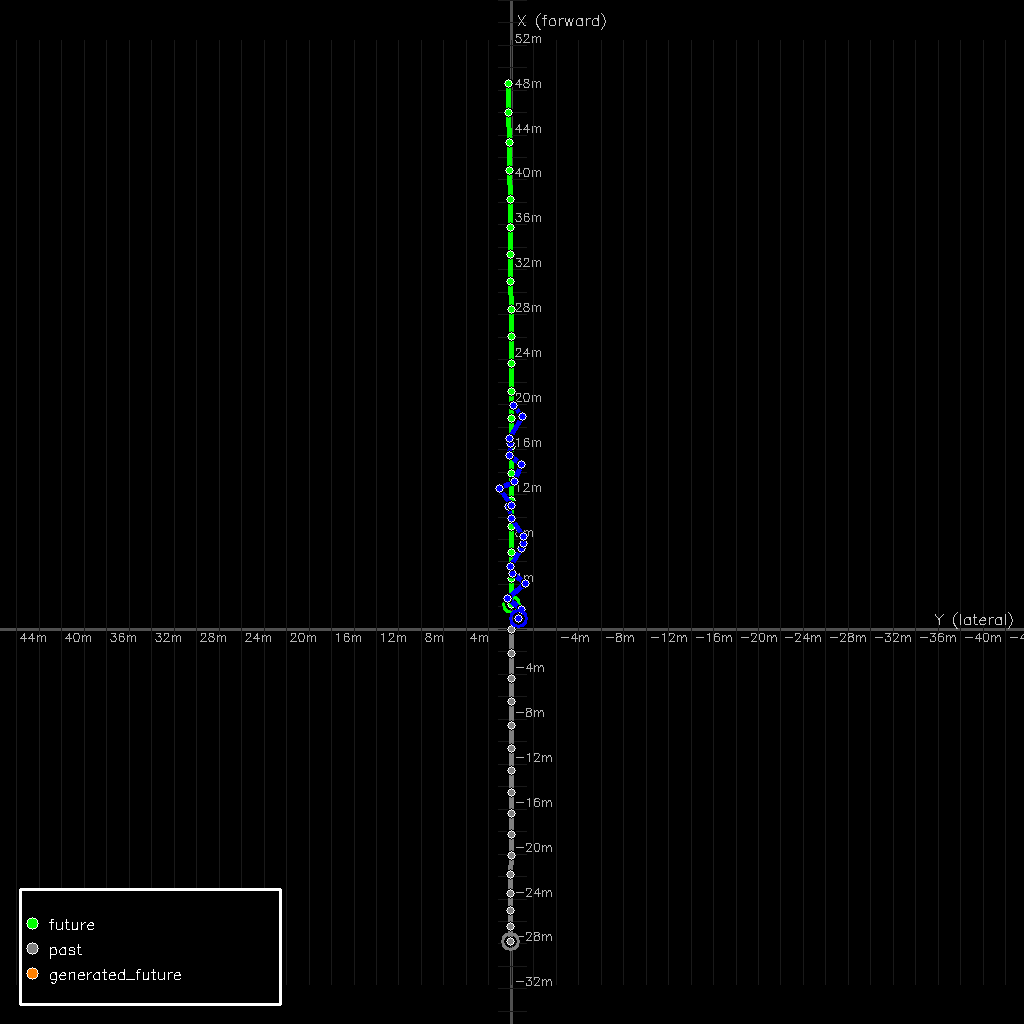}\\[-0.4em]
  {\footnotesize Step $2$}
\end{minipage}\hfill
\begin{minipage}[t]{0.19\linewidth}
  \centering
  \includegraphics[width=\linewidth]{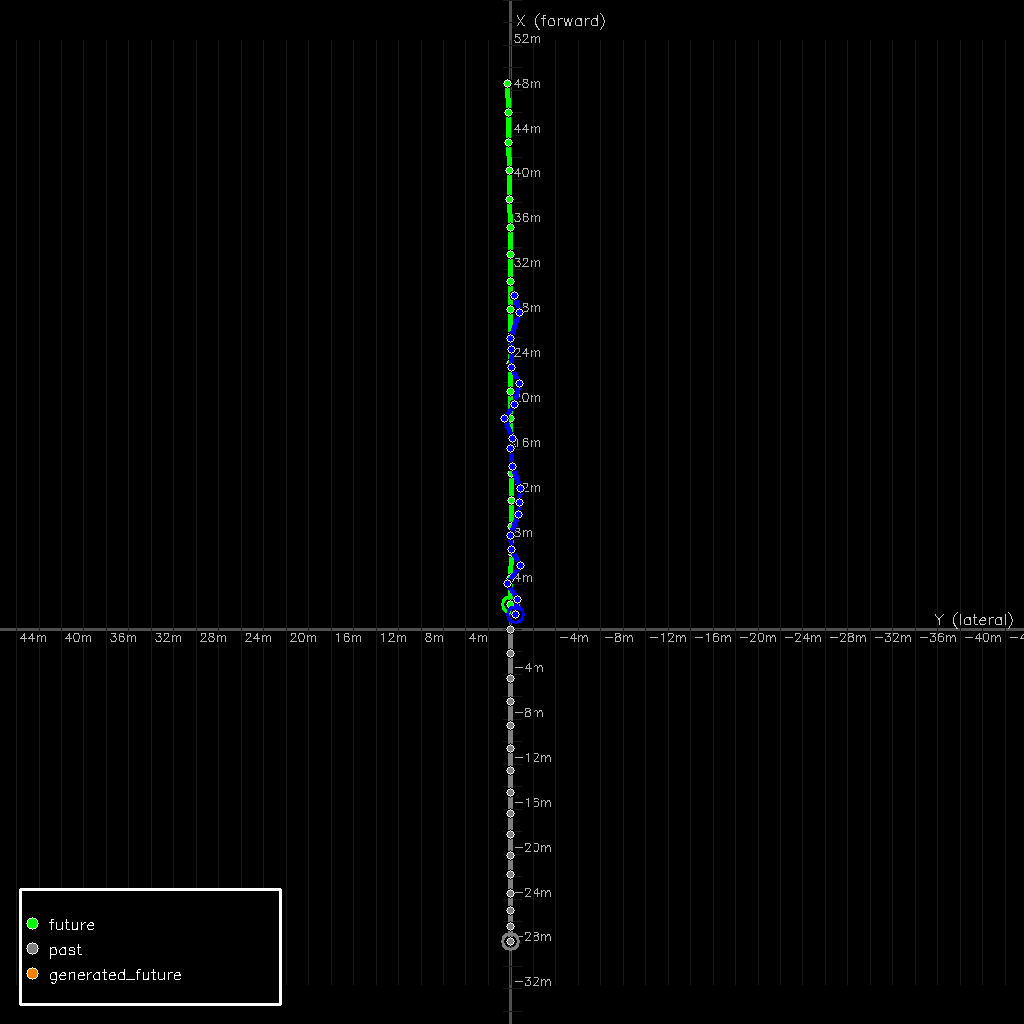}\\[-0.4em]
  {\footnotesize Step $3$}
\end{minipage}\hfill
\begin{minipage}[t]{0.19\linewidth}
  \centering
  \includegraphics[width=\linewidth]{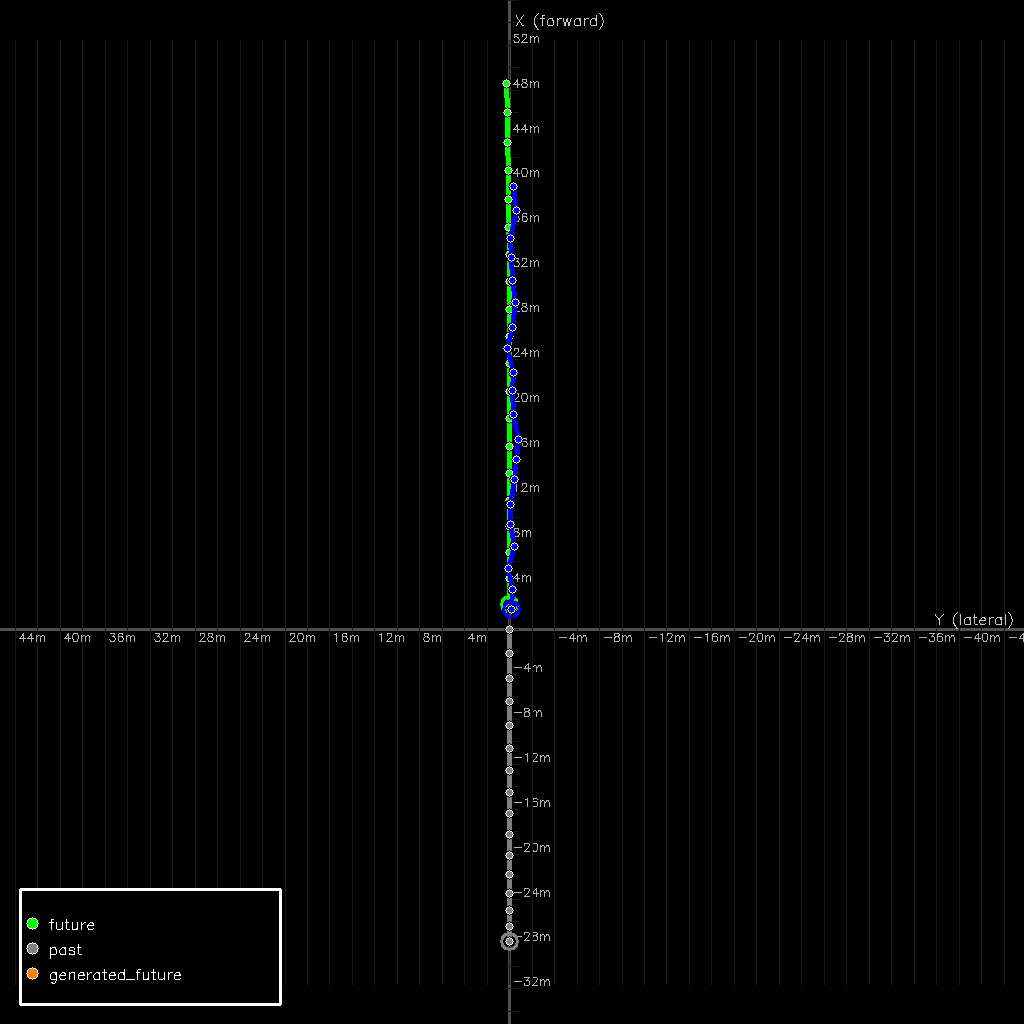}\\[-0.4em]
  {\footnotesize Step $4$}
\end{minipage}\hfill
\begin{minipage}[t]{0.19\linewidth}
  \centering
  \includegraphics[width=\linewidth]{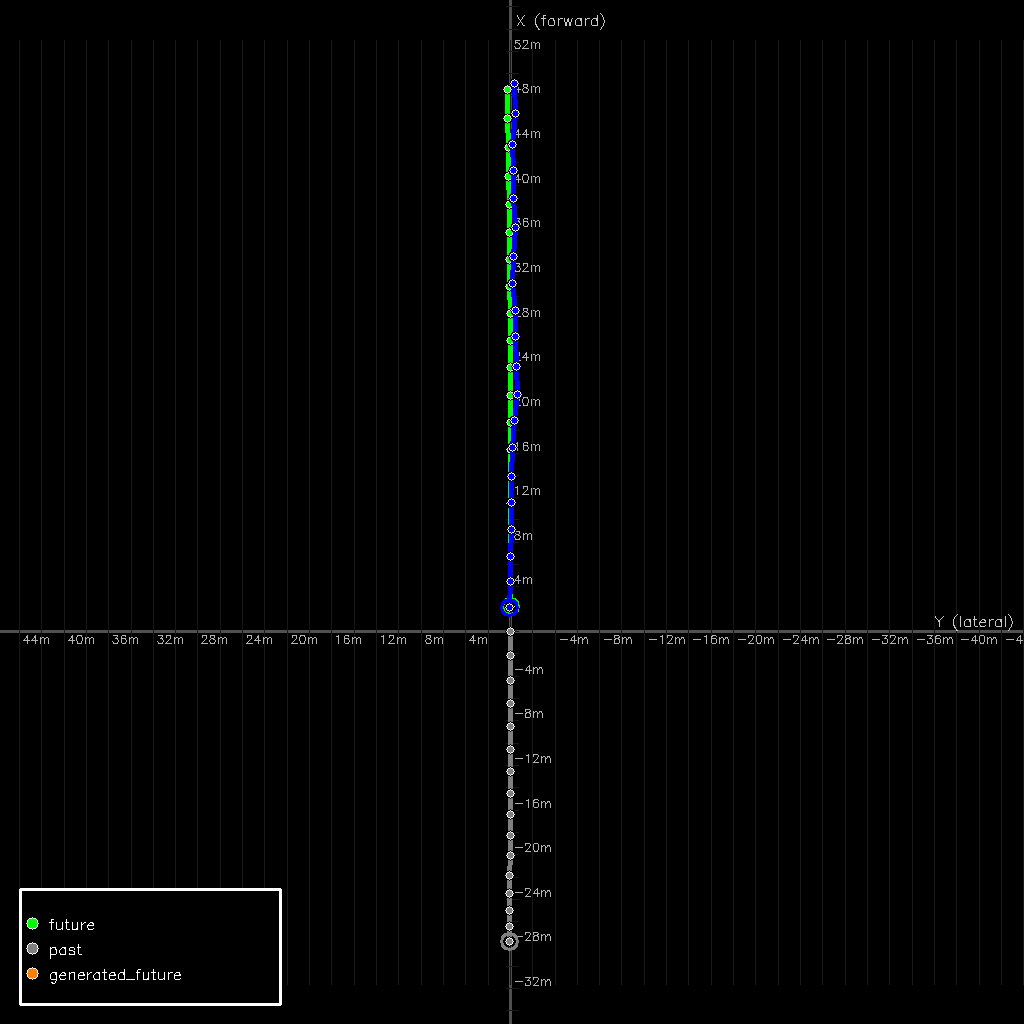}\\[-0.4em]
  {\footnotesize Step $5$}
\end{minipage}
\caption{\textbf{Flow-matching denoising over $5$ Euler steps} on one WOD-E2E frame (BEV: ego forward $+X$, lateral $+Y$) to better demonstrate the denoising process than 2 steps. Green: GT future; gray: past; blue: predicted trajectory after each denoise step. The Gaussian noise input that precedes Step\,$1$ is not shown.}
\label{fig:diffusion_steps}
\end{figure}

  \paragraph{Denoising visualisation.}
  Figure~\ref{fig:diffusion_steps} shows the flow-matching action head denoising over $5$ Euler steps on a representative WOD-E2E frame: starting from Gaussian noise, the predicted trajectory (blue) is progressively integrated along the learned flow toward the ground-truth corridor (green).

\begin{figure}[!htbp]
\centering
\includegraphics[width=0.7\linewidth]{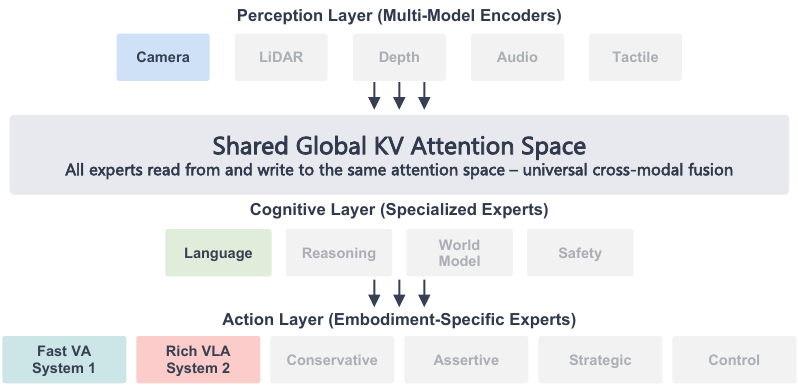}
\caption{\textbf{Foundation architecture vision.} Three-stage generalization of the two-group MoT (\S\ref{sec:unified}, \S\ref{sec:supp_mot_arch}): perception, cognitive (context-group experts), and action (action-group experts). Highlighted: populated in MindVLA-U1; grey: extension slots on the same shared K/V pool.}
\label{fig:mot_foundation}
\end{figure}

\paragraph{Foundation architecture: extension beyond driving.}
\label{sec:mot_foundation}
The deployed two-group MoT generalizes without changing the shared backbone: extensions enter as token-roles on the same K/V pool, not as new graph topologies. Figure~\ref{fig:mot_foundation} sketches the three-stage design space --- perception (modality encoders), cognitive (context-group experts: language, reasoning, world model, safety), and action (action-group experts: behaviors, embodiments) --- of which MindVLA-U1 populates one instantiation per stage (Camera; Language; Fast VA / Rich VLA). The other multiple instantiations are left to future work (\S\ref{sec:future_roadmap}).

\subsection{Streaming Memory for Efficient Temporal Modeling}
\label{sec:ablations_streaming}

The streaming paradigm makes two architectural commitments that we ablate separately: \emph{streaming training} (consecutive frames processed in order so each forward pass sees prior-frame context within the same sample) and a \emph{memory channel} (a learned channel carrying compact temporal state across stream steps, updated end-to-end). We additionally compare against the most direct alternative interface for temporal context --- pushing it into the VLM as multi-frame video tokens.

\paragraph{Streaming training and memory channel.}
Table~\ref{tab:abl_masmp} ablates streaming and memory separately. The two contributions separate cleanly: chunk-wise $\to$ streaming-no-memory gives $+0.04$ RFS, and streaming-no-memory $\to$ streaming + memory gives a further $+0.10$ RFS. The memory channel improves all planning metrics in the streaming setting: rater-matched ADE 5s drops from $1.14$\,m to $\mathbf{1.05}$\,m, and long-horizon sequence ADE drops at every horizon (e.g., $25$\,s ADE $1.54 \to \mathbf{1.50}$\,m). RFS-GT ADE 3s is roughly flat between the two streaming variants, consistent with memory carrying scene-style and behavioral-context cues that affect rater-judged quality more than raw L2 distance to logged GT. This ablation does not isolate gradient-connected training from a detached-gradient variant; we leave that comparison to future work.

\begin{table*}[!htbp]
\footnotesize
\caption{Streaming training and memory channel ablation on WOD-E2E val. $^\flat$ chunk-wise + VLM-side temporal modeling (Qwen3-VL DeepStack).}
\label{tab:abl_masmp}
\centering
\resizebox{\linewidth}{!}{%
\begin{tabular}{l|c|c|c|c|>{\columncolor[gray]{0.95}}c}
\specialrule{1pt}{0pt}{1pt}
\toprule
Configuration & L B/R$\uparrow$ & RFS-GT ADE 3/5$\downarrow$ & RFS-matched ADE 3/5$\downarrow$ & seq ADE 3/5/10/15/20/25$\downarrow$ & RFS$\uparrow$ \\
\midrule
\multicolumn{6}{l}{\textbf{Chunk-wise Training Baseline (single-clip per forward, no streaming memory)}}\\
Chunk-wise, single frame                          & 0.29 / 0.51             & 0.92 / 2.15                     & 0.52 / 1.13                     & ---                                                                                          & 7.69 \\
Chunk-wise + image sequence (4 frames)$^\flat$    & \textbf{0.31 / 0.53}    & 1.14 / 2.53                     & 0.63 / 1.27                     & ---                                                                                          & 7.61 \\
\midrule
\multicolumn{6}{l}{\textbf{Streaming Training Variants}}\\
Streaming training, no memory channel              & 0.31 / 0.49            & 0.98 / 2.30                     & 0.50 / 1.14                     & 0.48 / 0.98 / 1.03 / 1.22 / 1.33 / 1.54                                                      & 7.73 \\
MindVLA-U1 (streaming + memory, Ours)              & 0.30 / 0.49            & \textbf{0.92} / \textbf{2.14}   & \textbf{0.50 / 1.05}            & \textbf{0.46 / 0.94 / 0.99 / 1.19 / 1.29 / 1.50}                                             & \textbf{7.83} \\
\bottomrule
\specialrule{1pt}{1pt}{2pt}
\end{tabular}
}
\end{table*}

\paragraph{Temporal VLM modeling vs.\ streaming memory.}
A central design choice in driving VLAs is \emph{where} temporal context enters the model. The streaming-memory design (\S\ref{sec:streaming_memory}) keeps the per-step VLM input single-frame and propagates a compact learned memory channel across frame-steps. The most obvious alternative is to keep the VLM in charge of temporal mixing by feeding the $K$ most recent frames into the VLM directly and letting attention handle history; we instantiate this with Qwen3-VL DeepStack, which stacks vision tokens from $K{=}4$ frames at the visual encoder (Table~\ref{tab:abl_masmp}, $^\flat$ row). The DeepStack variant slightly degrades planning quality below the chunk-wise single-frame baseline (RFS $7.61$ vs.\ $7.69$), and does not recover the streaming-memory gain ($7.83$) --- the additional vision tokens are heavily redundant across multi-view driving frames, and a generic VLM is not architecturally trained to compress them into a planning-relevant temporal state, while the streaming memory channel is supervised end-to-end through the propagation transformer and is therefore a temporal channel the model has been trained to read and write.

\subsection{RL Post-Training}
\label{sec:rl_post_training}

As a final step toward state-of-the-art planning quality, we post-train the SFT checkpoint with Group Relative Policy Optimization (GRPO)~\cite{guo2025deepseek} using the rater-feedback score (RFS) as the only reward signal --- no auxiliary ADE, FDE, or smoothness terms --- to test whether the unified streaming interface absorbs a rater-aligned reward without architectural change. RL post-training pushes RFS from $7.83$ (SFT init) to $\mathbf{8.20}$ on the validation split and to a leading $\mathbf{7.87}$ on the official test split (Tables~\ref{tab:waymo_e2e_val} and~\ref{tab:waymo_e2e_test}); the trust-region rate (the share of predicted trajectories falling inside at least one rater's trust region) rises from $66.0\%$ to $\mathbf{73.1\%}$. RL hyperparameters and ADE-rater trade-off observations are in \S\ref{sec:supp_rl_extended}.

\subsection{Val/Test Distribution Diagnostic}
\label{sec:val_test_diagnostic}

A central caveat for any WOD-E2E result is that the validation and official test splits are not drawn from the same intent distribution. Applying MindLabel's intent labels (\S\ref{sec:supp_mindlabel_dreaming}) to both splits, Table~\ref{tab:intent_dist_short} reports the four most-shifted intents (full $15$-row distribution: \S\ref{sec:supp_mindlabel_dist_shift}). On val, the distribution is dominated by \emph{active-driving} intents (\emph{accelerating} $19.5\%$, \emph{starting} $10.5\%$). On test, a single \emph{slowdown} intent --- \emph{waiting} ($20.2\%$) --- dominates, while \emph{accelerating} collapses to $4.1\%$ ($-78.9\%$ relative). This structurally explains the val-to-test RFS drop observed across all methods on the leaderboard (MindVLA-U1: $-0.24$ RFS), and reframes the gap as a benchmark-level distribution shift rather than a per-method weakness. MindLabel's intent annotations turn the benchmark itself into an analyzable object --- a diagnostic capability that is part of the interface story.

\begin{table}[!htbp]
\caption{Top-shifted intent classes on WOD-E2E val (RFS-anchored frames) vs.\ test (end-frame protocol), derived from MindLabel intent labels. Full $15$-intent distribution in \S\ref{sec:supp_mindlabel_dist_shift}.}
\label{tab:intent_dist_short}
\footnotesize
\centering
\begin{tabular}{l|cc|c}
\toprule
Intent              & Val share & Test share & $\Delta$ relative \\
\midrule
\emph{accelerating}        & $19.5\%$  & $4.1\%$    & $-78.9\%$ \\
\emph{starting}            & $10.5\%$  & $4.5\%$    & $-57.1\%$ \\
\emph{following}           & $2.7\%$   & $5.6\%$    & $+107\%$ \\
\emph{waiting}             & $7.6\%$   & $\mathbf{20.2\%}$    & $+167\%$ \\
\bottomrule
\end{tabular}
\end{table}

\subsection{VLM Backbone Scaling and the Language-Action Decoupling}
\label{sec:backbone_scaling}

To probe whether driving-VLA planning quality scales with VLM backbone size, we train the same architecture, data, and schedule with Qwen3.5-VL at $0.8$B, $2$B, $4$B, and $9$B (Table~\ref{tab:abl_scaling}). At our default budget, \emph{RFS is non-monotonic in backbone size}: $7.81 \to 7.94 \to 7.86 \to 7.84$, peaking at $2$B and slightly regressing thereafter; extending training to $200$ epochs at $9$B recovers $+0.07$ ($7.91$), suggesting the larger backbones are undertrained on the default schedule. We treat this as evidence about the experimental regime, not a paradigm claim --- a careful framing is needed for both halves of the result, and we give them separately below.

\begin{table*}[!htbp]
\caption{VLM backbone scaling on WOD-E2E val: identical architecture, data, and schedule, only the Qwen3.5-VL size varied. \textbf{Bold} = best per column, \underline{underline} = second-best.}
\label{tab:abl_scaling}
\footnotesize
\centering
\resizebox{\linewidth}{!}{%
\begin{tabular}{l|c|c|c|>{\columncolor[gray]{0.95}}c}
\specialrule{1pt}{0pt}{1pt}
\toprule
Backbone & L B/R$\uparrow$ & RFS-GT ADE 3/5$\downarrow$ & RFS-matched ADE 3/5$\downarrow$ & RFS$\uparrow$ \\
\midrule
\multicolumn{5}{l}{\textbf{Instruction-tuned VLM, default schedule}}\\
Qwen3.5-VL $0.8$B           & 0.27 / 0.48            & 0.85 / 2.04            & 0.48 / 1.07            & 7.81 \\
Qwen3.5-VL $2$B             & 0.27 / 0.48            & 0.85 / 2.04            & 0.48 / 1.07            & \textbf{7.94} \\
Qwen3.5-VL $4$B             & 0.28 / 0.48            & 0.85 / \underline{2.03} & 0.47 / 1.04           & 7.86 \\
Qwen3.5-VL $9$B             & \textbf{0.28 / 0.49}   & 0.88 / 2.09            & \underline{0.47} / 1.04 & 7.84 \\
\midrule
\multicolumn{5}{l}{\textbf{Base VLM (no instruction tuning)}}\\
Qwen3.5-VL $2$B (base)      & 0.12 / 0.29            & 0.85 / 2.08            & 0.48 / 1.06            & 7.91 \\
Qwen3.5-VL $4$B (base)      & 0.13 / 0.30            & 0.86 / 2.05            & 0.47 / 1.03            & 7.86 \\
\midrule
\multicolumn{5}{l}{\textbf{Extended training schedule}}\\
Qwen3.5-VL $9$B ($200$ ep)  & 0.28 / 0.48            & \textbf{0.87 / 2.05}   & \textbf{0.46 / 1.00}   & \underline{7.91} \\
\bottomrule
\specialrule{1pt}{1pt}{2pt}
\end{tabular}%
}
\end{table*}

\paragraph{Language quality and action quality are structurally decoupled.}
The base-vs.\ instruction-tuned comparison at $2$B is the crisper finding: BLEU-4 / ROUGE-L collapse from $0.27 / 0.48$ to $0.12 / 0.29$ (a $\sim$$2{\times}$ drop in VQA quality) while RFS is essentially preserved ($7.94 \to 7.91$); the same pattern repeats at $4$B. This is mechanistic evidence for the unified backbone's central design property: Intent-CFG (\S\ref{sec:unified_capability_diagnostics}) does not consume the VLM's \emph{language generation} capacity. It consumes the VLM's \emph{scene representation}. The intent token enters the action diffusion through its conditioning embedding, not through its decoded answer-token surface form; a VLM whose instruction-following fluency is damaged or never trained still produces an intent token whose embedding routes action correctly via CFG, as long as the shared backbone has learned a scene representation in which intent is classifiable. Controllability is therefore a property of the shared-backbone interface, not a byproduct of strong LLM language quality --- the Intent-CFG result of \S\ref{sec:unified_capability_diagnostics} and the language-quality result are separable contributions, not co-conditioned on backbone scale.

\paragraph{Why this is not a VLA scaling-law claim --- and what one would need.}
We do \emph{not} claim that driving VLA cannot scale. Three confounds prevent the non-monotonic curve in Table~\ref{tab:abl_scaling} from being read as a paradigm result. \emph{Schedule}: the $9$B model gains $+0.07$ RFS when extended from the default schedule to $200$ epochs, indicating that larger backbones are undertrained at fixed-epoch budgets --- compute-matched (FLOPs-equivalent) training, not parameter-matched, is the right scaling axis. \emph{Data}: WOD-E2E has ${\sim}2$K training sequences; at this data scale, larger backbones may saturate against the dataset rather than against representational capacity, so a clean scaling study has to scale data with model jointly. MindLabel addresses this lever on the language/preference side; on-vehicle telemetry is the larger reservoir. \emph{Action-interface capacity}: the flow-matching action head is held constant across all backbone sizes, so if the bottleneck at our budget is the action head, backbone scale cannot relax it; an honest scaling study has to vary action-interface capacity as a separate axis. Additionally, open-loop RFS may saturate before closed-loop policy quality does, so without a closed-loop evaluation channel a scaling curve on RFS understates the value of scale. A clean VLA scaling law requires jointly varying \emph{model, data, action-interface, schedule, and evaluation channel} --- a research thread we describe in \S\ref{sec:future_roadmap} rather than relitigate here.

\paragraph{Implications for deployment.}
At this scale, the unified backbone already delivers state-of-the-art planning at $0.8$B (RFS $7.81$, on par with the strongest non-MindVLA-U1 VLA results) --- useful for deployment-latency budgets, and matched to the fast-mode throughput result of \S\ref{sec:ablations_fast_slow}. Combined with the decoupling above, this gives MindVLA-U1 a property that is operationally useful: planning quality does not require a flagship-scale VLM, and Intent-CFG controllability does not degrade with backbone-quality variation. Whether \emph{aggressively} larger backbones unlock further gains is a question for the proper scaling study, not for the present paper.

\section{Related Work}
\label{sec:related}

\emph{Vision-to-Action (VA) models for end-to-end autonomous driving}~\cite{hu2023planning,jiang2023vad,chen2024vadv2,sun2025sparsedrive,zheng2024genad,Weng_2024_CVPR,li2024ego,chitta2022transfuser,liao2025diffusiondrive,zheng2025diffusion,feng2025rap} map camera observations directly to continuous trajectories. Their directness is the strength: action stays continuous, the optimization target is close to the control objective, and strong VA systems set a high bar for trajectory quality. The limitation is semantic: VA representations are implicit, with no language-mediated interface for inspecting scene concepts, conditioning on intent, or testing whether language-grounded knowledge changes the action. MindVLA-U1 keeps VA's continuous-control discipline while adding a measurable language-to-action pathway.

\emph{Driving Vision-Language-Action (VLA) models}~\cite{shao2024lmdrive,tian2024drivevlm,jiang2024senna,sima2024drivelm,hwang2024emma,wang2025omnidrive,chi2025impromptu,zhou2025autovla,zeng2025futuresightdrive,yuan2025autodrive,rowe2025poutine,fu2025orion,renz2025simlingo,li2025recogdrive,li2025drivevla,wang2025alpamayo} attach VLM-style semantic reasoning to action generation, but often inherit interfaces convenient for language models rather than natural for control. The relevant design axes expand on the gaps in \S\ref{sec:intro}: action representation (token-quantized vs.\ continuous), temporal continuity (fixed action chunks vs.\ streaming), temporal-token efficiency (redundant multi-frame VLM input vs.\ compact memory), language-action coupling (asserted vs.\ measured), and driving-native supervision (generic VQA vs.\ scene-grounded language and trajectory-preference data). A practical issue cuts across these: heavy reliance on simulated benchmarks (\eg, CARLA~\cite{dosovitskiy2017carla}) whose closed-loop scores routinely fail to predict on-road behavior. MindVLA-U1 is positioned as an interface correction across these axes rather than a single isolated module, evaluated entirely on real-world WOD-E2E. Extended discussion --- including dual-head VLA precedents, intent-conditioned policies in broader robot learning, modality-routed compute, and the open question of VLA backbone scaling --- is deferred to \S\ref{sec:supp_related}.

\section{Conclusion, Limitations, and Future Roadmap}
\label{sec:future}

\subsection{Conclusion}
\label{sec:conclusion}

MindVLA-U1 argues that the VLA--VA planning gap on real-world driving is not a paradigm cost of semantic reasoning but largely an interface problem, and resolves it with a unified shared backbone that produces autoregressive language and flow-matching continuous action in one forward pass over one shared representation, a streaming memory channel that replaces redundant multi-frame VLM modeling and fixed action chunks, an Intent-CFG bridge that gives language a measurable causal route into action, and dense/sparse-MoT fast/slow execution that recovers VA-class throughput from a single checkpoint. On WOD-E2E, the resulting system reaches the highest reported RFS on the official test split with two diffusion steps, and after RL post-training surpasses experienced human raters ($8.20$ vs.\ $8.13$ GT RFS, val). The val/test distribution-shift diagnostic (\S\ref{sec:val_test_diagnostic}) reframes the residual gap as a benchmark-level intent shift rather than a per-method weakness, and the scaling study (\S\ref{sec:backbone_scaling}) shows that the language-to-action interface is structurally decoupled from VLM language quality --- the Intent-CFG controllability result holds even when VQA quality collapses by $2{\times}$. Taken together, these results validate the central thesis: \emph{the VLA-VA gap on real-world long-tail driving is interface-level, and the unified interface closes it without sacrificing precision, throughput, or the natural-language interface}.

\subsection{Limitations}
\label{sec:limitations}

We list the scope conditions of the results above, framed as what the paper does \emph{not} claim, so follow-up work can address each axis cleanly. \emph{(i) Open-loop evaluation only.} WOD-E2E is logged, not reactive; the RFS metric and the rater panel test open-loop trajectory quality, not closed-loop policy quality under the model's own action distribution. Surpassing the $8.13$ GT RFS establishes open-loop superiority; closed-loop on-vehicle behavior is a separate evaluation channel that we have not measured. \emph{(ii) Single benchmark.} All headline results are on WOD-E2E. Cross-benchmark transfer to nuScenes, NAVSIM, or on-vehicle deployment is not yet verified; the architectural arguments generalize, but the empirical specifics may not. \emph{(iii) No definitive VLA scaling-law claim.} Per \S\ref{sec:backbone_scaling}, naive parameter scaling does not return monotonically at our budget. We have not separated schedule, data, action-interface, and evaluation-channel confounds cleanly enough to call this a paradigm result; the broader question of whether driving VLA scales is left open. \emph{(iv) Partial use of MindLabel.} The main results consume only the basic scene-grounded VQA stream and the GT $3$-class intent label. Dreamed alternative trajectories, GT/dreamed trajectory-evaluation QA, $20$-class intent, and chain-of-thought rationales are released as part of MindLabel but unexploited in the model trained here.

\subsection{Future Roadmap}
\label{sec:future_roadmap}

The limitations above invite natural follow-up directions on top of MindVLA-U1's unified interface; below we list several worth exploring:

\emph{1. Deeper RL post-training.} The current RL stage is a single SFT$\to$GRPO step on the RFS scalar (\S\ref{sec:rl_post_training}); the unified backbone admits richer reward sources (rater-panel ensembles, MindLabel trajectory-evaluation QA as a learned reward) and longer-horizon credit assignment, which we plan to explore.

\emph{2. Intent diversity and language alignment.} Intent-CFG currently uses a $3$-class GT-supervised intent; MindLabel's $20$-class intent vocabulary, dreamed-trajectory diversity, and free-form language conditioning are natural extensions toward closing the predicted-intent vs.\ GT-intent gap of \S\ref{sec:unified_capability_diagnostics}.

\emph{3. Cross-benchmark verification.} Headline results in this paper are confined to WOD-E2E (limitation (ii)); extending evaluation to additional open AD benchmarks --- nuScenes, NAVSIM, and similar --- together with on-vehicle deployment would test whether the architectural arguments transfer empirically as well as conceptually.

\emph{4. Toward a controlled VLA scaling-law study.} As \S\ref{sec:backbone_scaling} argues, a clean scaling result would need compute-matched (not parameter-matched) training, data scaled jointly with model size, action-interface capacity varied as a separate axis, and closed-loop evaluation. We see this as a longer-term effort rather than a next step.

\emph{5. Closed-loop training with a learned world model.} A learned world model routed through the existing backbone as another modality --- rather than a separate subsystem or a hand-built reactive simulator --- would serve as the reactive environment in which the policy rolls out counterfactual futures under reward, addressing limitation (i) above.

\emph{6. MoT expansion to additional experts.} The routing principle in \S\ref{sec:ablations_fast_slow} (perceptual modalities $\to$ context, motor/proprioceptive $\to$ action) extends naturally to additional context-group experts (\eg reasoning, world-model rollout, safety) that share the K/V pool with the language expert but maintain independent FFN capacity; how far this extends in practice is an open question.

\emph{7. VLA inference acceleration.} Several orthogonal speedups remain open for the fast path: distilling the $2$-step flow into a single-step model, reusing the prefix KV cache across Euler steps and across consecutive frames, and decoupling the slow language pathway's update rate from the fast action loop. Each tightens the deployment budget without changing the unified architecture, and brings real-time control closer to VA-class throughput at the deployed VLA scale.

\emph{8. Transfer to embodied tasks.} The unified-interface philosophy generalizes beyond driving; whether the structural decoupling observed in \S\ref{sec:backbone_scaling} is driving-specific or a general property of unified backbones is something we hope to test on manipulation and navigation in future work.

The unified interface presented here is one step on the longer path toward fully autonomous driving (\emph{i.e.}, L4); many more will be needed --- in benchmarks, in on-vehicle validation, in closed-loop training with learned world models, and in directions we cannot yet anticipate. We present MindVLA-U1 in that spirit, and look forward to the perspectives, datasets, and use cases the community will bring --- both to driving, and to the broader physical-intelligence agenda it ultimately serves.

\bibliographystyle{unsrtnat}
\bibliography{cite}

@inproceedings{hu2023planning,
  title={Planning-oriented autonomous driving},
  author={Hu, Yihan and Yang, Jiazhi and Chen, Li and Li, Keyu and Sima, Chonghao and Zhu, Xizhou and Chai, Siqi and Du, Senyao and Lin, Tianwei and Wang, Wenhai and others},
  booktitle={Proceedings of the IEEE/CVF conference on computer vision and pattern recognition},
  pages={17853--17862},
  year={2023}
}

@inproceedings{jiang2023vad,
  title={Vad: Vectorized scene representation for efficient autonomous driving},
  author={Jiang, Bo and Chen, Shaoyu and Xu, Qing and Liao, Bencheng and Chen, Jiajie and Zhou, Helong and Zhang, Qian and Liu, Wenyu and Huang, Chang and Wang, Xinggang},
  booktitle={Proceedings of the IEEE/CVF International Conference on Computer Vision},
  pages={8340--8350},
  year={2023}
}

@article{chen2024vadv2,
  title={Vadv2: End-to-end vectorized autonomous driving via probabilistic planning},
  author={Chen, Shaoyu and Jiang, Bo and Gao, Hao and Liao, Bencheng and Xu, Qing and Zhang, Qian and Huang, Chang and Liu, Wenyu and Wang, Xinggang},
  journal={arXiv preprint arXiv:2402.13243},
  year={2024}
}

@inproceedings{sun2025sparsedrive,
  title={Sparsedrive: End-to-end autonomous driving via sparse scene representation},
  author={Sun, Wenchao and Lin, Xuewu and Shi, Yining and Zhang, Chuang and Wu, Haoran and Zheng, Sifa},
  booktitle={2025 IEEE International Conference on Robotics and Automation (ICRA)},
  pages={8795--8801},
  year={2025},
  organization={IEEE}
}

@inproceedings{zheng2024genad,
  title={Genad: Generative end-to-end autonomous driving},
  author={Zheng, Wenzhao and Song, Ruiqi and Guo, Xianda and Zhang, Chenming and Chen, Long},
  booktitle={European Conference on Computer Vision},
  pages={87--104},
  year={2024},
  organization={Springer}
}

@InProceedings{Weng_2024_CVPR,
    author    = {Weng, Xinshuo and Ivanovic, Boris and Wang, Yan and Wang, Yue and Pavone, Marco},
    title     = {PARA-Drive: Parallelized Architecture for Real-time Autonomous Driving},
    booktitle = {Proceedings of the IEEE/CVF Conference on Computer Vision and Pattern Recognition (CVPR)},
    month     = {June},
    year      = {2024},
    pages     = {15449-15458}
}

@inproceedings{li2024ego,
  title={Is ego status all you need for open-loop end-to-end autonomous driving?},
  author={Li, Zhiqi and Yu, Zhiding and Lan, Shiyi and Li, Jiahan and Kautz, Jan and Lu, Tong and Alvarez, Jose M},
  booktitle={Proceedings of the IEEE/CVF Conference on Computer Vision and Pattern Recognition},
  pages={14864--14873},
  year={2024}
}

@article{chitta2022transfuser,
  title={Transfuser: Imitation with transformer-based sensor fusion for autonomous driving},
  author={Chitta, Kashyap and Prakash, Aditya and Jaeger, Bernhard and Yu, Zehao and Renz, Katrin and Geiger, Andreas},
  journal={IEEE transactions on pattern analysis and machine intelligence},
  volume={45},
  number={11},
  pages={12878--12895},
  year={2022},
  publisher={IEEE}
}

@inproceedings{liao2025diffusiondrive,
  title={Diffusiondrive: Truncated diffusion model for end-to-end autonomous driving},
  author={Liao, Bencheng and Chen, Shaoyu and Yin, Haoran and Jiang, Bo and Wang, Cheng and Yan, Sixu and Zhang, Xinbang and Li, Xiangyu and Zhang, Ying and Zhang, Qian and others},
  booktitle={Proceedings of the Computer Vision and Pattern Recognition Conference},
  pages={12037--12047},
  year={2025}
}

@article{zheng2025diffusion,
  title={Diffusion-based planning for autonomous driving with flexible guidance},
  author={Zheng, Yinan and Liang, Ruiming and Zheng, Kexin and Zheng, Jinliang and Mao, Liyuan and Li, Jianxiong and Gu, Weihao and Ai, Rui and Li, Shengbo Eben and Zhan, Xianyuan and others},
  journal={arXiv preprint arXiv:2501.15564},
  year={2025}
}

@article{feng2025rap,
  title={Rap: 3d rasterization augmented end-to-end planning},
  author={Feng, Lan and Gao, Yang and Zablocki, Eloi and Li, Quanyi and Li, Wuyang and Liu, Sichao and Cord, Matthieu and Alahi, Alexandre},
  journal={arXiv preprint arXiv:2510.04333},
  year={2025}
}

@inproceedings{shao2024lmdrive,
  title={Lmdrive: Closed-loop end-to-end driving with large language models},
  author={Shao, Hao and Hu, Yuxuan and Wang, Letian and Song, Guanglu and Waslander, Steven L and Liu, Yu and Li, Hongsheng},
  booktitle={Proceedings of the IEEE/CVF conference on computer vision and pattern recognition},
  pages={15120--15130},
  year={2024}
}

@article{tian2024drivevlm,
  title={Drivevlm: The convergence of autonomous driving and large vision-language models},
  author={Tian, Xiaoyu and Gu, Junru and Li, Bailin and Liu, Yicheng and Wang, Yang and Zhao, Zhiyong and Zhan, Kun and Jia, Peng and Lang, Xianpeng and Zhao, Hang},
  journal={arXiv preprint arXiv:2402.12289},
  year={2024}
}

@article{jiang2024senna,
  title={Senna: Bridging large vision-language models and end-to-end autonomous driving},
  author={Jiang, Bo and Chen, Shaoyu and Liao, Bencheng and Zhang, Xingyu and Yin, Wei and Zhang, Qian and Huang, Chang and Liu, Wenyu and Wang, Xinggang},
  journal={arXiv preprint arXiv:2410.22313},
  year={2024}
}

@inproceedings{sima2024drivelm,
  title={Drivelm: Driving with graph visual question answering},
  author={Sima, Chonghao and Renz, Katrin and Chitta, Kashyap and Chen, Li and Zhang, Hanxue and Xie, Chengen and Bei{\ss}wenger, Jens and Luo, Ping and Geiger, Andreas and Li, Hongyang},
  booktitle={European conference on computer vision},
  pages={256--274},
  year={2024},
  organization={Springer}
}

@article{hwang2024emma,
  title={Emma: End-to-end multimodal model for autonomous driving},
  author={Hwang, Jyh-Jing and Xu, Runsheng and Lin, Hubert and Hung, Wei-Chih and Ji, Jingwei and Choi, Kristy and Huang, Di and He, Tong and Covington, Paul and Sapp, Benjamin and others},
  journal={arXiv preprint arXiv:2410.23262},
  year={2024}
}

@inproceedings{wang2025omnidrive,
  title={Omnidrive: A holistic vision-language dataset for autonomous driving with counterfactual reasoning},
  author={Wang, Shihao and Yu, Zhiding and Jiang, Xiaohui and Lan, Shiyi and Shi, Min and Chang, Nadine and Kautz, Jan and Li, Ying and Alvarez, Jose M},
  booktitle={Proceedings of the computer vision and pattern recognition conference},
  pages={22442--22452},
  year={2025}
}

@article{chi2025impromptu,
  title={Impromptu vla: Open weights and open data for driving vision-language-action models},
  author={Chi, Haohan and Gao, Huan-ang and Liu, Ziming and Liu, Jianing and Liu, Chenyu and Li, Jinwei and Yang, Kaisen and Yu, Yangcheng and Wang, Zeda and Li, Wenyi and others},
  journal={arXiv preprint arXiv:2505.23757},
  year={2025}
}

@article{zhou2025autovla,
  title={Autovla: A vision-language-action model for end-to-end autonomous driving with adaptive reasoning and reinforcement fine-tuning},
  author={Zhou, Zewei and Cai, Tianhui and Zhao, Seth Z and Zhang, Yun and Huang, Zhiyu and Zhou, Bolei and Ma, Jiaqi},
  journal={arXiv preprint arXiv:2506.13757},
  year={2025}
}

@article{zeng2025futuresightdrive,
  title={Futuresightdrive: Thinking visually with spatio-temporal cot for autonomous driving},
  author={Zeng, Shuang and Chang, Xinyuan and Xie, Mengwei and Liu, Xinran and Bai, Yifan and Pan, Zheng and Xu, Mu and Wei, Xing and Guo, Ning},
  journal={arXiv preprint arXiv:2505.17685},
  year={2025}
}

@article{yuan2025autodrive,
  title={AutoDrive-R$^{2}$: Incentivizing Reasoning and Self-Reflection Capacity for VLA Model in Autonomous Driving},
  author={Yuan, Zhenlong and Qian, Chengxuan and Tang, Jing and Chen, Rui and Song, Zijian and Sun, Lei and Chu, Xiangxiang and Cai, Yujun and Zhang, Dapeng and Li, Shuo},
  journal={arXiv preprint arXiv:2509.01944},
  year={2025}
}

@article{rowe2025poutine,
  title={Poutine: Vision-language-trajectory pre-training and reinforcement learning post-training enable robust end-to-end autonomous driving},
  author={Rowe, Luke and de Schaetzen, Rodrigue and Girgis, Roger and Pal, Christopher and Paull, Liam},
  journal={arXiv preprint arXiv:2506.11234},
  year={2025}
}

@inproceedings{fu2025orion,
  title={Orion: A holistic end-to-end autonomous driving framework by vision-language instructed action generation},
  author={Fu, Haoyu and Zhang, Diankun and Zhao, Zongchuang and Cui, Jianfeng and Liang, Dingkang and Zhang, Chong and Zhang, Dingyuan and Xie, Hongwei and Wang, Bing and Bai, Xiang},
  booktitle={Proceedings of the IEEE/CVF International Conference on Computer Vision},
  pages={24823--24834},
  year={2025}
}

@inproceedings{renz2025simlingo,
  title={Simlingo: Vision-only closed-loop autonomous driving with language-action alignment},
  author={Renz, Katrin and Chen, Long and Arani, Elahe and Sinavski, Oleg},
  booktitle={Proceedings of the Computer Vision and Pattern Recognition Conference},
  pages={11993--12003},
  year={2025}
}

@inproceedings{zhang2025adadrive,
  title={Adadrive: Self-adaptive slow-fast system for language-grounded autonomous driving},
  author={Zhang, Ruifei and Xie, Junlin and Zhang, Wei and Chen, Weikai and Tan, Xiao and Wan, Xiang and Li, Guanbin},
  booktitle={Proceedings of the IEEE/CVF International Conference on Computer Vision},
  pages={5112--5121},
  year={2025}
}

@article{luo2025adathinkdrive,
  title={Adathinkdrive: Adaptive thinking via reinforcement learning for autonomous driving},
  author={Luo, Yuechen and Li, Fang and Xu, Shaoqing and Lai, Zhiyi and Yang, Lei and Chen, Qimao and Luo, Ziang and Xie, Zixun and Jiang, Shengyin and Liu, Jiaxin and others},
  journal={arXiv preprint arXiv:2509.13769},
  year={2025}
}

@article{li2025recogdrive,
  title={Recogdrive: A reinforced cognitive framework for end-to-end autonomous driving},
  author={Li, Yongkang and Xiong, Kaixin and Guo, Xiangyu and Li, Fang and Yan, Sixu and Xu, Gangwei and Zhou, Lijun and Chen, Long and Sun, Haiyang and Wang, Bing and others},
  journal={arXiv preprint arXiv:2506.08052},
  year={2025}
}

@article{li2025drivevla,
  title={DriveVLA-W0: World models amplify data scaling law in autonomous driving},
  author={Li, Yingyan and Shang, Shuyao and Liu, Weisong and Zhan, Bing and Wang, Haochen and Wang, Yuqi and Chen, Yuntao and Wang, Xiaoman and An, Yasong and Tang, Chufeng and others},
  journal={arXiv preprint arXiv:2510.12796},
  year={2025}
}

@article{wang2025alpamayo,
  title={Alpamayo-r1: Bridging reasoning and action prediction for generalizable autonomous driving in the long tail},
  author={Wang, Yan and Luo, Wenjie and Bai, Junjie and Cao, Yulong and Che, Tong and Chen, Ke and Chen, Yuxiao and Diamond, Jenna and Ding, Yifan and Ding, Wenhao and others},
  journal={arXiv preprint arXiv:2511.00088},
  year={2025}
}

@article{peng2025counterfactual,
  title={Counterfactual VLA: Self-Reflective Vision-Language-Action Model with Adaptive Reasoning},
  author={Peng, Zhenghao and Ding, Wenhao and You, Yurong and Chen, Yuxiao and Luo, Wenjie and Tian, Thomas and Cao, Yulong and Sharma, Apoorva and Xu, Danfei and Ivanovic, Boris and others},
  journal={arXiv preprint arXiv:2512.24426},
  year={2025}
}

@article{huang2026automot,
  title={Automot: A unified vision-language-action model with asynchronous mixture-of-transformers for end-to-end autonomous driving},
  author={Huang, Wenhui and Zhang, Songyan and Huang, Qihang and Wang, Zhidong and Mao, Zhiqi and Chua, Collister and Chen, Zhan and Chen, Long and Lv, Chen},
  journal={arXiv preprint arXiv:2603.14851},
  year={2026}
}

@article{qwen2.5,
    title   = {Qwen2.5 Technical Report}, 
    author  = {An Yang and Baosong Yang and Beichen Zhang and Binyuan Hui and Bo Zheng and Bowen Yu and Chengyuan Li and Dayiheng Liu and Fei Huang and Haoran Wei and Huan Lin and Jian Yang and Jianhong Tu and Jianwei Zhang and Jianxin Yang and Jiaxi Yang and Jingren Zhou and Junyang Lin and Kai Dang and Keming Lu and Keqin Bao and Kexin Yang and Le Yu and Mei Li and Mingfeng Xue and Pei Zhang and Qin Zhu and Rui Men and Runji Lin and Tianhao Li and Tingyu Xia and Xingzhang Ren and Xuancheng Ren and Yang Fan and Yang Su and Yichang Zhang and Yu Wan and Yuqiong Liu and Zeyu Cui and Zhenru Zhang and Zihan Qiu},
    journal = {arXiv preprint arXiv:2412.15115},
    year    = {2024}
}

@article{yang2025qwen3,
  title={Qwen3 technical report},
  author={Yang, An and Li, Anfeng and Yang, Baosong and Zhang, Beichen and Hui, Binyuan and Zheng, Bo and Yu, Bowen and Gao, Chang and Huang, Chengen and Lv, Chenxu and others},
  journal={arXiv preprint arXiv:2505.09388},
  year={2025}
}

@misc{qwen3.5,
    title  = {{Qwen3.5}: Towards Native Multimodal Agents},
    author = {{Qwen Team}},
    month  = {February},
    year   = {2026},
    url    = {https://qwen.ai/blog?id=qwen3.5}
}

@article{guo2025deepseek,
  title={Deepseek-r1: Incentivizing reasoning capability in llms via reinforcement learning},
  author={Guo, Daya and Yang, Dejian and Zhang, Haowei and Song, Junxiao and Wang, Peiyi and Zhu, Qihao and Xu, Runxin and Zhang, Ruoyu and Ma, Shirong and Bi, Xiao and others},
  journal={arXiv preprint arXiv:2501.12948},
  year={2025}
}

@inproceedings{chen2024internvl,
  title={Internvl: Scaling up vision foundation models and aligning for generic visual-linguistic tasks},
  author={Chen, Zhe and Wu, Jiannan and Wang, Wenhai and Su, Weijie and Chen, Guo and Xing, Sen and Zhong, Muyan and Zhang, Qinglong and Zhu, Xizhou and Lu, Lewei and others},
  booktitle={Proceedings of the IEEE/CVF conference on computer vision and pattern recognition},
  pages={24185--24198},
  year={2024}
}

@article{black2024pi_0,
  title={$\pi_0$: A Vision-Language-Action Flow Model for General Robot Control},
  author={Black, Kevin and Brown, Noah and Driess, Danny and Esmail, Adnan and Equi, Michael and Finn, Chelsea and Fusai, Niccolo and Groom, Lachy and Hausman, Karol and Ichter, Brian and others},
  journal={arXiv preprint arXiv:2410.24164},
  year={2024}
}

@article{intelligence2025pi_,
  title={$\pi_{0.5}$: A Vision-Language-Action Model with Open-World Generalization},
  author={Intelligence, Physical and Black, Kevin and Brown, Noah and Darpinian, James and Dhabalia, Karan and Driess, Danny and Esmail, Adnan and Equi, Michael and Finn, Chelsea and Fusai, Niccolo and others},
  journal={arXiv preprint arXiv:2504.16054},
  year={2025}
}

@article{vaswani2017attention,
  title={Attention is all you need},
  author={Vaswani, Ashish and Shazeer, Noam and Parmar, Niki and Uszkoreit, Jakob and Jones, Llion and Gomez, Aidan N and Kaiser, {\L}ukasz and Polosukhin, Illia},
  journal={Advances in neural information processing systems},
  volume={30},
  year={2017}
}

@article{liang2024mixture,
  title={Mixture-of-transformers: A sparse and scalable architecture for multi-modal foundation models},
  author={Liang, Weixin and Yu, Lili and Luo, Liang and Iyer, Srinivasan and Dong, Ning and Zhou, Chunting and Ghosh, Gargi and Lewis, Mike and Yih, Wen-tau and Zettlemoyer, Luke and others},
  journal={arXiv preprint arXiv:2411.04996},
  year={2024}
}

@article{deng2025emerging,
  title={Emerging properties in unified multimodal pretraining},
  author={Deng, Chaorui and Zhu, Deyao and Li, Kunchang and Gou, Chenhui and Li, Feng and Wang, Zeyu and Zhong, Shu and Yu, Weihao and Nie, Xiaonan and Song, Ziang and others},
  journal={arXiv preprint arXiv:2505.14683},
  year={2025}
}

@article{ho2022classifier,
  title={Classifier-free diffusion guidance},
  author={Ho, Jonathan and Salimans, Tim},
  journal={arXiv preprint arXiv:2207.12598},
  year={2022}
}

@article{ho2020denoising,
  title={Denoising diffusion probabilistic models},
  author={Ho, Jonathan and Jain, Ajay and Abbeel, Pieter},
  journal={Advances in neural information processing systems},
  volume={33},
  pages={6840--6851},
  year={2020}
}

@inproceedings{esser2024scaling,
  title={Scaling rectified flow transformers for high-resolution image synthesis},
  author={Esser, Patrick and Kulal, Sumith and Blattmann, Andreas and Entezari, Rahim and M{\"u}ller, Jonas and Saini, Harry and Levi, Yam and Lorenz, Dominik and Sauer, Axel and Boesel, Frederic and others},
  booktitle={Forty-first international conference on machine learning},
  year={2024}
}

@article{liu2022flow,
  title={Flow straight and fast: Learning to generate and transfer data with rectified flow},
  author={Liu, Xingchao and Gong, Chengyue and Liu, Qiang},
  journal={arXiv preprint arXiv:2209.03003},
  year={2022}
}

@article{xu2025wod,
  title={Wod-e2e: Waymo open dataset for end-to-end driving in challenging long-tail scenarios},
  author={Xu, Runsheng and Lin, Hubert and Jeon, Wonseok and Feng, Hao and Zou, Yuliang and Sun, Liting and Gorman, John and Tolstaya, Ekaterina and Tang, Sarah and White, Brandyn and others},
  journal={arXiv preprint arXiv:2510.26125},
  year={2025}
}

@article{song2020score,
  title={Score-based generative modeling through stochastic differential equations},
  author={Song, Yang and Sohl-Dickstein, Jascha and Kingma, Diederik P and Kumar, Abhishek and Ermon, Stefano and Poole, Ben},
  journal={arXiv preprint arXiv:2011.13456},
  year={2020}
}

@article{wei2022chain,
  title={Chain-of-thought prompting elicits reasoning in large language models},
  author={Wei, Jason and Wang, Xuezhi and Schuurmans, Dale and Bosma, Maarten and Xia, Fei and Chi, Ed and Le, Quoc V and Zhou, Denny and others},
  journal={Advances in neural information processing systems},
  volume={35},
  pages={24824--24837},
  year={2022}
}

@inproceedings{li2023blip,
  title={Blip-2: Bootstrapping language-image pre-training with frozen image encoders and large language models},
  author={Li, Junnan and Li, Dongxu and Savarese, Silvio and Hoi, Steven},
  booktitle={International conference on machine learning},
  pages={19730--19742},
  year={2023},
  organization={PMLR}
}

@inproceedings{papineni2002bleu,
  title={Bleu: a method for automatic evaluation of machine translation},
  author={Papineni, Kishore and Roukos, Salim and Ward, Todd and Zhu, Wei-Jing},
  booktitle={Proceedings of the 40th annual meeting of the Association for Computational Linguistics},
  pages={311--318},
  year={2002}
}

@inproceedings{lin2004rouge,
  title={Rouge: A package for automatic evaluation of summaries},
  author={Lin, Chin-Yew},
  booktitle={Text summarization branches out},
  pages={74--81},
  year={2004}
}

@inproceedings{rasley2020deepspeed,
  title={Deepspeed: System optimizations enable training deep learning models with over 100 billion parameters},
  author={Rasley, Jeff and Rajbhandari, Samyam and Ruwase, Olatunji and He, Yuxiong},
  booktitle={Proceedings of the 26th ACM SIGKDD international conference on knowledge discovery \& data mining},
  pages={3505--3506},
  year={2020}
}

@inproceedings{dosovitskiy2017carla,
  title={CARLA: An open urban driving simulator},
  author={Dosovitskiy, Alexey and Ros, German and Codevilla, Felipe and Lopez, Antonio and Koltun, Vladlen},
  booktitle={Conference on robot learning},
  pages={1--16},
  year={2017},
  organization={PMLR}
}

@article{dosovitskiy2020image,
  title={An image is worth 16x16 words: Transformers for image recognition at scale},
  author={Dosovitskiy, Alexey and Beyer, Lucas and Kolesnikov, Alexander and Weissenborn, Dirk and Zhai, Xiaohua and Unterthiner, Thomas and Dehghani, Mostafa and Minderer, Matthias and Heigold, Georg and Gelly, Sylvain and others},
  journal={arXiv preprint arXiv:2010.11929},
  year={2020}
}

@article{simeoni2025dinov3,
  title={Dinov3},
  author={Sim{\'e}oni, Oriane and Vo, Huy V and Seitzer, Maximilian and Baldassarre, Federico and Oquab, Maxime and Jose, Cijo and Khalidov, Vasil and Szafraniec, Marc and Yi, Seungeun and Ramamonjisoa, Micha{\"e}l and others},
  journal={arXiv preprint arXiv:2508.10104},
  year={2025}
}

@inproceedings{liu2021swin,
  title={Swin transformer: Hierarchical vision transformer using shifted windows},
  author={Liu, Ze and Lin, Yutong and Cao, Yue and Hu, Han and Wei, Yixuan and Zhang, Zheng and Lin, Stephen and Guo, Baining},
  booktitle={Proceedings of the IEEE/CVF international conference on computer vision},
  pages={10012--10022},
  year={2021}
}

@article{ma2025dvlm,
  title={dVLM-AD: Enhance diffusion vision-language-model for driving via controllable reasoning},
  author={Ma, Yingzi and Cao, Yulong and Ding, Wenhao and Zhang, Shuibai and Wang, Yan and Ivanovic, Boris and Jiang, Ming and Pavone, Marco and Xiao, Chaowei},
  journal={arXiv preprint arXiv:2512.04459},
  year={2025}
}

@article{wang2025hmvlm,
  title={HMVLM: Multistage reasoning-enhanced vision-language model for long-tailed driving scenarios},
  author={Wang, Daming and Song, Yuhao and He, Zijian and Chen, Kangliang and Pan, Xing and Deng, Lu and Gu, Weihao},
  journal={arXiv preprint arXiv:2506.05883},
  year={2025}
}

@article{touvron2023llama,
  title={Llama: Open and efficient foundation language models},
  author={Touvron, Hugo and Lavril, Thibaut and Izacard, Gautier and Martinet, Xavier and Lachaux, Marie-Anne and Lacroix, Timoth{\'e}e and Rozi{\`e}re, Baptiste and Goyal, Naman and Hambro, Eric and Azhar, Faisal and others},
  journal={arXiv preprint arXiv:2302.13971},
  year={2023}
}

@article{team2023gemini,
  title={Gemini: a family of highly capable multimodal models},
  author={Team, Gemini and Anil, Rohan and Borgeaud, Sebastian and Alayrac, Jean-Baptiste and Yu, Jiahui and Soricut, Radu and Schalkwyk, Johan and Dai, Andrew M and Hauth, Anja and Millican, Katie and others},
  journal={arXiv preprint arXiv:2312.11805},
  year={2023}
}

@article{you2025llada,
  title={Llada-v: Large language diffusion models with visual instruction tuning},
  author={You, Zebin and Nie, Shen and Zhang, Xiaolu and Hu, Jun and Zhou, Jun and Lu, Zhiwu and Wen, Ji-Rong and Li, Chongxuan},
  journal={arXiv preprint arXiv:2505.16933},
  year={2025}
}

@article{tschannen2025siglip,
  title={Siglip 2: Multilingual vision-language encoders with improved semantic understanding, localization, and dense features},
  author={Tschannen, Michael and Gritsenko, Alexey and Wang, Xiao and Naeem, Muhammad Ferjad and Alabdulmohsin, Ibrahim and Parthasarathy, Nikhil and Evans, Talfan and Beyer, Lucas and Xia, Ye and Mustafa, Basil and others},
  journal={arXiv preprint arXiv:2502.14786},
  year={2025}
}

@inproceedings{zitkovich2023rt,
  title={Rt-2: Vision-language-action models transfer web knowledge to robotic control},
  author={Zitkovich, Brianna and Yu, Tianhe and Xu, Sichun and Xu, Peng and Xiao, Ted and Xia, Fei and Wu, Jialin and Wohlhart, Paul and Welker, Stefan and Wahid, Ayzaan and others},
  booktitle={Conference on Robot Learning},
  pages={2165--2183},
  year={2023},
  organization={PMLR}
}

@article{team2024octo,
  title={Octo: An open-source generalist robot policy},
  author={Team, Octo Model and Ghosh, Dibya and Walke, Homer and Pertsch, Karl and Black, Kevin and Mees, Oier and Dasari, Sudeep and Hejna, Joey and Kreiman, Tobias and Xu, Charles and others},
  journal={arXiv preprint arXiv:2405.12213},
  year={2024}
}

@article{kim2024openvla,
  title={Openvla: An open-source vision-language-action model},
  author={Kim, Moo Jin and Pertsch, Karl and Karamcheti, Siddharth and Xiao, Ted and Balakrishna, Ashwin and Nair, Suraj and Rafailov, Rafael and Foster, Ethan and Lam, Grace and Sanketi, Pannag and others},
  journal={arXiv preprint arXiv:2406.09246},
  year={2024}
}

@article{liu2024rdt,
  title={Rdt-1b: a diffusion foundation model for bimanual manipulation},
  author={Liu, Songming and Wu, Lingxuan and Li, Bangguo and Tan, Hengkai and Chen, Huayu and Wang, Zhengyi and Xu, Ke and Su, Hang and Zhu, Jun},
  journal={arXiv preprint arXiv:2410.07864},
  year={2024}
}

@article{chi2025diffusion,
  title={Diffusion policy: Visuomotor policy learning via action diffusion},
  author={Chi, Cheng and Xu, Zhenjia and Feng, Siyuan and Cousineau, Eric and Du, Yilun and Burchfiel, Benjamin and Tedrake, Russ and Song, Shuran},
  journal={The International Journal of Robotics Research},
  volume={44},
  number={10-11},
  pages={1684--1704},
  year={2025},
  publisher={Sage Publications Sage UK: London, England}
}

@article{qu2025eo,
  title={Eo-1: Interleaved vision-text-action pretraining for general robot control},
  author={Qu, Delin and Song, Haoming and Chen, Qizhi and Chen, Zhaoqing and Gao, Xianqiang and Ye, Xinyi and Lv, Qi and Shi, Modi and Ren, Guanghui and Ruan, Cheng and others},
  journal={arXiv preprint arXiv:2508.21112},
  year={2025}
}

@inproceedings{sun2020scalability,
  title={Scalability in perception for autonomous driving: Waymo open dataset},
  author={Sun, Pei and Kretzschmar, Henrik and Dotiwalla, Xerxes and Chouard, Aurelien and Patnaik, Vijaysai and Tsui, Paul and Guo, James and Zhou, Yin and Chai, Yuning and Caine, Benjamin and others},
  booktitle={Proceedings of the IEEE/CVF conference on computer vision and pattern recognition},
  pages={2446--2454},
  year={2020}
}

@article{fang2025robix,
  title={Robix: A unified model for robot interaction, reasoning and planning},
  author={Fang, Huang and Zhang, Mengxi and Dong, Heng and Li, Wei and Wang, Zixuan and Zhang, Qifeng and Tian, Xueyun and Hu, Yucheng and Li, Hang},
  journal={arXiv preprint arXiv:2509.01106},
  year={2025}
}

@article{kaplan2020scaling,
  title={Scaling laws for neural language models},
  author={Kaplan, Jared and McCandlish, Sam and Henighan, Tom and Brown, Tom B and Chess, Benjamin and Child, Rewon and Gray, Scott and Radford, Alec and Wu, Jeffrey and Amodei, Dario},
  journal={arXiv preprint arXiv:2001.08361},
  year={2020}
}

@article{hoffmann2022training,
  title={Training compute-optimal large language models},
  author={Hoffmann, Jordan and Borgeaud, Sebastian and Mensch, Arthur and Buchatskaya, Elena and Cai, Trevor and Rutherford, Eliza and Casas, DDL and Hendricks, Lisa Anne and Welbl, Johannes and Clark, Aidan and others},
  journal={arXiv preprint arXiv:2203.15556},
  volume={10},
  year={2022}
}

@inproceedings{wang2023exploring,
  title={Exploring object-centric temporal modeling for efficient multi-view 3d object detection},
  author={Wang, Shihao and Liu, Yingfei and Wang, Tiancai and Li, Ying and Zhang, Xiangyu},
  booktitle={Proceedings of the IEEE/CVF international conference on computer vision},
  pages={3621--3631},
  year={2023}
}

@inproceedings{li2024think2drive,
  title={Think2drive: Efficient reinforcement learning by thinking with latent world model for autonomous driving (in carla-v2)},
  author={Li, Qifeng and Jia, Xiaosong and Wang, Shaobo and Yan, Junchi},
  booktitle={European conference on computer vision},
  pages={142--158},
  year={2024},
  organization={Springer}
}

@article{zhang20254d,
  title={4d-vla: Spatiotemporal vision-language-action pretraining with cross-scene calibration},
  author={Zhang, Jiahui and Chen, Yurui and Xu, Yueming and Huang, Ze and Zhou, Yanpeng and Yuan, Yu-Jie and Cai, Xinyue and Huang, Guowei and Quan, Xingyue and Xu, Hang and others},
  journal={arXiv preprint arXiv:2506.22242},
  year={2025}
}

@article{shi2025memoryvla,
  title={Memoryvla: Perceptual-cognitive memory in vision-language-action models for robotic manipulation},
  author={Shi, Hao and Xie, Bin and Liu, Yingfei and Sun, Lin and Liu, Fengrong and Wang, Tiancai and Zhou, Erjin and Fan, Haoqiang and Zhang, Xiangyu and Huang, Gao},
  journal={arXiv preprint arXiv:2508.19236},
  year={2025}
}

@article{zeng2025janusvln,
  title={Janusvln: Decoupling semantics and spatiality with dual implicit memory for vision-language navigation},
  author={Zeng, Shuang and Qi, Dekang and Chang, Xinyuan and Xiong, Feng and Xie, Shichao and Wu, Xiaolong and Liang, Shiyi and Xu, Mu and Wei, Xing and Guo, Ning},
  journal={arXiv preprint arXiv:2509.22548},
  year={2025}
}

@article{wei2025streamvln,
  title={Streamvln: Streaming vision-and-language navigation via slowfast context modeling},
  author={Wei, Meng and Wan, Chenyang and Yu, Xiqian and Wang, Tai and Yang, Yuqiang and Mao, Xiaohan and Zhu, Chenming and Cai, Wenzhe and Wang, Hanqing and Chen, Yilun and others},
  journal={arXiv preprint arXiv:2507.05240},
  year={2025}
}

@article{nguyen2026video,
  title={Video Understanding: Through A Temporal Lens},
  author={Nguyen, Thong Thanh},
  journal={arXiv preprint arXiv:2602.00683},
  year={2026}
}

@article{wu2023unleashing,
  title={Unleashing large-scale video generative pre-training for visual robot manipulation},
  author={Wu, Hongtao and Jing, Ya and Cheang, Chilam and Chen, Guangzeng and Xu, Jiafeng and Li, Xinghang and Liu, Minghuan and Li, Hang and Kong, Tao},
  journal={arXiv preprint arXiv:2312.13139},
  year={2023}
}

@article{chen2025intentionvla,
  title={IntentionVLA: Generalizable and Efficient Embodied Intention Reasoning for Human-Robot Interaction},
  author={Chen, Yandu and Gu, Kefan and Wen, Yuqing and Zhao, Yucheng and Wang, Tiancai and Nie, Liqiang},
  journal={arXiv preprint arXiv:2510.07778},
  year={2025}
}

@article{ma2025cdp,
  title={Cdp: Towards robust autoregressive visuomotor policy learning via causal diffusion},
  author={Ma, Jiahua and Qin, Yiran and Li, Yixiong and Liao, Xuanqi and Guo, Yulan and Zhang, Ruimao},
  journal={arXiv preprint arXiv:2506.14769},
  year={2025}
}

\newpage
\appendix

\section{Extended Related Work}
\label{sec:supp_related}

This section expands the related work discussion from \S\ref{sec:related}, surveying VLA methods primarily from the autonomous driving perspective along the design axes below. Although embodied intelligence operates in different observation modalities and action spaces, both domains face the same structural challenges. We believe insights developed in one domain can directly or indirectly inform the other; accordingly, citations below draw from both driving and embodied VLA literature. We acknowledge the inspiration that embodied intelligence research has provided to MindVLA-U1, and hope our work offers reciprocal insights to the embodied community.

\paragraph{Continuous action and open-vocabulary grounding.}
Tokenizing trajectories makes action compatible with an autoregressive language decoder but imposes a precision floor on spatial regression~\cite{sima2024drivelm,hwang2024emma,wang2025omnidrive,chi2025impromptu,zhou2025autovla,zeng2025futuresightdrive,yuan2025autodrive,rowe2025poutine,zitkovich2023rt,kim2024openvla}. 
Bypassing language recovers continuous control but gives up the explicit semantic representation that motivates VLA in the first place. Dual-head VLA designs~\cite{tian2024drivevlm,jiang2024senna,li2025recogdrive,liu2024rdt,black2024pi_0,intelligence2025pi_} avoid this false choice by sharing representation while preserving modality-native output spaces. MindVLA-U1 follows this principle with autoregressive language and continuous flow-matching action~\cite{chi2025diffusion,qu2025eo} on one backbone.

\paragraph{Language-action coupling.}
A recurring under-treated question in VLA-AD is whether the language reasoning produced by the VLM actually contributes to the action --- or whether the language head is a costly side-task the action head learns to ignore. In broader robot learning, goal-conditioned and intent-conditioned policies~\cite{zitkovich2023rt,wu2023unleashing,chen2025intentionvla} provide mechanistic precedents for explicit language-to-action routing; in the driving VLA setting, however, an explicit and measurable bridge from language-side state to continuous action has not yet been demonstrated. Prior driving-VLA work typically reports a single planning number with language enabled, without ablating against parallel or action-only variants that isolate the contribution of the causal language-to-action conditioning~\cite{wang2025alpamayo,peng2025counterfactual}. The result is that the field cannot distinguish ``language helped planning'' from ``the model with language got better numbers for unrelated reasons.''

\paragraph{Fast/Slow Execution System.}
VLA backbones introduce inference latency incompatible with real-time control, yet most driving moments are routine reflex needing no deliberation --- making fast/slow separation a deployment-level necessity. Prior approaches fall into two categories, each with structural limitations. 
The \emph{cascaded dual-module pipeline}~\cite{tian2024drivevlm,jiang2024senna,li2025recogdrive,black2024pi_0,intelligence2025pi_,fang2025robix} delegates reasoning to a separate VLM whose discrete outputs feed a downstream planner; this severs end-to-end gradient flow and adds serial latency as an unavoidable bottleneck. \emph{Heuristic routing}~\cite{zhang2025adadrive,luo2025adathinkdrive,zhou2025autovla} uses model inference outputs to decide per frame whether to invoke full deliberation or bypass it; the routing decision lacks distributional robustness and provides no architectural guarantee that the fast path produces valid actions. Deployment-grade fast/slow separation must instead be a structural property of the architecture itself: MoT~\cite{liang2024mixture,deng2025emerging,huang2026automot} embeds modality-specific attention groups and feed-forward experts directly in the forward pass, making the fast/slow boundary an architectural invariant rather than a runtime heuristic (\S\ref{sec:unified}).

\paragraph{Action chunking and temporal context modeling.}
Most published VLA systems predict action chunks and stitch them at clip boundaries, accumulating seam error and breaking the continuity that physical control requires. Because driving is inherently temporally causal --- intent, momentum, and scene evolution carry across observation boundaries --- the training paradigm should satisfy three requirements: continuously renewed memory propagation that mirrors the sequential nature of the task, deployment-friendly bounded compute, and end-to-end joint optimization of the memory component with the VLA model. We adopt framewise streaming memory propagation as this paradigm (\S\ref{sec:streaming_memory}). Prior approaches either train on independent chunks without cross-boundary propagation~\cite{zhou2025autovla,yuan2025autodrive,rowe2025poutine,renz2025simlingo,luo2025adathinkdrive,li2025recogdrive,wang2025alpamayo,peng2025counterfactual,huang2026automot,zitkovich2023rt,kim2024openvla,black2024pi_0,intelligence2025pi_}, or fail to properly design memory propagation and jointly optimize it with the VLA model~\cite{wang2023exploring,li2024think2drive,fu2025orion,zhang2025adadrive,team2024octo,zhang20254d,shi2025memoryvla,qu2025eo,zeng2025janusvln,wei2025streamvln} --- remaining a suboptimal paradigm choice.

\paragraph{Driving-native VLA supervision.}
Standard driving datasets present two gaps that generic VLA training cannot close. First, logged trajectories teach control but not language-grounded scene reasoning: a model trained only to reproduce one trajectory per scene receives no explicit supervision for spatial cognition or planning-relevant interpretation. Second, one logged trajectory does not expose the quality spectrum of possible behaviors the scene admits --- standard imitation learning models a single mode from a multi-modal safe-action distribution. 
The literature has explored perturbation-based and exemplar-based trajectory augmentation~\cite{ma2025cdp,renz2025simlingo}, but the open question is how to ground synthesized diversity in scene-specific intent so that it is semantically meaningful rather than arbitrary, and how to convert that diversity into a learning signal that distinguishes good driving from poor. MindLabel addresses both gaps in a unified pipeline: scene-understanding QA binds language to the same frames used for action supervision, while intent-conditioned trajectory generation expands behavioral coverage with semantically grounded alternatives.

\paragraph{Backbone scale.}
Despite the rapid progress of language and vision-language model scaling laws~\cite{kaplan2020scaling,hoffmann2022training}, current driving VLA results do not yet establish an obvious monotonic VLM scaling law for planning~\cite{shao2024lmdrive,tian2024drivevlm,rowe2025poutine}. 
Backbone scale remains important, but in this setting it is entangled with optimization stability, memory pressure, data mixture, and the action interface~\cite{kim2024openvla,black2024pi_0,intelligence2025pi_}. We characterize the regime --- including a structural decoupling of language quality and action quality --- in \S\ref{sec:backbone_scaling}.

\paragraph{Simulator-bound evaluation.}
A practical shortcoming that cuts across the above is the heavy reliance on simulated benchmarks --- particularly CARLA~\cite{dosovitskiy2017carla} --- in academic VLA-AD work. Simulator visual realism, agent behavior, and long-tail composition do not match on-road conditions; closed-loop scores reported on simulator benchmarks routinely fail to predict real-world performance. The field has begun to migrate to real-world benchmarks such as Waymo Open Dataset End-to-End~\cite{sun2020scalability,xu2025wod}, but most published VLA results remain simulator-anchored. MindVLA-U1 is evaluated entirely on real-world WOD-E2E.

\section{MindLabel Dataset}
\label{sec:supp_mindlabel}

MindLabel supplies the driving-native language and trajectory-diversity supervision used by MindVLA-U1. Its design is motivated by two gaps in logged trajectory imitation. First, a model trained only to reproduce one
logged trajectory per scene receives no explicit supervision for scene cognition: perception of complex layouts, spatiotemporal reasoning about agents, and planning-relevant interpretation. Second, one logged trajectory
does not expose the quality spectrum of possible behaviors in the same scene. MindLabel addresses both gaps by generating scene-understanding QA and trajectory-evaluation QA from the same driving frames used for action
supervision.

\begin{figure*}[ht!]
  \centering
  \includegraphics[width=0.9\linewidth]{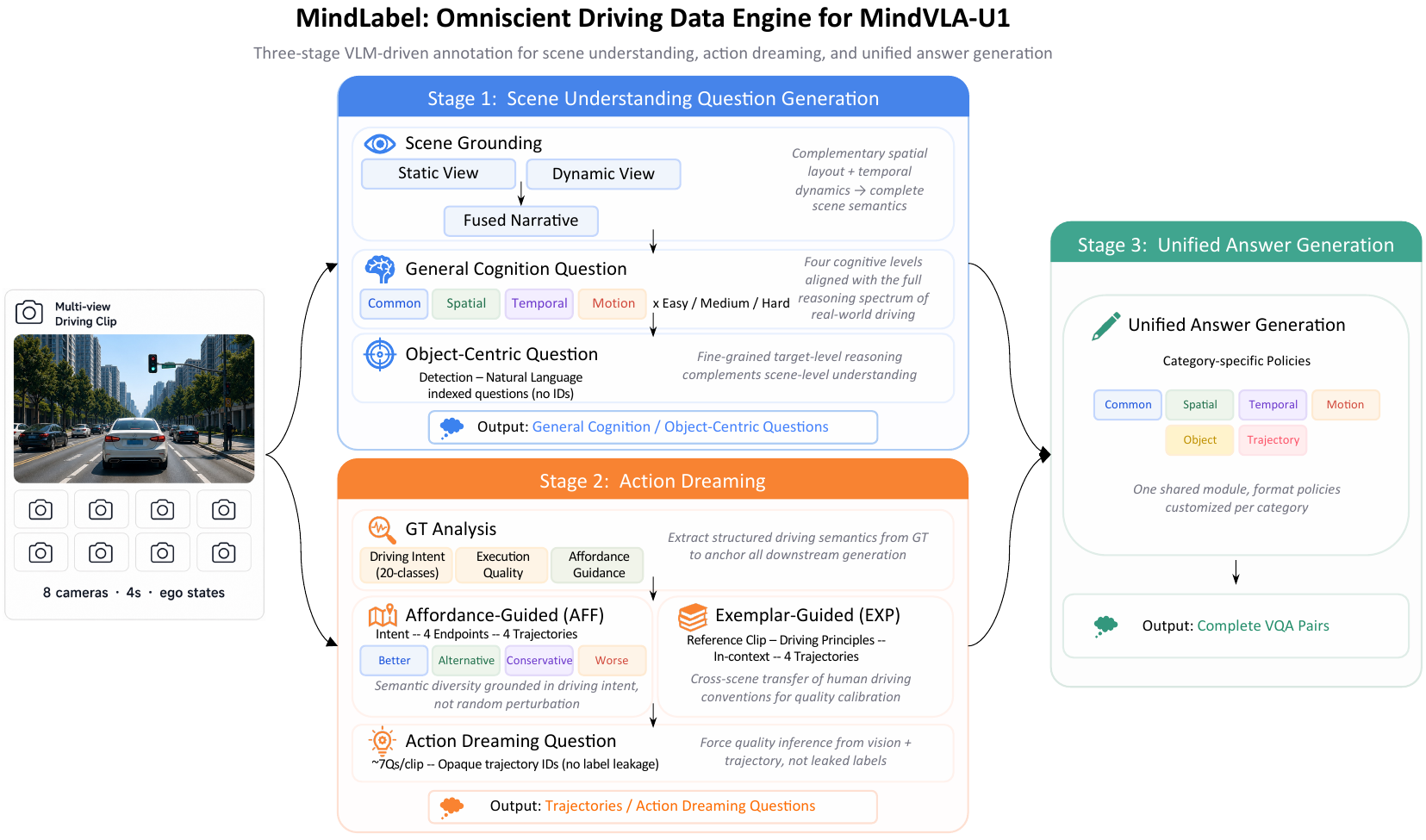}
  \caption{\textbf{MindLabel pipeline overview.} Scene Understanding Question Generation and Action Dreaming run in parallel on each driving frame, producing complementary question sets that are jointly answered by a unified answer-generation module with category-specific policies.}
  \label{fig:mindlabel_pipeline}
\end{figure*}

\begin{figure*}[ht!]
  \centering
  \includegraphics[width=0.9\linewidth]{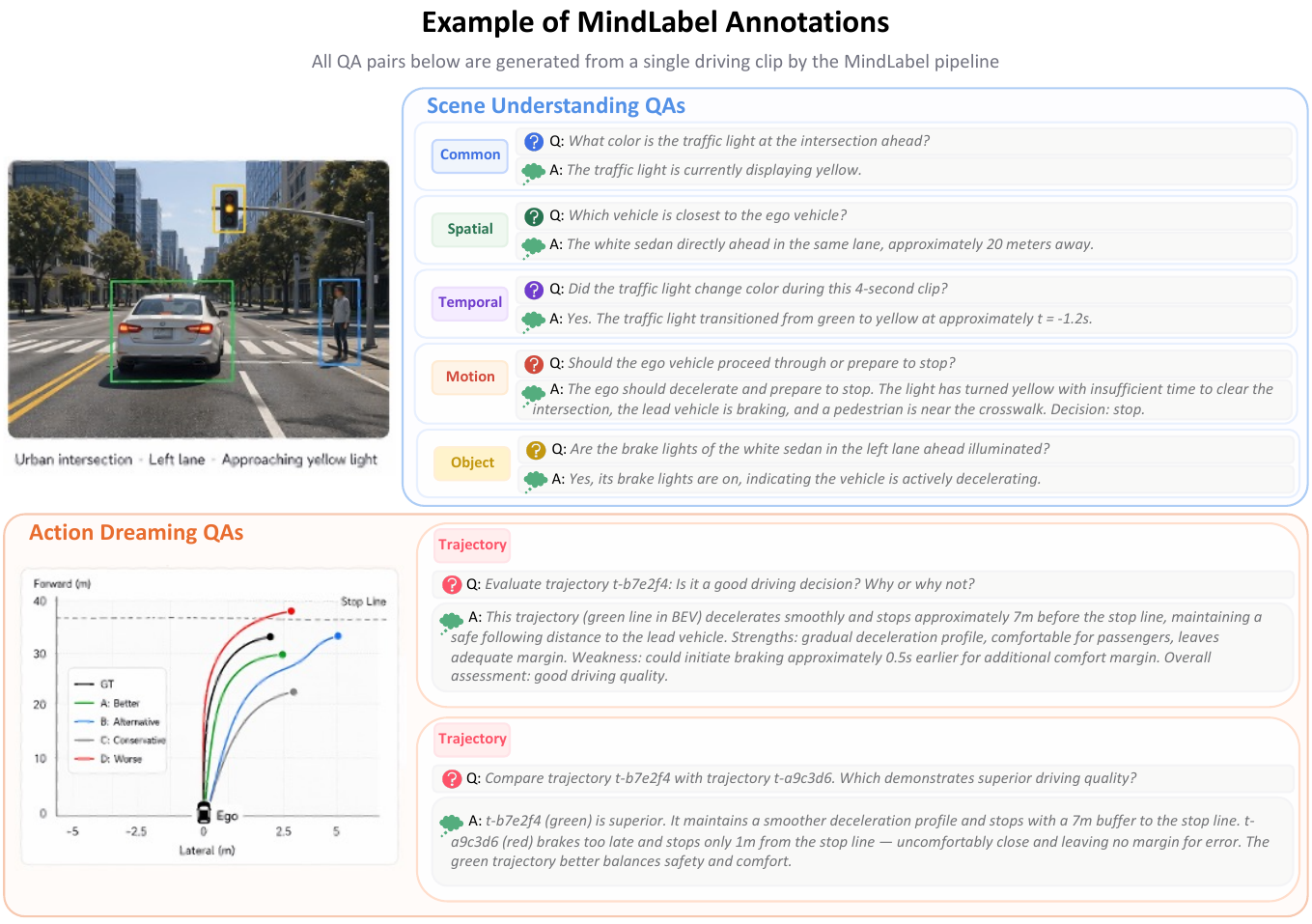}
  \caption{\textbf{Example MindLabel annotations from a single driving frame.} The pipeline produces scene-understanding QA pairs across five categories (Common, Spatial, Temporal, Motion, Object-Centric) and action-dreaming QA pairs that evaluate synthesized trajectories using opaque identifiers.}
  \label{fig:mindlabel_example}
\end{figure*}

\begin{figure*}[ht!]
\centering
\includegraphics[width=0.75\linewidth]{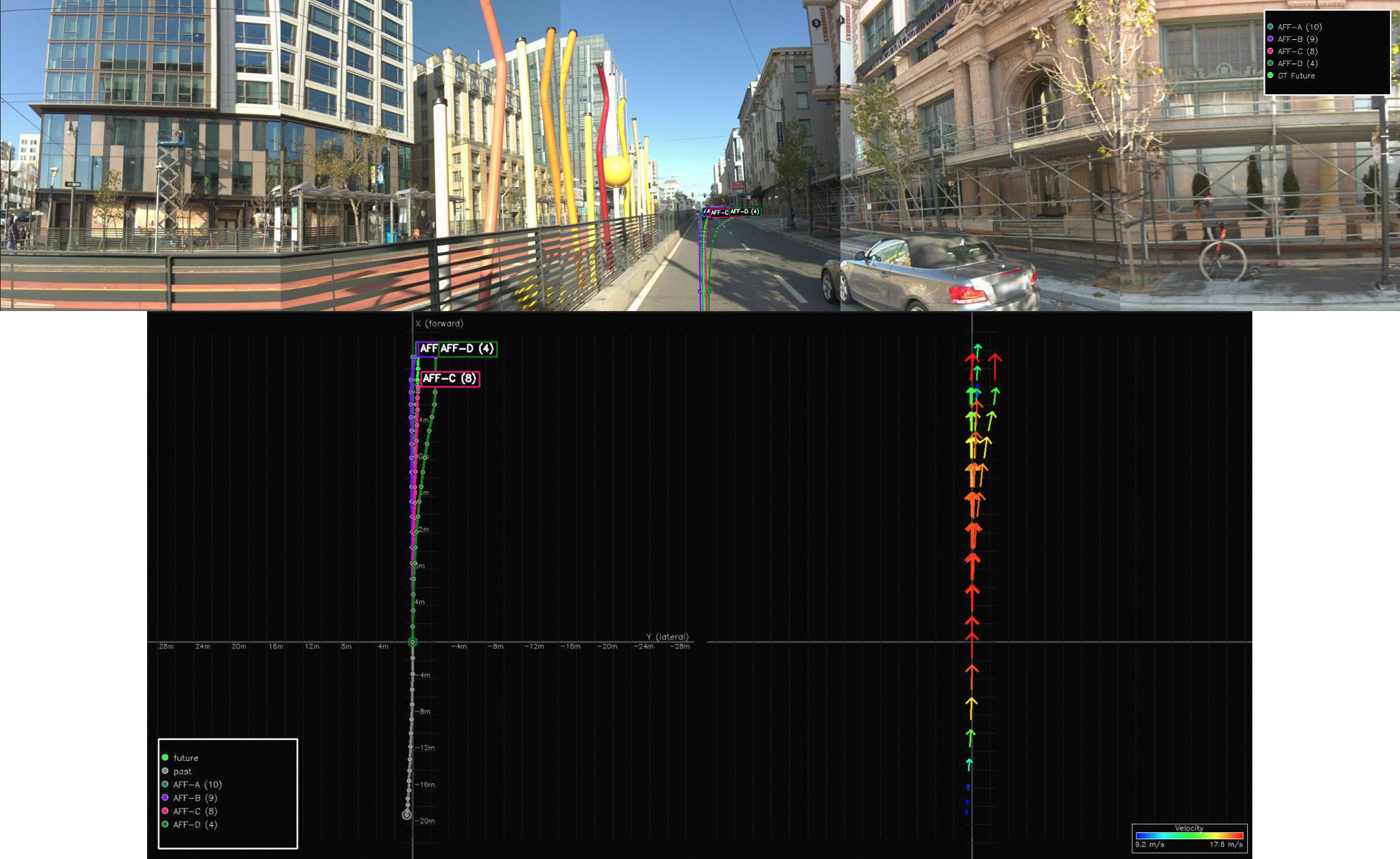}\\[-0.1em]
{\footnotesize (a) Daytime urban: panorama (top) + BEV (bottom).}\\[0.5em]
\includegraphics[width=0.75\linewidth]{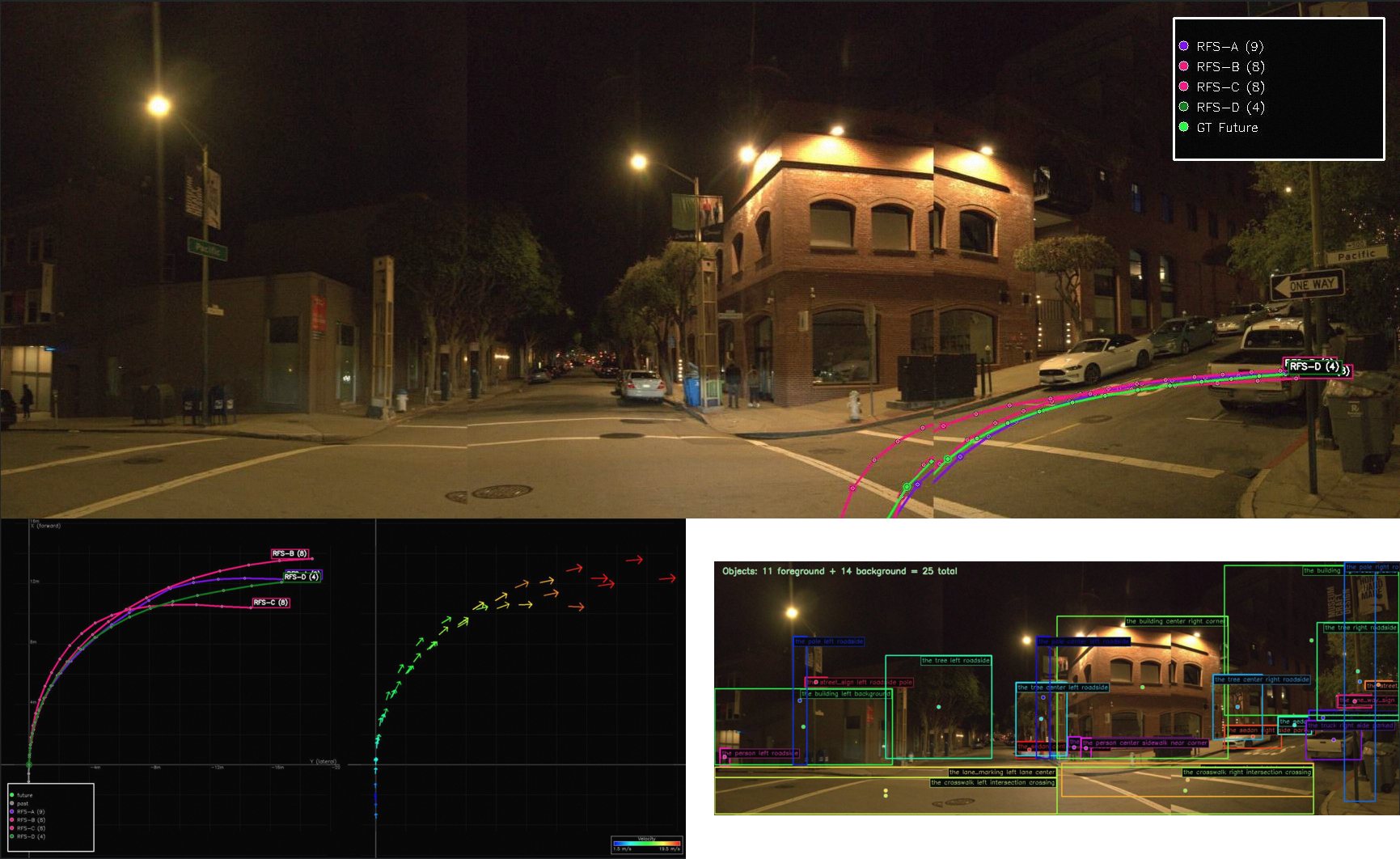}\\[-0.1em]
{\footnotesize (b) Nighttime intersection: panorama (top) + BEV + Object-Centric (bottom).}
\caption{\textbf{MindLabel annotations on real WOD-E2E frames.} Two example scenes stacked vertically. In each panel, the front-camera panorama overlays dreamed trajectories (four AFF candidates A--D from \S\ref{sec:supp_mindlabel_dreaming} plus the GT future, color-coded by RFS quality) and the BEV view shows trajectories with per-step motion vectors. Scene B additionally exposes the Object-Centric annotation pass (\S\ref{sec:supp_mindlabel_scene}): $25$ bounding boxes ($11$ foreground / $14$ background) each grounded by a natural-language descriptor (\eg \emph{``the car in front''}, \emph{``intersection right corner''}).}
\label{fig:mindlabel_visual_examples}
\end{figure*}

These two motivations give rise to three pipeline stages (Figure~\ref{fig:mindlabel_pipeline}): \textbf{Scene Understanding Question Generation} (\S\ref{sec:supp_mindlabel_scene}), which constructs multi-level cognitive
questions from driving video; \textbf{Action Dreaming} (\S\ref{sec:supp_mindlabel_dreaming}), which synthesizes diverse trajectories and trajectory-evaluation questions grounded in GT driving intent; and \textbf{Unified
Answer Generation} (\S\ref{sec:supp_mindlabel_answer}), which produces answers for all question types through a shared module with category-specific policies. Figure~\ref{fig:mindlabel_example} shows a concrete
annotation example from a single driving frame; Figure~\ref{fig:mindlabel_visual_examples} shows real WOD-E2E frames overlaid with the dreamed-trajectory and object-centric annotation outputs.

\subsection{Scene Understanding Question Generation}
\label{sec:supp_mindlabel_scene}

This stage generates structured questions across multiple levels of scene understanding in three steps.

\paragraph{Scene Grounding.}
The pipeline fuses a \textit{static} view of the most recent frame's spatial layout with a \textit{dynamic} view of the preceding 4-second history window into a single free-form natural-language scene description, intentionally without metric coordinates or structured object lists; precise per-target attributes are handled by Object-Centric Question Generation below.

\paragraph{General Cognition Question Generation.}
Building on the scene narrative, the pipeline generates questions across four cognitive categories: \textit{Common} (factual scene perception, \eg \textit{``What color is the traffic light ahead?''}), \textit{Spatial} (relative positions and distances), \textit{Temporal} (sequential changes and causal reasoning), and \textit{Motion} (safety analysis and planning decisions, \eg \textit{``Should the ego vehicle proceed, yield, or stop?''}). Each question is independently assigned a difficulty level (Easy/Medium/Hard) and a chain-of-thought flag indicating whether multi-step reasoning is required.

\paragraph{Object-Centric Question Generation.}
The pipeline performs unified detection across vehicles, pedestrians, cyclists, traffic signs, and signals in the current frame, and generates questions anchored to each detected instance, referencing it through a natural-language description (\eg \textit{``the red truck in the left lane approximately 15 meters ahead''}) rather than a detection ID. Question types span object existence, visual attributes, OCR on traffic signs, signal state classification, and counting queries.

\subsection{Action Dreaming: Diverse Trajectory Construction}
\label{sec:supp_mindlabel_dreaming}

Standard imitation from a single GT trajectory per frame leaves two structural gaps: the model sees only one behavioral realization per scenario, and has no signal about what constitutes suboptimal behavior. Action Dreaming addresses both by generating a diverse set of synthetic trajectories per frame, conditioned on GT-extracted driving semantics, plus matching trajectory-evaluation questions.

\paragraph{GT Analysis.}
For each frame we extract from the ground-truth future trajectory: (1) a $20$-class driving intent label spanning longitudinal maneuvers (\eg cruising, decelerating), lateral maneuvers (\eg lane change, left turn), and complex cases (\eg U-turn, emergency stop); (2) an execution quality assessment of the GT trajectory; and (3) affordance guidance describing scene-specific opportunities and constraints. This anchors dreamed trajectories to actual driving behavior rather than arbitrary perturbation.

\paragraph{Affordance-Guided Generation (AFF).}
Conditioned on the GT-derived intent, the pipeline generates four \textit{affordance endpoints} corresponding to qualitatively distinct executions of the same intent (Better / Alternative / Conservative / Worse), each completed into a full 5-second trajectory using the GT trajectory's curvature as a shape reference. This intent-conditioned, geometry-consistent design ensures the trajectory set spans the quality spectrum relevant to the current scene while remaining semantically grounded.

\paragraph{Exemplar-Guided Generation (EXP).}
EXP retrieves a \textit{reference frame} from the same driving sequence that carries existing human preference annotations, extracts driving principles from it (recommended behaviors, patterns to avoid), and injects them as in-context guidance during trajectory generation, transferring human-preference-derived driving conventions into the generated set.

\paragraph{Action Dreaming Question Generation.}
The generated trajectories are converted into trajectory-evaluation questions: per frame, four fixed questions ask the model to evaluate each dreamed trajectory's quality (\eg \textit{``Evaluate this trajectory: Is it a good driving decision? Why or why not?''}), plus three sampled questions covering reference-trajectory intent, comparative ranking among dreamed trajectories, and pairwise comparisons. Each trajectory is referenced by an opaque identifier rather than its quality label --- the model must infer quality from the driving video and the BEV waypoint sequence, without any direct signal that a trajectory is labeled ``Better'' or ``Worse.''

\subsection{Unified Answer Generation}
\label{sec:supp_mindlabel_answer}

All questions from the preceding stages are answered by a shared module with category-specific formatting policies: \textit{Common} produces concise factual responses, \textit{Spatial} adds explicit units and relative positions, \textit{Temporal} cites time points and causal relationships, and \textit{Motion} concludes with an actionable driving recommendation (\eg \textit{proceed}, \textit{yield}, \textit{stop}). Trajectory-evaluation answers are conditioned on the driving video and the BEV waypoint sequence of each trajectory, producing chain-of-thought reasoning over driving quality. Questions carrying the chain-of-thought flag receive multi-step inference chains with cited evidence.

\subsection{Dataset Statistics}
\label{sec:supp_mindlabel_stats}

Table~\ref{tab:supp_mindlabel_stats} reports the total MindLabel annotation scale across the full WOD-E2E (training $+$ validation $+$ test). Each clip is annotated independently by both backbones (Qwen3-VL and Qwen3.5-Plus), and the totals below count VQA pairs and dreamed trajectories \emph{aggregated across both backbones' outputs} on the same clip set. Per-split clip counts on training and validation are measured from the labeling-pipeline runs (front view, $2$\,Hz over a $4$\,s window); the test clip count uses the WOD-E2E test split's $1{,}505$ sequences each annotated at $3$ clips per sequence (the released end frame plus its two preceding clips, matching the WOD-E2E test protocol). The aggregate per-clip yield observed in the partial run is $\sim$$200$ VQA pairs and $\sim$$13$--$14$ dreamed trajectories (combined AFF $+$ EXP at $\sim$$95\%$ frame coverage); the two backbones contribute roughly equally to this aggregate but \emph{not} exactly $1{:}1$, since each backbone makes its own decisions about how many spatial/temporal/motion follow-ups to emit per scene. Totals are extrapolated from this measured aggregate rate rather than from a single backbone doubled. Valid-QA category proportions are likewise extrapolated from the partial run and applied to the full-scale aggregate. MindVLA-U1's main result consumes only the basic scene-grounded VQA stream and the GT $3$-class intent label; the rest --- dreamed trajectories, GT/dreamed trajectory-evaluation QAs, GT commentary, $20$-class intent, chain-of-thought rationales --- is released as part of MindLabel to support broader driving-VLA tasks (preference learning, trajectory ranking, reasoning, world-model conditioning) in future work.

\begin{table}[h]
\caption{Total MindLabel annotation scale across the full WOD-E2E benchmark (training $+$ validation $+$ test). Each clip is independently annotated by both backbones (Qwen3-VL and Qwen3.5-Plus); VQA and dreamed-trajectory totals are aggregate counts \emph{across both backbones' outputs}, extrapolated from the measured aggregate per-clip yield in the partial run ($\sim$$200$ VQAs/clip; $\sim$$13$--$14$ dreamed/clip at $\sim$$95\%$ coverage). Per-backbone contributions are approximately $1{:}1$ but not exactly so.}
\label{tab:supp_mindlabel_stats}
\centering
\footnotesize
\setlength{\tabcolsep}{6pt}
\begin{tabular}{lrrrr}
\toprule
\textbf{Statistic} & \textbf{Training} & \textbf{Validation} & \textbf{Test} & \textbf{Total} \\
\midrule
Sequences (WOD-E2E)                                  & $2{,}037$    & $479$         & $1{,}505$    & $4{,}021$    \\
Annotated clips (per backbone)                       & $10{,}542$   & $3{,}771$     & $4{,}515$    & $18{,}828$   \\
\midrule
Total VQA pairs                                      & $\sim$$2.1$M  & $\sim$$750$K & $\sim$$900$K & $\boldsymbol{\sim 3.8\text{M}}$ \\
\quad Common ($\sim$$11.3\%$)                        & ---           & ---           & ---           & $\sim$$430$K  \\
\quad Spatial ($\sim$$13.6\%$)                       & ---           & ---           & ---           & $\sim$$520$K  \\
\quad Temporal ($\sim$$14.7\%$)                      & ---           & ---           & ---           & $\sim$$560$K  \\
\quad Motion ($\sim$$14.7\%$)                        & ---           & ---           & ---           & $\sim$$560$K  \\
\quad Object-Centric ($\sim$$11.4\%$)                & ---           & ---           & ---           & $\sim$$430$K  \\
\quad Trajectory-evaluation ($\sim$$7.5\%$)          & ---           & ---           & ---           & $\sim$$285$K  \\
\quad Chain-of-thought, cross-category ($\sim$$26.8\%$) & ---        & ---           & ---           & $\sim$$1.0$M  \\
\midrule
Total dreamed trajectories ($\sim$$95\%$ cov.)        & $\sim$$140$K & $\sim$$50$K  & $\sim$$60$K  & $\boldsymbol{\sim 250\text{K}}$ \\
\midrule
Annotation backbones                                 & \multicolumn{4}{c}{Qwen3-VL $+$ Qwen3.5-Plus}                          \\
\bottomrule
\end{tabular}
\end{table}

\subsection{Val/Test Intent Distribution Shift}
\label{sec:supp_mindlabel_dist_shift}

This section provides the full $15$-intent distribution underlying the val/test diagnostic in main text \S\ref{sec:val_test_diagnostic} (which shows only the four most-shifted intents in Table~\ref{tab:intent_dist_short}). Table~\ref{tab:supp_intent_dist} reports the per-intent count and share on val (anchored to the RFS-evaluated frames) and test (end-frame protocol). Beyond the top movers, several other slowdown / yielding categories (\emph{decelerating}, \emph{following}, \emph{yielding}, \emph{lane changes}) are markedly over-represented on test, reinforcing the diagnostic in the main text: the val-to-test RFS drop seen across all methods (e.g.\ MindVLA-U1's $-0.24$~RFS) tracks a structural, dataset-level shift in intent mix, concentrated on the under-represented \emph{accelerate / start / turn-right} clusters and on the over-represented \emph{waiting} cluster.

\begin{table}[h]
\caption{MindLabel-derived intent distribution on WOD-E2E val and test. Per-intent count $n$ and share (\%); sorted by descending val share. \textbf{Bold} = dominant intent on each split.}
\label{tab:supp_intent_dist}
\centering
\footnotesize
\setlength{\tabcolsep}{6pt}
\begin{tabular}{l|rr|rr}
\toprule
                       & \multicolumn{2}{c|}{Val (RFS-anchored)} & \multicolumn{2}{c}{Test (end-frame)} \\
\cmidrule(lr){2-3}\cmidrule(lr){4-5}
Intent                 & $n$ & \% & $n$ & \% \\
\midrule
accelerating           & 93  & \textbf{19.54} &  62 &  4.12 \\
cruising               & 65  & 13.66          & 230 & 15.29 \\
turning\_right         & 58  & 12.18          &  86 &  5.72 \\
starting               & 50  & 10.50          &  68 &  4.52 \\
stopping               & 42  &  8.82          & 113 &  7.51 \\
avoiding\_obstacle     & 38  &  7.98          &  37 &  2.46 \\
waiting                & 36  &  7.56          & 304 & \textbf{20.21} \\
turning\_left          & 25  &  5.25          &  75 &  4.99 \\
decelerating           & 21  &  4.41          &  87 &  5.78 \\
following              & 13  &  2.73          &  84 &  5.59 \\
lane\_change\_right    &  8  &  1.68          &  38 &  2.53 \\
braking                &  8  &  1.68          &  16 &  1.06 \\
lane\_change\_left     &  6  &  1.26          &  36 &  2.39 \\
yielding               &  5  &  1.05          &  35 &  2.33 \\
u\_turn                &  5  &  1.05          &   3 &  0.20 \\
\bottomrule
\end{tabular}
\end{table}

\section{Implementation Setup and Backbone Scaling}
\label{sec:supp_impl}

\subsection{Architectural and Optimization Details}
\label{sec:supp_setup}

This section provides the dimensional and optimization detail deferred from \S\ref{sec:setup}.

\paragraph{Network architecture.}
MindVLA-U1 uses Qwen3-VL-2B as the default VLM backbone (hidden size $H{=}2048$). The vision encoder is frozen; the visual merger, the language model, ego-history encoders, the streaming memory module, and the action head are jointly trained. Ego-history is encoded by three lightweight MLPs (one each for position, velocity, and acceleration) consuming $16$ historical states sampled at $2$\,Hz. The propagation transformer reads/writes $N_m{=}128$ memory tokens per frame via $6$ cross-attention layers with $16$ heads; the FIFO memory channel holds $N_g{=}2$ frame-step entries (total capacity $N_g \cdot N_m {=} 256$ tokens). The action head is a $2$-layer MLP with SiLU activation that predicts a $6$-dimensional output (position $+$ velocity $+$ acceleration) over $L_f{=}20$ future waypoints at $4$\,Hz (5-second horizon), using $2$ Euler integration steps at inference. Backbone-size results and the language-action decoupling analysis are in \S\ref{sec:backbone_scaling}.

\paragraph{Implementation details.}
We optimize with AdamW ($\mathrm{lr}{=}10^{-4}$, weight decay $0.1$, $\beta_1{=}0.9$, $\beta_2{=}0.999$) under a linear-warmup cosine-annealed schedule (warmup $1{,}000$ iterations, $\eta_\mathrm{min}{=}0.1\!\cdot\!\eta_\mathrm{max}$), trained for $50$ epochs (\S\ref{sec:setup}). Mixed-precision BF16 with DeepSpeed ZeRO-2 is used across $8$ GPUs. The flow-matching loss applies per-component weights $w_\mathrm{pos}{=}1.0$, $w_\mathrm{vel}{=}0.5$, $w_\mathrm{acc}{=}0.5$. The action representation uses a delta-position channel: future positions are predicted as incremental displacements while velocity and acceleration remain auxiliary absolute channels.

\subsection{VLM Backbone Scaling (Moved)}
\label{sec:supp_scaling}

The VLM backbone scaling study and Table~\ref{tab:abl_scaling} are now in main text \S\ref{sec:backbone_scaling}.

\section{Intent-CFG: Implementation Details}
\label{sec:supp_intent_cfg}

This section documents the three conditioning sources compared in Table~\ref{tab:language_helps_action} and the CFG mechanism that mixes their conditional and unconditional velocity fields.

\paragraph{Conditioning sources.}
Three intent signals are evaluated against the no-intent baseline. \emph{Trajectory-derived} extracts an intent class from the GT trajectory geometry. \emph{GT-supplied} uses the WOD-E2E raw intent $z \in \{\text{left}, \text{right}, \text{straight}\}$ label directly --- an oracle that decouples CFG-mechanism quality from intent-prediction quality. \emph{NTP-predicted} (the deployed primary) decodes the intent token from the language head via standard next-token prediction on the same scenes that supply the action labels (\S\ref{sec:semantic_intent}), and is paired with the prototype-grounded refinement of the intent embedder referenced in main text.

\paragraph{Embedding and CFG conditioning.}
An embedding table stores one row per intent class plus a learned unconditional row $\emptyset$. The selected row is projected and added residually to the action MLP's time embedding, preserving the joint AR$+$FM forward pass of \S\ref{sec:unified}. At training, the conditioning intent is replaced by $\emptyset$ with probability $p_{\text{drop}}{=}0.15$, so the same parameters learn both conditional and unconditional velocity fields. At inference, two backbone passes are run --- one with the predicted $z$ and one with $\emptyset$ --- and their velocity predictions are mixed at every Euler step using Eq.~\ref{eq:intent_cfg} with guidance scale $s{=}1.5$.

\section{Streaming Memory and Long-Sequence Training}
\label{sec:supp_streaming}

\subsection{Architectural Details}
\label{sec:supp_streaming_arch}

This section gives the architectural details deferred from \S\ref{sec:streaming_memory}. Three coupled components produce the memory feature $\mathbf{m}_i \in \mathbb{R}^{N_m \times H}$ (Figure~\ref{fig:memory_bank}):
a FIFO memory channel, a motion-aware modulator, and a propagation transformer.

\paragraph{FIFO memory channel.}
The memory channel $\mathcal{M}$ is a bounded FIFO buffer holding at most $N_g$ frame-step entries, each consisting of $N_m$ tokens in $\mathbb{R}^H$, for total capacity $N_g \times N_m$ tokens. When full, the oldest entry
is evicted, keeping per-step cost constant.

\paragraph{Motion-aware modulation.}
Each entry in $\mathcal{M}$ is stored in the ego coordinate system of the frame-step that produced it. Before reading, each historical entry $j$ is re-expressed in the current step $i$'s ego coordinate system via the
relative SE(2) transform:
\begin{equation}
  T_{j \to i} = P_i \cdot P_j^{-1},
  \label{eq:se2_warp}
\end{equation}
where $P_i, P_j \in \mathrm{SE}(2)$ are vehicle poses extracted from ego-states $\mathbf{e}_i, \mathbf{e}_j$. From $T_{j \to i}$ we form a 5-dimensional feature: rotation $(\cos\phi, \sin\phi)$, translation $(\delta_x,
\delta_y)$, and a normalized temporal offset. A lightweight MLP maps this feature to a modulation vector in $\mathbb{R}^H$, added elementwise to all $N_m$ tokens of entry $j$, yielding the modulated channel contents
$\tilde{\mathcal{M}}_i$.

\paragraph{Propagation transformer.}
$N_m$ learnable query vectors cross-attend to $\tilde{\mathcal{M}}_i$ via a Q-Former-style transformer, producing $\mathbf{m}_i$, which is prepended to the current frame's input sequence. After the backbone forward pass,
the same transformer symmetrically compresses the backbone outputs $\mathbf{h}_i$ into $N_m$ tokens that are written back to $\mathcal{M}$ for subsequent frame-steps. Gradients flow through the propagation transformer across stream steps, so
step-$i$ losses directly supervise the memory written at step $i{-}1$ (\S\ref{sec:streaming_memory}). An empirical comparison against the alternative of feeding multi-frame video directly into the VLM is given in \S\ref{sec:ablations_streaming} (Table~\ref{tab:abl_masmp}, $^\flat$ row).

\subsection{Full-Sequence Pose Recovery for Streaming Training}
\label{sec:supp_pose_recovery}

WOD-E2E releases each driving segment as a sequence of $\sim$$5$-second clips, with each clip's trajectory expressed in its own local ego frame. To run streaming training over full segments rather than isolated clips, MindVLA-U1 first recovers a global pose chain that stitches consecutive clips into a single ego-anchored coordinate system, so the streaming memory channel and the per-frame ego-state encoder see consistent positions across the whole segment.

\paragraph{SE(2) pose alignment.}
At each clip boundary, the relative SE(2) transform between two adjacent clips' ego frames is estimated by aligning the overlapping waypoints (the tail of clip $j$ with the head of clip $j{+}1$) under a rigid $2$D rotation $+$ translation. The resulting $T_{(j+1) \to j} \in \mathrm{SE}(2)$ takes coordinates from clip $j{+}1$'s frame into clip $j$'s frame; cumulative composition along the clip chain expresses every clip's local trajectory in a single global frame anchored to the segment's first clip. This is the same SE(2) algebra used inside the streaming memory channel for motion-aware modulation (\S\ref{sec:supp_streaming_arch}, Eq.~\ref{eq:se2_warp}); here it is applied at preprocessing time across clip boundaries.

\begin{figure}[!htbp]
  \centering
  \includegraphics[width=0.8\linewidth]{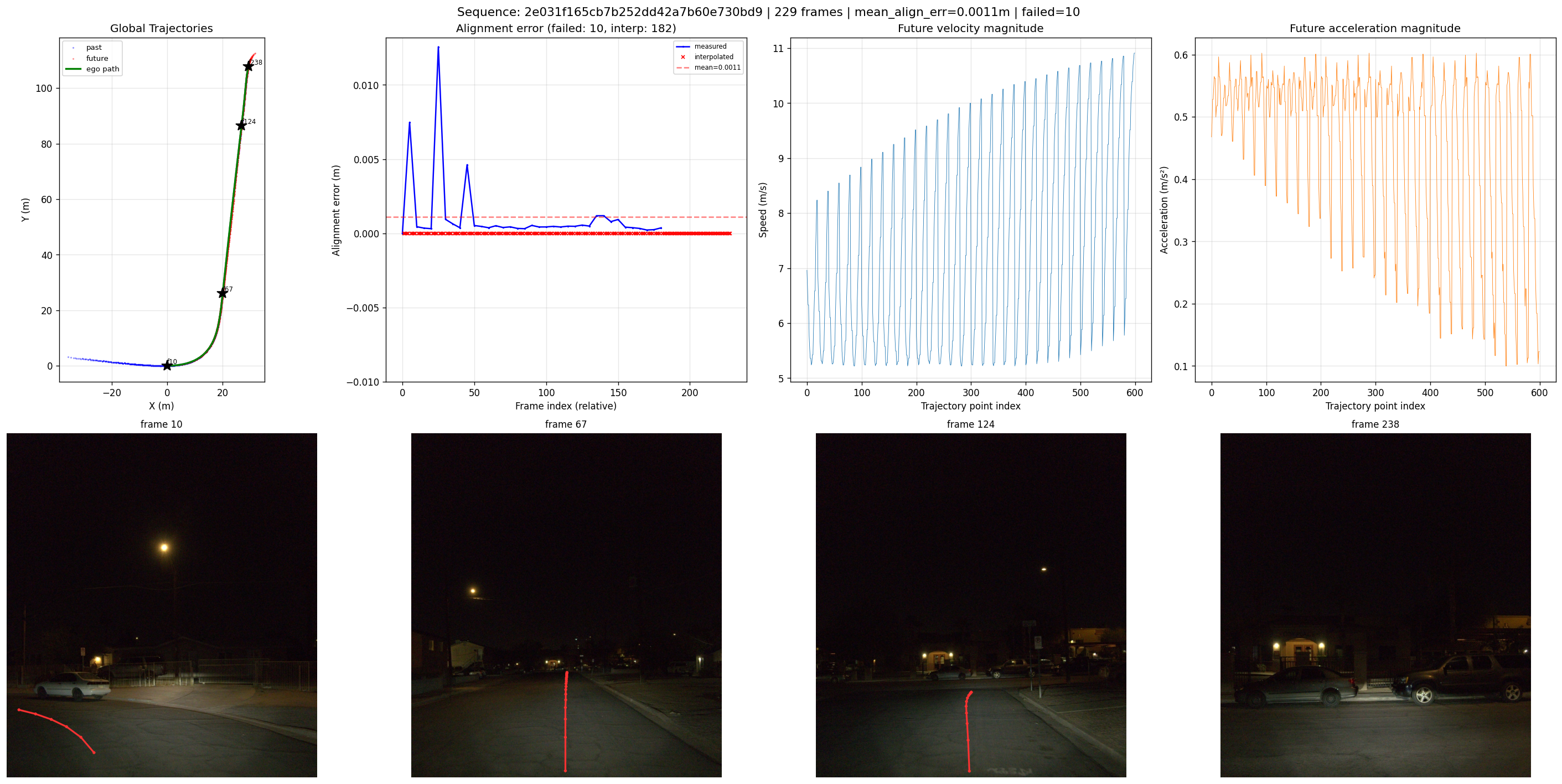}
  \caption{\textbf{Full-sequence pose recovery on a representative WOD-E2E segment ($229$ frames).} Top, left to right: recovered global trajectory in segment-anchor coordinates; per-frame SE(2) alignment residual (mean $\sim$$0.0011$\,m, $10$ inliers per join); speed-magnitude profile across the full sequence; acceleration-magnitude profile. Bottom: sampled front-view frames ($\#32$, $\#67$, $\#124$, $\#198$) with projected ego trajectory overlaid in red. Sub-cm alignment residual confirms that consecutive WOD-E2E clips can be stitched into a single global frame, enabling streaming training (\S\ref{sec:streaming_memory}) over full segments rather than isolated clips.}
  \label{fig:supp_pose_recovery}
\end{figure}

\paragraph{Recovery quality.}
Figure~\ref{fig:supp_pose_recovery} visualizes the recovered pose chain on a representative WOD-E2E segment ($229$ frames). Per-frame alignment residual stays well below $\sim$$0.005$\,m (mean $\sim$$0.0011$\,m, $10$ inliers per join), the recovered global trajectory traces a geometrically consistent path, and the derived velocity and acceleration profiles are smooth and physically plausible across the full segment. Bottom: four sampled front-view frames with the projected ego trajectory overlaid in red. With this preprocessing in place, MindVLA-U1's streaming forward pass can be trained on full segments, so the FIFO memory channel sees coherent context across many tens of frames and the ego-state encoder consumes consistent positions throughout.

\subsection{Qualitative Streaming Examples}
\label{sec:supp_streaming_qualitative}

We provide two visualizations of MindVLA-U1's streaming inference behavior to complement the architectural and ablation results.

\begin{figure}[!htbp]
  \centering
  \includegraphics[width=0.8\linewidth]{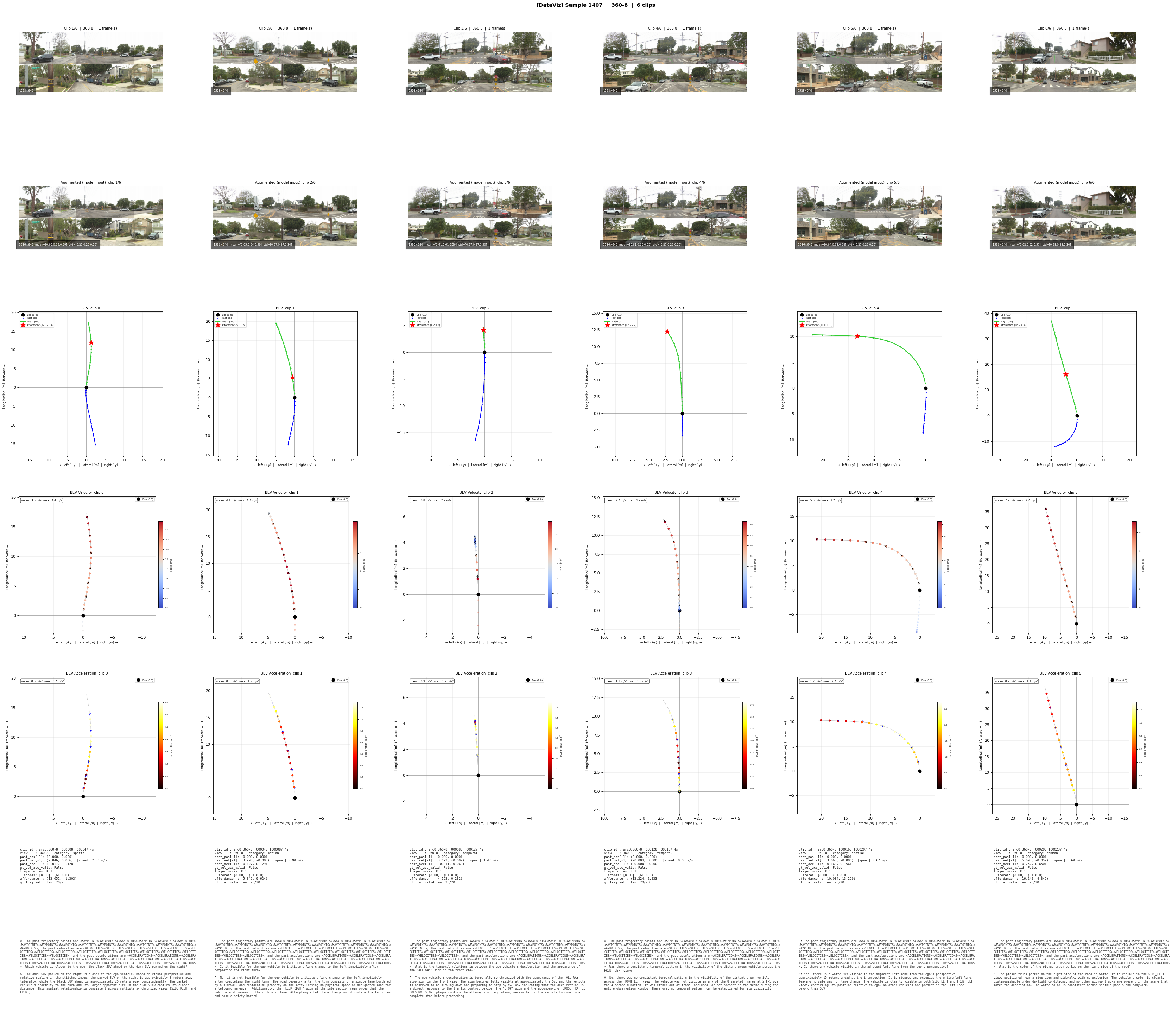}
  \caption{\textbf{Per-frame streaming inference across six consecutive frames of one streaming sample.} Per column: front-view input (top two rows), predicted BEV trajectory (middle), per-waypoint confidence heatmaps (bottom two rows). The streaming memory channel (\S\ref{sec:streaming_memory}, \S\ref{sec:supp_streaming_arch}) carries scene context across frames; planned trajectories evolve smoothly with no fixed-chunk discontinuities.}
  \label{fig:supp_streaming_sample}
\end{figure}

\paragraph{Per-frame streaming inference (Figure~\ref{fig:supp_streaming_sample}).}
Six consecutive frames from a single streaming sample are shown side-by-side, each column reporting one streamed frame: top --- the front-view RGB input image; middle --- the predicted BEV trajectory ($5$\,s horizon, $20$ waypoints) under the streaming forward pass; bottom (heat-mapped panels) --- per-waypoint confidence over the prediction horizon. Predictions evolve smoothly across frames as the streaming memory channel propagates context, with no chunk-boundary discontinuities and no stale waypoints persisting across the stream.

\begin{figure}[!htbp]
  \centering
  \begin{minipage}[t]{0.49\linewidth}
    \centering
    \includegraphics[width=\linewidth]{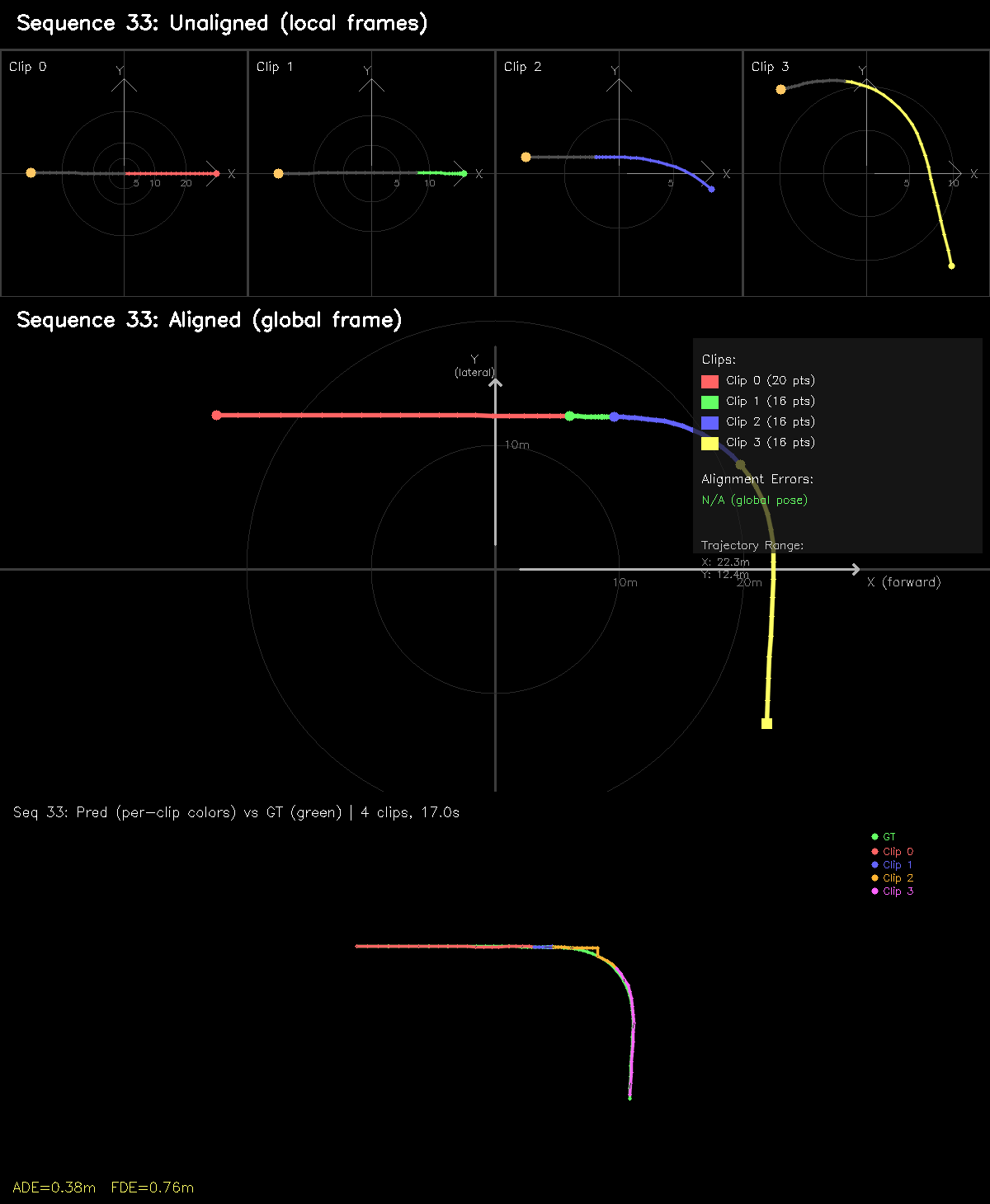}\\[2pt]
    {\footnotesize (a) Seq 33 --- right turn (ADE $0.38$\,m, FDE $0.76$\,m)}
  \end{minipage}\hfill
  \begin{minipage}[t]{0.49\linewidth}
    \centering
    \includegraphics[width=\linewidth]{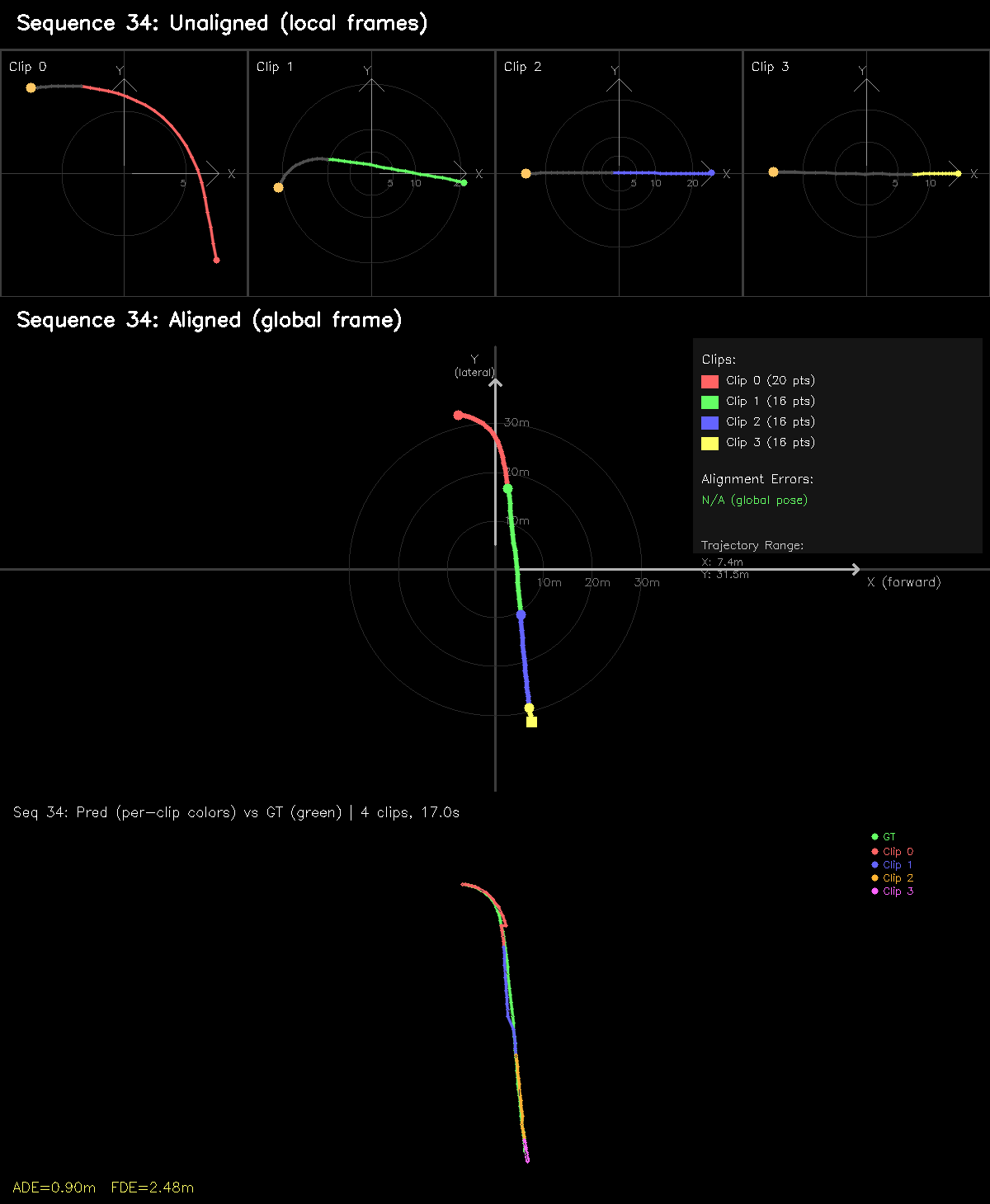}\\[2pt]
    {\footnotesize (b) Seq 34 --- leftward curve (ADE $0.90$\,m, FDE $2.48$\,m)}
  \end{minipage}\\[6pt]
  \begin{minipage}[t]{0.49\linewidth}
    \centering
    \includegraphics[width=\linewidth]{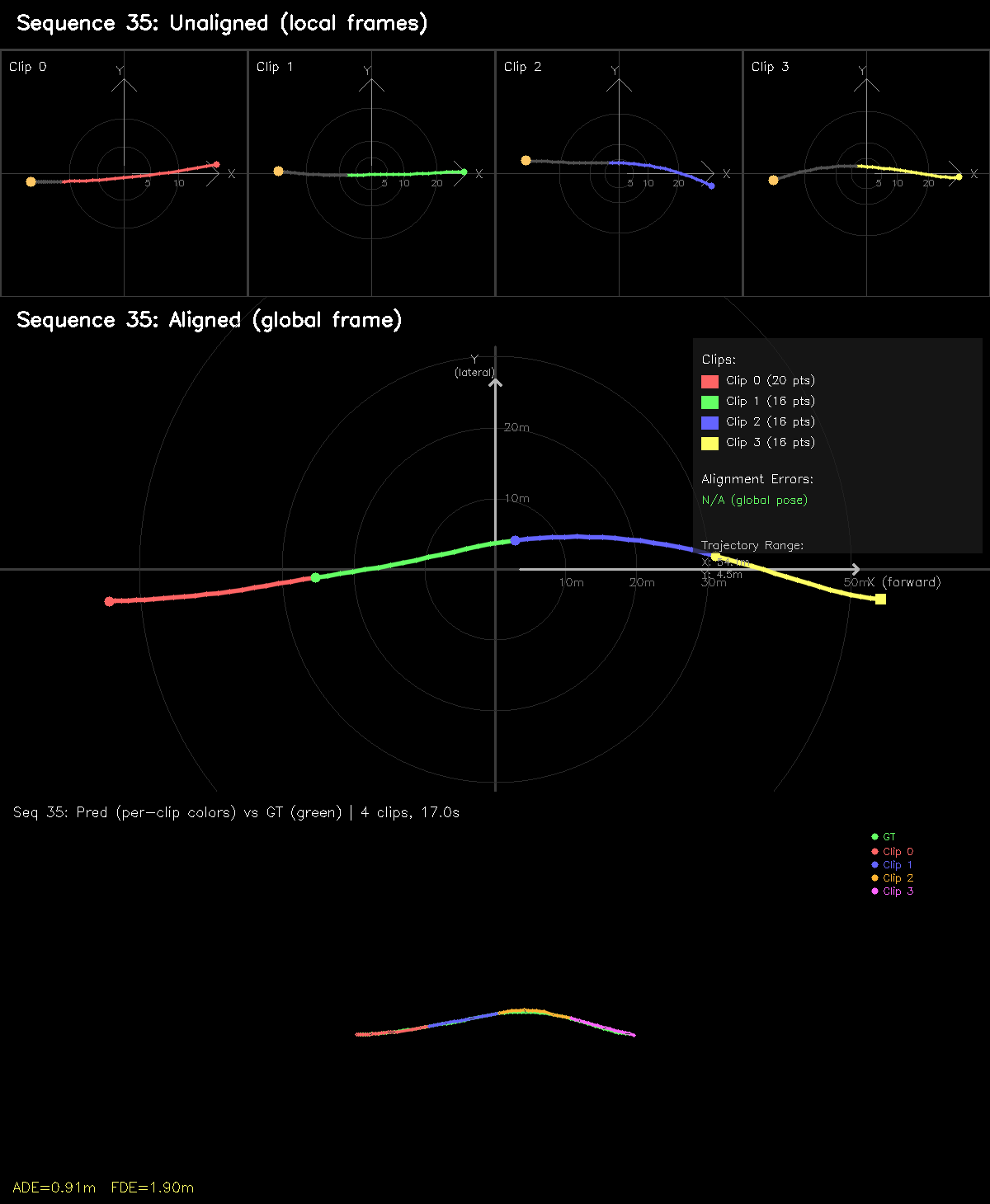}\\[2pt]
    {\footnotesize (c) Seq 35 --- curved cruise (ADE $0.91$\,m, FDE $1.90$\,m)}
  \end{minipage}\hfill
  \begin{minipage}[t]{0.49\linewidth}
    \centering
    \includegraphics[width=\linewidth]{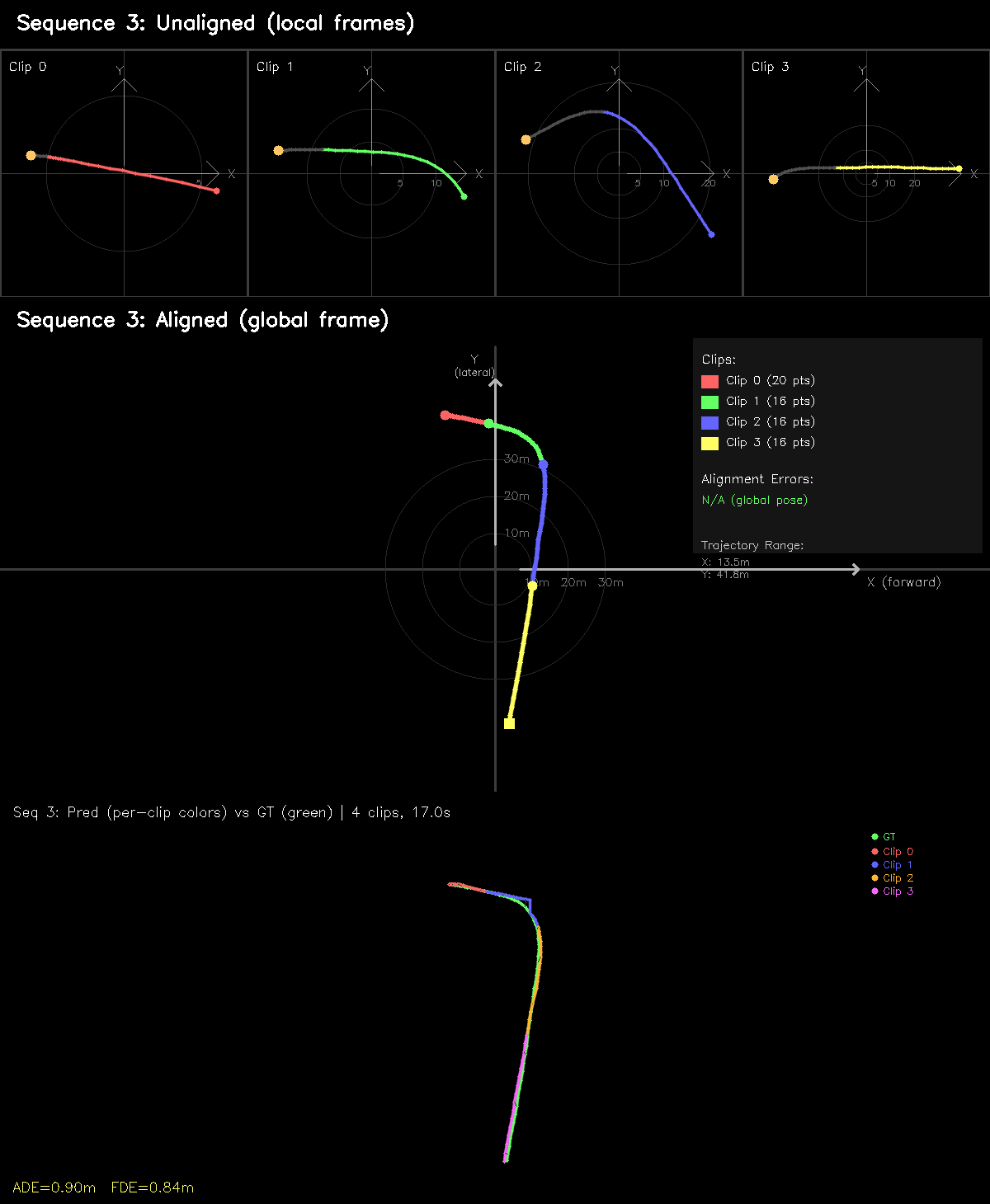}\\[2pt]
    {\footnotesize (d) Seq 3 --- sharp curving maneuver (ADE $0.90$\,m, FDE $0.84$\,m)}
  \end{minipage}
  \caption{\textbf{Long-horizon streaming consistency over $4$ consecutive clips ($\sim$$17$\,s, $68$ waypoints).} Per sequence: top row shows per-clip predictions in their own local ego frames (Clips $0$--$3$); middle stitches the four predictions in a single global frame via the streaming pose chain; bottom overlays the stitched prediction against the logged GT (green). Sub-meter ADEs hold across all four scenarios --- right turn (a), leftward curve (b), curved cruise (c), and sharp curving maneuver (d) --- with end-point errors remaining within a few meters at the $\sim$$17$\,s horizon despite per-clip ego-frame resets at clip boundaries.}
  \label{fig:supp_long_horizon}
\end{figure}

\paragraph{Long-horizon trajectory consistency (Figure~\ref{fig:supp_long_horizon}).}
Four full driving sequences (Seqs $33$, $34$, $35$, $3$) are each evaluated over $4$ consecutive clips ($\sim$$17$\,s, $68$ predicted waypoints per sequence). For each sequence, the top row shows the per-clip predictions in their own local ego frames (Clip $0$--$3$, each starting at the local origin and reset at clip boundaries); the middle panel stitches the four predictions into a single global coordinate system using the streaming pose chain; the bottom overlay compares the stitched prediction (per-clip colors) against the logged ground truth (green), reporting sequence-level ADE and FDE. The global-frame stitches stay coherent across clip boundaries despite per-clip ego-frame resets --- sub-meter ADEs hold across all four sequences (right turn, leftward curve, curved cruise, sharp curving maneuver), and end-point errors stay within a few meters at the $\sim$$17$\,s horizon.

\section{Mixture-of-Transformers Backbone}
\label{sec:supp_mot}

\subsection{Architectural Details}
\label{sec:supp_mot_arch}

This section gives the MoT routing details deferred from \S\ref{sec:unified}. All numbers below describe the deployed \emph{(V,L)+(M,S,A)} grouping (the ``Ours'' row in Table~\ref{tab:abl_mot}).

\paragraph{Modality-routed attention.}
In the MoT variant, each layer splits attention into two independent groups: a \emph{context} group serving visual and language tokens, and an \emph{action} group serving memory, ego-state, and action tokens. The groups use independent Q/K/V/O projections, and the action group uses fewer attention heads ($4$) than the context group, which inherits the dense backbone's head count. Self-attention remains shared across the two groups within each layer, so the context representation is built with action context, preserving the unified-backbone property of \S\ref{sec:unified}.

\paragraph{Per-modality experts.}
The feed-forward stage replaces the single SwiGLU with group-specific experts. The context expert is cloned from the dense backbone and keeps its original intermediate width, while the action expert is initialized from scratch with a narrower intermediate width ($d_{\mathrm{ff}}{=}1024$). Capacity is kept high for perception/language representation and made compact for motor decoding.

\paragraph{Fast-mode subgraph.}
In fast mode, answer/thinking tokens are physically removed before the backbone pass. The remaining memory, ego-state, and action tokens route through the action group, while visual and question tokens are kept as conditioning prefix on the context side rather than decoded into language. This is different from mask-only fast/slow variants: the sequence length seen by attention is smaller, so the fast path can translate into actual compute reduction. The independence of the groups also makes temporal-frequency decoupling --- caching context key-value states from a slow step and reusing them across subsequent fast steps --- structurally straightforward; we leave this extension to future work. The throughput consequences of the dense vs.\ MoT fast paths are measured in \S\ref{sec:ablations_fast_slow} (Table~\ref{tab:fast_slow_fps}).

\section{RL Post-Training: Extended Details}
\label{sec:supp_rl_extended}

This section gives the optimization hyperparameters, training-dynamics, checkpoint-selection, and ADE-rater trade-off details deferred from \S\ref{sec:rl_post_training}.

\paragraph{Hyperparameters.}
Optimizer: AdamW, learning rate $5\!\times\!10^{-7}$ constant, $\beta_1{=}0.9$, $\beta_2{=}0.999$, weight decay $0.1$, gradient norm clip $0.3$. KL regulariser: $\beta_{\mathrm{kl}}{=}0.008$ with the k3 estimator $\beta\!\cdot\!(\exp(r{-}c){-}(r{-}c){-}1)$ against a snapshot of the SFT weights. PPO clip $\epsilon{=}\pm 0.2$. Rollouts: $8$ trajectories per sample, group-scaled rewards. The RFS reward is computed at $4$\,Hz over a $5$-second horizon. RL is run on $8$ GPUs at batch size $1$/GPU; RFS plateaus smoothly without the ``collapse-then-recovery'' pattern reported in some RL-from-rater systems, which we attribute to the conservative $\beta_{\mathrm{kl}}$ and the small constant learning rate.

\paragraph{ADE--rater trade-off under RFS-only reward.}
The detailed eval reveals an interesting trade-off that the headline RFS hides. RL post-training \emph{improves} the rater-matched distance (minADE-5s drops $1.16 \to 1.07$, $-0.09$\,m) while modestly \emph{worsening} the GT-matched distance (ADE-5s rises $2.18 \to 2.22$, $+0.04$\,m). The model has learned to plan trajectories closer to the rater panel, which is exactly what RFS rewards, at the cost of modest divergence from the single logged GT trajectory. This is consistent with the structure of the RFS reward: the ground-truth logged trajectory is one valid behavior but rarely the rater-preferred one in a multi-rater panel, so optimizing RFS pulls the model toward the panel mode. The effect is larger at the longer horizon, which we read as RL primarily sharpening late-trajectory rater alignment.


\end{document}